\documentclass{article}
\PassOptionsToPackage{numbers}{natbib}
\usepackage{datetime}
\usepackage[preprint]{neurips_2024}
\usepackage{blindtext}
\usepackage[labelfont=bf]{caption} 
\usepackage{graphicx}
\usepackage[usenames,dvipsnames]{color}
\usepackage{siunitx}
\usepackage[most]{tcolorbox}
\usepackage[hyphens]{url}
\usepackage[export]{adjustbox}
\usepackage{hyperref}
\usepackage{subcaption}
\usepackage[capitalize]{cleveref}
\usepackage{multirow}
\usepackage{booktabs}
\usepackage{tikz}

\newtcolorbox{white}{colback=white!10!white,boxrule=0pt, top=0pt,bottom=0pt, left=0pt}
\newtcolorbox{blue}{colback=blue!10!white,boxrule=0pt, top=0pt,bottom=0pt, left=0pt}
\newtcolorbox{cyan}{colback=cyan!10!white,boxrule=0pt, top=0pt,bottom=0pt, left=0pt}
\newtcolorbox{red}{colback=red!10!white,boxrule=0pt,top=0pt,bottom=0pt, left=0pt}
\newtcolorbox{green}{colback=green!10!white,boxrule=0pt, top=0pt,bottom=0pt, left=0pt}
\colorlet{bluec}{cyan!20!white}
\colorlet{greenc}{green!20!white}
\colorlet{redc}{red!20!white}
\colorlet{orangec}{orange!20!white}
\colorlet{violetc}{violet!20!white}
\usepackage{setspace}
\usepackage{algorithm}
\usepackage[noend]{algpseudocode}
\algdef{SE}[SUBALG]{Indent}{EndIndent}{}{\algorithmicend\ }%
\algtext*{Indent}
\algtext*{EndIndent}
\algdef{SE}[SUBALG]{ForP}{EndForP}[1]%
{\textbf{for }#1 \textbf{do in parallel}}%
{}
\algtext*{EndForP} 

\makeatletter
\newcommand{\myfnsymbol}[1]{%
  \expandafter\@myfnsymbol\csname c@#1\endcsname
}
\newcommand{\@myfnsymbol}[1]{%
  \ifcase #1
  \or 1
  \or 2
  \or \TextOrMath{\textasteriskcentered}{*}
  \or \TextOrMath{\textasteriskcentered}{*}\TextOrMath{\textasteriskcentered}{*}
  \or \TextOrMath{\textdagger}{\dagger}
  \or \TextOrMath{\textasteriskcentered}{*},\TextOrMath{\textasteriskcentered}{*}\TextOrMath{\textasteriskcentered}{*}
  \fi
}
\newcommand{\affiliationA}{\@myfnsymbol{1}}
\newcommand{\affiliationB}{\@myfnsymbol{2}}
\newcommand{\equalcontributor}{\@myfnsymbol{3}}
\newcommand{\biequalcontributor}{\@myfnsymbol{4}}
\newcommand{\correspondingA}{\@myfnsymbol{5}}

\usepackage{soul,color,colortbl}
\definecolor{forestgreen}{rgb}{0.13, 0.55, 0.13}

\definecolor{lightgray}{gray}{0.9}

\usepackage{placeins}
\usepackage{booktabs}

\newcommand{\photon}{\textit{Photon}\xspace}
\newcommand{\photonclient}{\textit{Photon LLM Node}\xspace}
\newcommand{\photonclients}{\textit{Photon LLM Nodes}\xspace}
\newcommand{\photonserver}{\textit{Photon Aggregator}\xspace}
\newcommand{\photondata}{\textit{Photon Data Source}\xspace}
\newcommand{\photondatas}{\textit{Photon Data Sources}\xspace}
\newcommand{\photonlink}{\textit{Photon Link}\xspace}

\title{{\huge The Future of Large Language Model\\Pre-training is Federated}}

\author{
\large Lorenzo Sani\textsuperscript{\affiliationA,\affiliationB,\correspondingA}
\And\large 
Alex Iacob\textsuperscript{\affiliationA,\affiliationB}
\And\large 
Zeyu Cao\textsuperscript{\affiliationA,\equalcontributor}
\And\large 
Bill Marino\textsuperscript{\affiliationA,\equalcontributor}
\And\large 
Yan Gao\textsuperscript{\affiliationA,\affiliationB}
\And\large 
Tomas Paulik\textsuperscript{\affiliationA}
\And\large 
Wanru Zhao\textsuperscript{\affiliationA}
\And\large 
William F. Shen\textsuperscript{\affiliationA}
\And\large 
Preslav Aleksandrov\textsuperscript{\affiliationA}
\And\large 
Xinchi Qiu\textsuperscript{\affiliationA}
\And\large 
Nicholas D. Lane\textsuperscript{\affiliationA,\affiliationB}
}

\begin{document}

{
\begingroup
\begin{figure}[t]
    \quad
    \begin{subfigure}{0.1275\textwidth}
        \includegraphics[width=\textwidth]{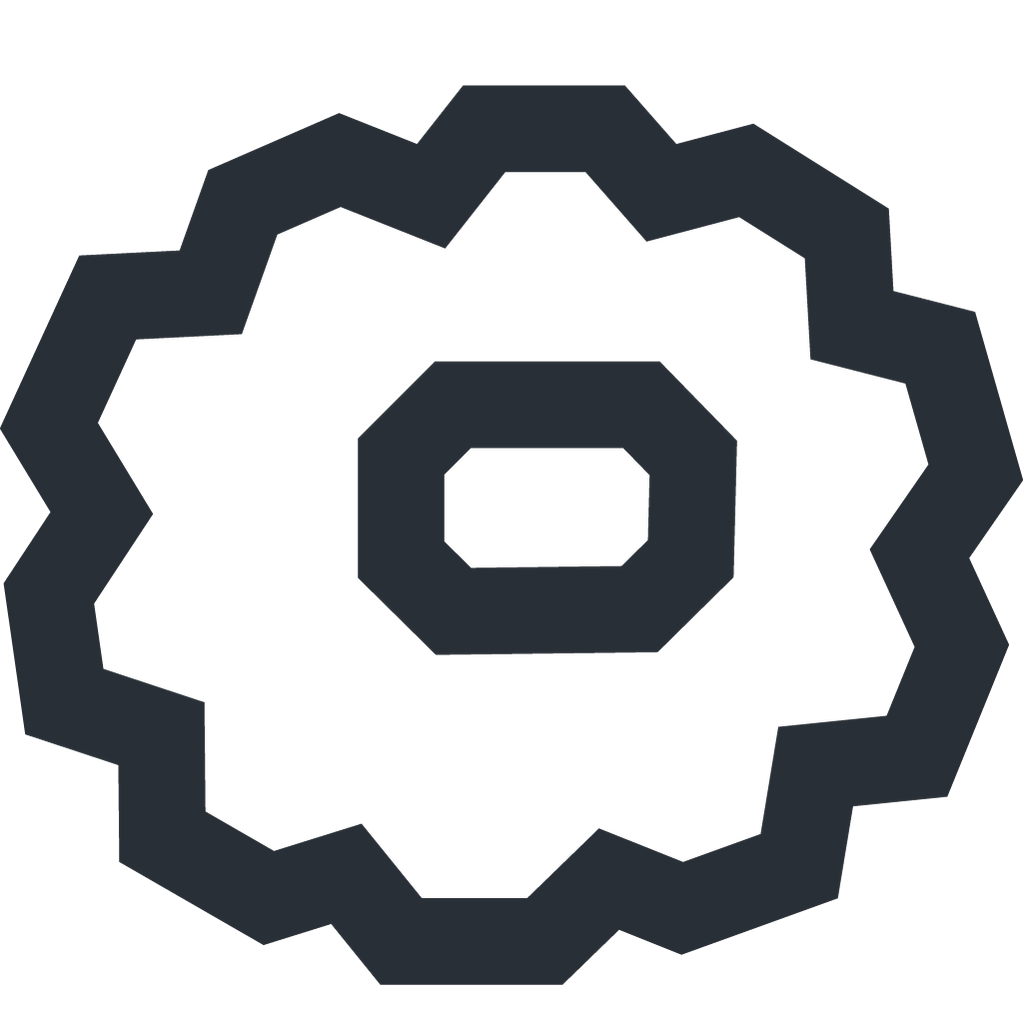}
    \end{subfigure}
    \hfill
    \begin{subfigure}{0.1\textwidth}
        \includegraphics[width=\textwidth]{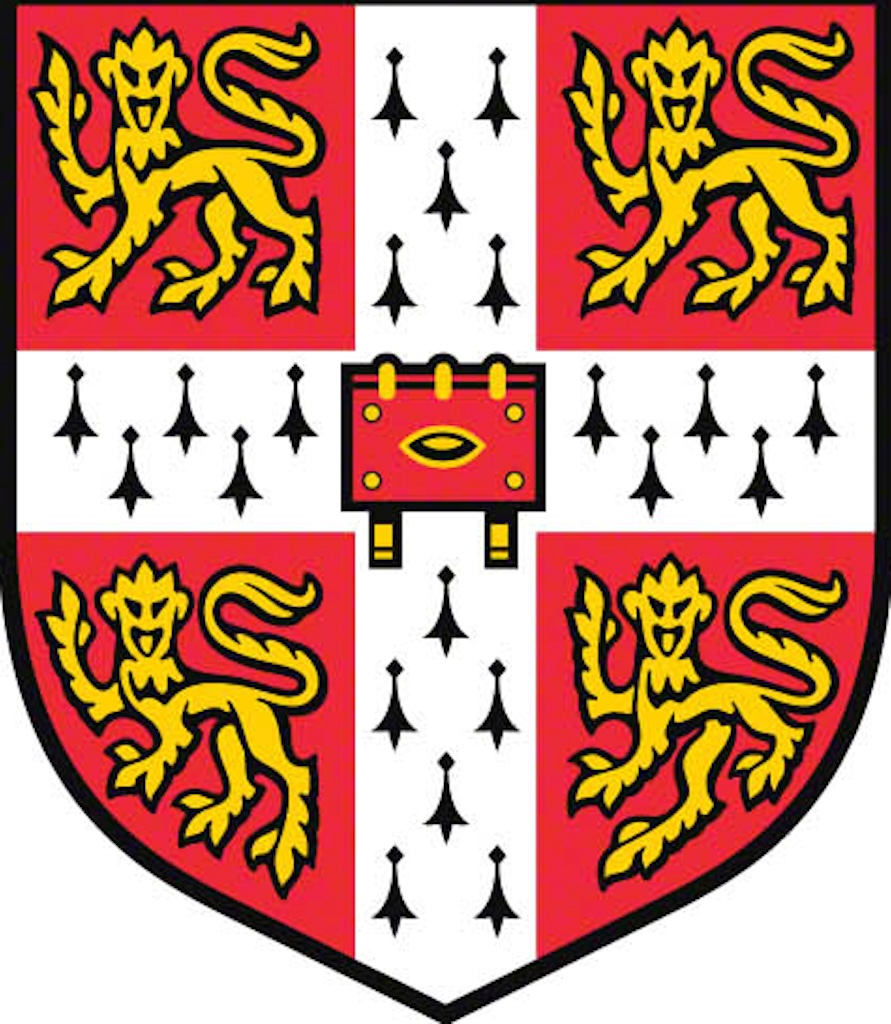}
    \end{subfigure}
    \hfill
    \begin{subfigure}{0.1275\textwidth}
        \includegraphics[width=\textwidth]{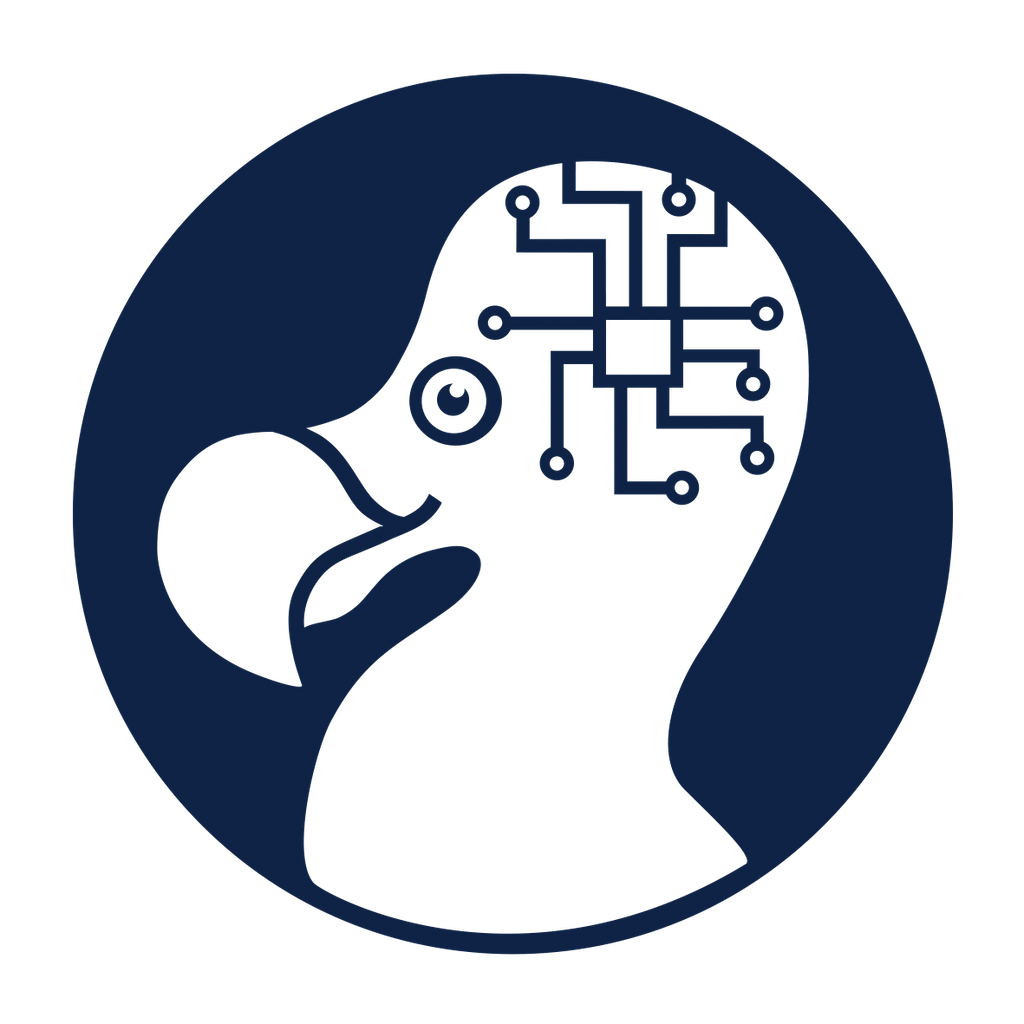}
    \end{subfigure}
\end{figure}
\endgroup
}
\setcounter{figure}{0}
\maketitle

\renewcommand{\thefootnote}{\myfnsymbol{footnote}}
\footnotetext[5]{Lead and corresponding author: \href{mailto:ls985@cam.ac.uk}{\nolinkurl{ls985@cam.ac.uk}}}
\footnotetext[3]{Equal contributors}
\footnotetext[1]{Department of Computer Science and Technology, University of Cambridge}
\footnotetext[2]{Flower Labs}

\setcounter{footnote}{0}
\renewcommand{\thefootnote}{\arabic{footnote}}

\begin{abstract}
Generative pre-trained large language models (LLMs) have demonstrated impressive performance over a wide range of tasks, thanks to the unprecedented amount of data they have been trained on.
As established scaling laws indicate, LLMs' future performance improvement depends on the amount of computing and data sources they can leverage for pre-training.
Federated learning (FL) has the potential to unleash the majority of the planet's data and computational resources, which are underutilized by the data-center-focused training methodology of current LLM practice.
Our work presents a robust, flexible, reproducible FL approach that enables large-scale collaboration across institutions to train LLMs.
We propose a scalable deployment system called \photon to enable the investigation and development of this new training paradigm for LLM pre-training.
We show that \photon can be used by organizations interested in collaborating with their private data sources and computational resources for pre-training LLMs with billions of parameters.
This paradigm would mobilize more computational and data resources while matching or potentially exceeding centralized performance.
We further show the effectiveness of the federated training scales with model size and present our approach for training billion-scale federated LLMs using limited resources.
Thus far, we have used \photon to train LLM models to the size of \textbf{$\mathbf{7}$B parameters} and anticipate larger models being completed in the near future.
Finally, we show that LLM training is highly resilient to the classical challenges of federated statistical and hardware heterogeneity.
Furthermore, we show that convergence is robust to partial participation, opening the avenue for compute-efficient collaborative training.
\photon will help data-rich actors to become the protagonists of LLMs pre-training instead of leaving the stage to compute-rich actors alone.
\end{abstract}

\section{Introduction}

The impressive performance of generative pre-trained large language models (LLMs) and their multi-modal derivations largely owes to their capacity to learn representations at scale~\citep{OgScalingLaws}.
Thus, a handful of well-resourced tech companies and institutions are using increasingly powerful computing facilities in the race to scale up LLMs and dataset sizes to achieve state-of-the-art~(SOTA) performance.
The thousands of hours of training to convergence on thousands of specialized and well-connected hardware accelerators in a single data center incur a high energy and monetary cost \citep{EnergyConsumptionForecast}.
Distributing training across multiple data centers in sparse geographical locations, for those companies who could afford it, would drive the cost even higher due to communication overheads~\citep{comm_eff_distr_learning_survey, training_cost_distr_data}.

\begin{figure}[b]
    \captionsetup{format=plain}
    \centering
    \noindent\includegraphics[width=\columnwidth]{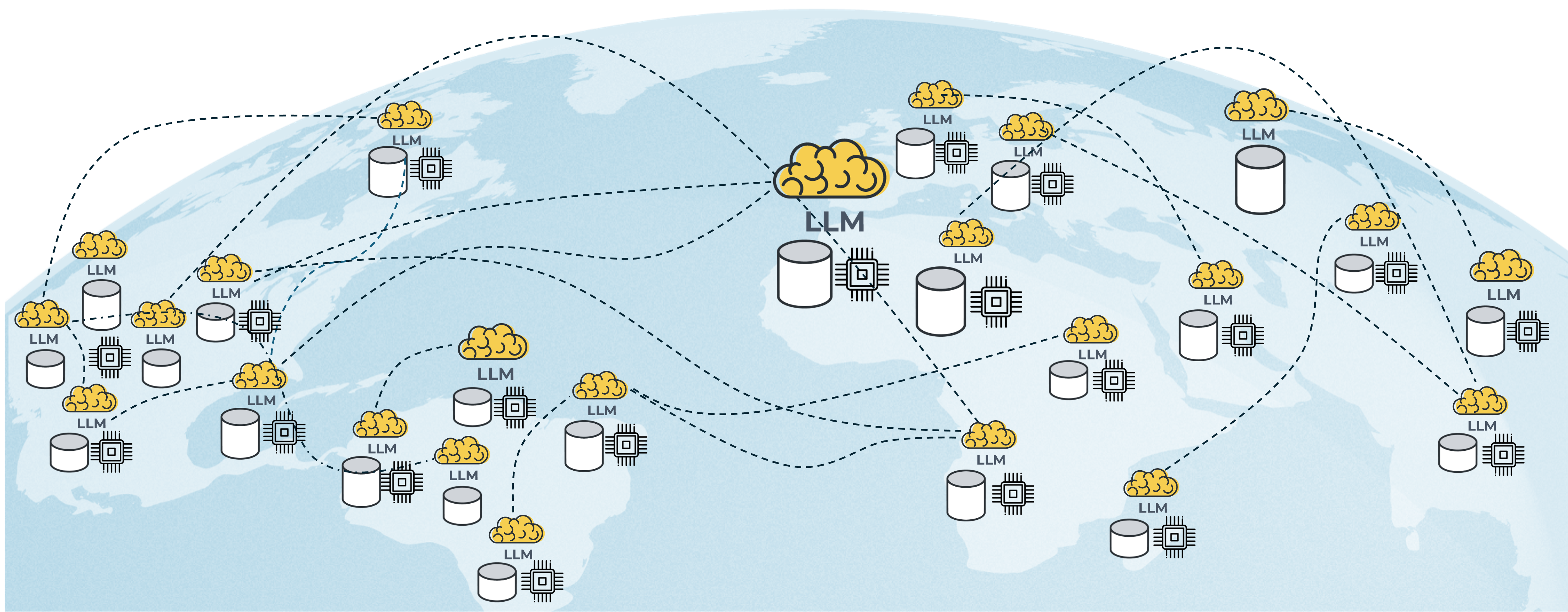}
    \caption{A hypothetical representation of the available data silos around the world. While scraping data from the web has taken foundation models quite far, most data remains under private entities' control. These organizations can collaborate in the federated generative pre-training of large language models to exploit their data towards the common goal of training LLMs they control. The collaborative nature of FL and its low communication requirements make this possible with only moderately powerful hardware, eliminating the prohibitive costs of pre-training.}
    \label{fig:pip}
    \vspace{-0.5cm}
\end{figure} 

\citet{TrainingComputeOptimalLLMs} showed that the effective performance improvement of increasingly large LLMs requires increasingly extensive training datasets.
Most of the world's data sources are unevenly distributed among private actors who often don't want to share it, even if the data regulations of their respective jurisdictions permit such sharing.
Since no organization independently owns the rights to a sufficient amount of text data, the multi-terabyte datasets used in these procedures must be obtained from publicly available sources.
These may include materials potentially protected by intellectual property laws~\citep{NytVsOpenAi} or be otherwise problematic~\citep{trusted_sources_alignment,the_curese_of_recursion, archival_pretraining_data, dp_pre_training_datasets}.
The potential means of collecting non-public data may require arranging independent deals with data providers~\citep{OpenAISpringerDeal} or taking model training directly to these private data sources.
It is anticipated that, by \num{2026}, the amount of high-quality data that LLMs will require to continue to improve their performance will exceed what is publicly available~\citep{WillWeRunOutOfData}.
Where datasets' sizes become constrained, but model growth is unbounded, this can lead to memorization and data leakage amongst current LLMs~\cite {SecuringLLMs,WhatCanWeLearnDataLeakageLaw}.

The next generation of LLMs and foundation models (FMs) will benefit from effectively leveraging more data and computational resources than the centralized paradigm currently makes available.
Therefore, we argue that making FMs' future even brighter requires shifting the dominant training paradigm to a collaborative federated approach, where nodes control considerable (but not immense) computational or data sources according to their abilities.
Using federated learning (FL), we can expand the quality of our models by gaining access to previously untapped data and computational sources~\citep{fed_FMs}.
This broader data availability will, in turn, allow us to increase the size of models that can be efficiently trained compared to the centralized paradigm and to avoid both memorization and data leakage.

As shown in previous works \citep{LocalSGD, DontUseLargeBatchesUseLocalSGD, fedavg}, FL can relax the synchronization requirements of stochastic gradient descent (SGD) to accommodate such poorly connected nodes.
Also, more recent works \citep{DiLoCo,asyncDiLoCo,DiPaCo} showed that Local SGD could substantially reduce the communication overhead of training LLMs in data center settings with homogeneous and heterogeneous computational nodes.
Advancing beyond their valuable independent insights, we argue for an entirely federated approach to LLM training that can automatically balance the workload of multiple actors and privately merge the knowledge derived from their local datasets without divulging them directly to other participants.

We are the \textbf{first} \citep{flowerllm-blogpost} to fully utilize a practical and production-ready \textbf{geographically distributed} federation of \textbf{heterogeneous devices} and \textbf{data sources} for reproducible generative pre-training of \textbf{billion-scale LLMs}.
This work presents \photon, a complete system for pre-training federated LLMs in a collaborative, reproducible, and scalable fashion that can address the main challenges for a flexible execution in \textbf{arbitrary FL cross-silo settings}.
We built it on the \textbf{open-source} FL framework \emph{Flower}~\citep{Flower}.
Furthermore, we make this available as a \textbf{training recipe} with a fully \textbf{transparent} experimental configuration accompanying our results.
We emphasize that federated generative pre-training can be done with \textbf{affordable} hardware configurations by interlinking single nodes containing $1$-$8$ GPUs from standard cloud providers ---rather than renting multiple entire data centers concurrently.
We also argue that federated systems can and must evolve beyond networks of clients holding both compute and data in a world of increasing hardware costs and competing privacy interests.
To achieve this, \photon allows data providers~(sources) to come together with compute providers in private partnerships amongst trusted parties without leaking any such data to other participants.
We demonstrate that pre-training LLMs can be democratized through FL technologies, including data-rich actors \textbf{independent} of their \textbf{connectivity and computing resources}.

Developing this novel federated system led to numerous insights, including the fact that, in defiance of expectations, larger federated LLMs can more easily find a consensus across clients than their smaller counterparts.
Our \textbf{results} show that:
\begin{enumerate}
    \item
    Federated LLM training offers \textbf{competitive performance} with centralized training but with far \textbf{less communication overhead} in a broad range of model sizes up to \textbf{$\mathbf{7}$B parameters}.
    \item Under our system, \photon, \textbf{larger models require less frequent communication} while receiving a more significant boost in generalization performance than smaller ones.
    \item The classical challenges of statistical heterogeneity and partial participation are much less impactful for overparameterized LLMs than the small models typically used in the FL literature. This opens the door to compute-efficient collaboration at a massive scale.
    \item By utilizing our local execution engine, \emph{Pollen}~\citep{Pollen}, to balance loads across computational nodes, we can establish a training pipeline that is \textbf{faster and more robust than} its \textbf{centralized} counterpart.
\end{enumerate}

\section{The Landscape of LLM Training}

Generative pre-trained large language models (LLMs) have demonstrated powerful performance across various natural language processing tasks, leading to rapid and widespread adoption.
They are trained on massive corpora of text, incorporating mixtures of low-quality web-scraped data and high-quality curated datasets~\citep{gpt3, gpt4}.
Recently, several institutions have released their pre-trained LLMs, either in the closed form, such as GPT-4~\cite{gpt4}, Chinchilla~\citep{TrainingComputeOptimalLLMs}, and Gemini~\citep{GEMINI}, or the open-sourced form, such as BLOOM~\citep{BLOOM}, LLaMa~\citep{llama, llama2}, and Falcon~\citep{falcon}.
The scaling laws identified by \citet{OgScalingLaws, TrainingComputeOptimalLLMs} dictate that model size and dataset size should be increased in equal measure to improve model performance best.
These suggest a future race between entities interested in developing state-of-the-art LLMs to grab as many compute and data sources as possible.
Thus, LLMs are headed in a promising direction that can become even more luminous by gaining the trust of private entities possessing an unprecedented breadth of knowledge and computing resources~\citep{fed_FMs}.

In \cref{subsec:centralized_distributed_optimization}, we describe the current landscape for generative pre-training of LLMs with particular attention to the techniques for centralized distributed training.
We discuss federated learning (FL) and its impact as a communication efficient technique in \cref{subsec:fl_background}.
\cref{subsec:fed_finetuning}, for completeness, presents the effort the community has put into matching FL with LLMs fine-tuning.
However, we highlight that our work tackles the far more challenging problem of federated pre-training of LLMs.

\subsection{Centralized Distributed Optimization}\label{subsec:centralized_distributed_optimization}

LLM pre-training is based on two denoising pillars: leveraging huge batch sizes and very long contexts, i.e.,~sequence length of a single input sample.
The combination of model and batch sizes forces training LLMs to scale SGD beyond the confines of a single GPU.
Thus, it is often necessary (a) to process more training samples in parallel and (b) to split the model across GPUs if it cannot fit within the memory of one GPU.
In the following, we present a subset of optimizations and techniques the research community proposed to make this challenging training recipe possible.
As we will see in \cref{subsec:fl_background}, FL follows as a natural next step in the sequence of distributed training optimizations previously adopted.
Given the importance of these optimizations even for FL's local training step, our work supports most of them, as described in \cref{subsec:pollen_parallel_execution,subsec:reproducibility_by_design}.

\subsubsection{Data and Model Parallelism}

The number of trainable parameters and the size of the datasets make LLM training very sensitive to the stochastic fluctuations of the optimizer used, thus requiring a solid and robust regularization achieved by the denoising properties of enormous batch sizes.
Distributed Data Parallelism (DDP) replicates the model $N_d$ (number of devices) times across different devices to enable training with sufficiently large batch sizes.
Then, DDP splits each enormous batch into micro-batches, subsequently fed into each replica in parallel.
The number of samples in every micro-batch depends on the available memory in the hardware accelerator; as such, it is often referred to as device batch size.
Thus, each replica produces the gradients of its training micro-batch locally and accumulates them via simple summation.
This accumulation is eventually followed by a synchronization step where these gradients are averaged and applied to the model.
Since the batch size of an ML algorithm plays a crucial role in its convergence~\citep{LargeBatchTraining, OpenAIDota}, this synchronization step happens after each replica has accumulated sufficient micro-batches to match the desired true batch size.
Modern DDP implementations such as the one used by \texttt{PyTorch Distributed}~\citep{PyTorchDistributed} use the \texttt{Ring AllReduce} algorithm popularized by \texttt{Horovod}~\citep{Horovod} to reduce the gradients across replicas.
The algorithm is implemented with low-level collective communication libraries like Nvidia's \texttt{NCCL} or Facebook's \texttt{Gloo}.
\texttt{Ring AllReduce} scales proportionally to the amount of data being transmitted and the latency of the slowest connection between workers in the ring, making it highly sensitive to the network topology connecting GPUs. 

In the case of LLM training, the model size often mandates computing gradients over batch sizes larger than what GPUs can support, even with DDP.
Thus, practitioners are forced to use a micro-batch size dictated by the number of training samples that fit in the GPU, which is, in turn, limited by the VRAM occupancy due to model parameters and activations.
These factors result in a complex interplay between the model's memory consumption and the training pipeline's efficiency.
The resulting memory-efficiency trade-off is even more complex when considering how many ways the models may be partitioned across GPUs to reduce their VRAM consumption.

For sufficiently large models, the parameters must be split across GPU workers so that they fit in VRAM.
Traditionally, this was achieved via Model Parallelism (MP)~\citep{ModelParallelism, MeshTensorflow}, splitting individual tensors and their computation vertically across GPUs and communicating as needed.
While model parallelism can reduce the per-GPU memory requirements linearly by the model parallelism degree $N_m$ (number of model splits), it induces communication overheads that are tolerable in a single-machine context with high inter-GPU bandwidth but cannot scale to multi-machine contexts.

\subsubsection{Fully Sharded Data Parallelism}

An alternative approach is to shard the model into equally-sized units amongst GPUs, with units potentially containing multiple layers, and then materialize the units, as necessary, to compute the activations during the forward pass via collective communication.
This form of fully-sharded data parallelism~\citep{FSDP_ZeRO,FSDP_Pytorch} reduces memory consumption linearly in the data-parallelism degree $N_d$ while increasing communication by \num{1.5}$\times$ compared to standard DDP.
It is also possible to combine this methodology with other techniques for reducing memory consumption, such as model parallelism, activation checkpointing~\citep{ActivationCheckpointing}, or CPU offloading~\citep{ZeroOffload}.
Activation checkpointing functions by recomputing activations during the backward pass rather than saving them in memory.
CPU offloading refers to offloading either data to system RAM or operations to the CPU.

Given the memory requirements, it is crucial to consider the minimum number of GPUs necessary to train one model with reasonable efficiency, e.g., without extreme CPU offloading.
This estimate provides a lower bound on the hardware that an organization requires to participate in any distributed training of an LLM and is generally determined by the model size and the micro-batch size.
The micro-batch size is also known as the device batch size since it is bounded by the number of samples that can fit on a GPU device together with the model.
By accounting for each organization's independent resources and manipulating the amount of local computation they do together with the micro-batch size, our method can relax the lower bound within reasonable limits, thus allowing even organizations with weak hardware to participate in distributed training.

\subsubsection{Bottlenecks for generative pre-training of LLMs}

High-quality public language data is liable for exhaustion within the next decade, while low-quality language data may be exhausted in a few decades~\citep{WillWeRunOutOfData}.
When taken together with a growing interest from both individuals and corporations in constraining what data can be scraped from the Internet or used to train an LLM, this limits the size of models that can be efficiently trained.
Circumventing this either requires costly independent deals with data providers~\citep{OpenAISpringerDeal}, leaps in the effectiveness of synthetic data generation for model training~\citep{MLForSyntheticData}, or significant improvements in the data efficiency of ML optimization. 

Similarly, hardware accelerators with enough memory and throughput to support LLM training are scarce and increasingly unaffordable to anyone interested except for the best-funded organizations.
Hundreds to thousands of such accelerators are required with extremely high monetary costs for training and inference~\citep{BLOOM, llmcost}.
Moreover, such accelerators must be largely uniform in specifications to avoid stragglers and need to be extremely well-connected both within a computational node and across nodes due to the synchronization requirements of data parallel or fully shared data parallel SGD~\citep{PyTorchDistributed, FSDP_ZeRO, FSDP_Pytorch}.
These constraints pose significant system challenges that can be mitigated by extremely expensive data center configurations or by developing heterogeneity-aware pipelines that exploit the resources available despite their heterogeneity.
The difficulties described above scale with model size, as splitting the model across the memory of several GPUs further increases communication demands~\citep{FSDP_ZeRO,FSDP_Pytorch} and the optimization gaps.

\subsubsection{Mitigation of LLMs demands}

The wider AI community has gone to great lengths to circumvent the high requirements of LLM utilization and thus expand the data pool we can draw upon.
Such endeavors have widely focused on locally running efficient inference with pre-trained model weights \citep{llama,llama2}, quantization \citep{Quantized1bitLLMs}, or parameter-efficient fine-tuning \citep{hu2021lora}.
The recent work proposes Petal \citep{DistributedInferenceAndFineTuningLargeModelsOverTheInternet}, which enables wide-scale collaboration for inference and parameter-efficient fine-tuning over the Internet by joining the resources of multiple parties.
Their work assumes that clients provide inference jobs or data for fine-tuning, and servers execute the LLM inference/fine-tuning in a distributed pipeline parallel fashion.
Their system makes clients responsible for holding the trainable parameters to enable efficient fine-tuning, with servers merely running forward passes and returning gradients for the pre-trained weights they store.

We argue that while methods exploiting pre-trained weights are highly beneficial to the broader community, they do not resolve the bottleneck of pre-training.
Thus, the community is bound to the decisions made by organizations capable of training LLMs regarding their structure and the data they use, and they may suffer downstream consequences for such reliance.
Performance degradation may happen since the pre-trained model, further fine-tuned for a specific downstream task, is the dominant upper bound for the downstream model's performances.
Thus, we propose developing systems capable of the distributed \emph{pre-training} of LLMs that can accommodate the variety of hardware and data available in the AI community.
Nevertheless, this new paradigm may include training from scratch if the model's size permits, starting from pre-initialized weights and retraining the entire model instead of only a limited subset of parameters.

\subsection{Federated Learning and Local SGD}\label{subsec:fl_background}

Traditional machine learning involves using a central server that hosts the machine learning models and all the data in one place.
In contrast, when using federated learning (FL) \cite{fedavg} algorithms, client devices collaboratively learn a shared global model using their local compute and data. 

FL aims to collaboratively learn a global model while keeping private data on the device.
Such a training pipeline occurs over multiple communication rounds.
During each round, a fraction of the clients are selected and receive the global model from the server.
Those selected clients then perform local training with their local data before sending the updated models back to the central server.
Finally, the central server aggregates these updates via averaging~\citep{fedavg} into a \textbf{pseudo-gradient}. It then uses a federated optimizer~\citep{FedMOM} to update its model based on the pseudo-gradient, creating a new global model.
Then, this three-stage process is repeated.

Federated optimization has several properties that make it suitable as a new paradigm for LLM training: (a) it does not require the private data of participants to be directly shared, (b) it can naturally incorporate Differential Privacy \citep{brendan2018learning} or Secure Aggregation \citep{45808} to compile with privacy regulations at an actor level, (c) it allows for more control over the optimization and has less restriction on the connectivity as each data-source can be associated with a series of updates. Crucially, since FL allows previously unseen data to be accessed during training, it reduces the likelihood of data memorization and leakage, which have become increasingly common as model size has increased~\citep{SecuringLLMs,WhatCanWeLearnDataLeakageLaw}.

Despite these advantages, FL comes with two major challenges in the form of data and systems heterogeneity~\citep{AdancesAndOpenProblems}. Data heterogeneity refers to the tendency of naturally generated and partitioned data to fail the IID assumption, which is common in centralized ML optimization. Systems heterogeneity refers to the ability of client hardware to vary in terms of computational ability, communication efficiency, or availability~\citep{ScaleAndSystemDesign}. Both forms of heterogeneity are highly relevant for the distributed training of LLMs. Data heterogeneity may arise from participants holding texts in different languages, belonging to various genres, or varying in complexity. Systems heterogeneity is primarily present in the computational ability of the GPUs held by a specific client, their VRAM, and number, as well as the communication efficiency of said client.

Local SGD~\citep{LocalSGD,LocalSGD_Trade_Offs_At_Scale} is a data-parallel training paradigm where each replica applies independent gradient updates to its parameters for several local steps before averaging parameters rather than gradients.
While mathematically equivalent to the FedAvg~\citep{fedavg}, the context in which it is applied, lowering the communication costs in centralized training, lacks the hardware and potential data heterogeneity specific to FL clients.
Unlike previous work~\citep{DiLoCo}, we intend to go beyond local SGD towards fully federated training inclusive of the broadest possible range of participants.

\subsection{Federated Fine-tuning and Parameter Efficient Fine-tuning of LLMs}\label{subsec:fed_finetuning}

Until now, full federated pre-trained LLMs have not been accomplished because researchers could not solve the dual challenges of its communication overhead and pre-training large models on resource-challenged devices.
That said, researchers, eager to reap the benefits of federated learning, have nonetheless made progress on federating downstream LLM training tasks whose computational and communication demands are lower, such as fine-tuning, parameter-efficient fine-tuning (PEFT), and prompt-tuning.
We outline these developments here for a complete reference.

For example, \citet{hilmkil2021scaling} use FL to fine-tune all the model parameters of ALBERT~\citep{ALBERT} and BERT~\citep{BERT}, reaching 90\% of the accuracy achieved by a centrally trained model on text classification tasks.
Meanwhile, \citet{riedel2023} found that BERT fine-tuned in an FL setting could perform as well as a centralized model on multilingual text classification tasks.
\citet{wang2024public} and \citet{weller2022} also conducted federated fine-tuning on private data but preceded it with centralized pre-training on public data.
They found that this workflow improved fine-tuning accuracy, even when pre-training only on a 1\% sampling of a large public corpus or if the client's data was non-IID. 

Much progress has also been made on federated PEFT, whose computational and communication hurdles are lower than those of federated fine-tuning.
Researchers have shown that a model that has been subject to federated PEFT can outperform the original pre-trained model \citep{zhang2024building}, outperform siloed client models \citep{fan2023fatellm}, and even outperform federated fine-tuning \citep{kuang2023federatedscopellm, jiang2023lowparameter}, including in non-IID scenarios \citep{pmlr-v232-malaviya23a}, but with far lower computation and communication costs because clients only need to update and transmit the smaller set of parameters.
Federated LoRA, for example, may consume as little as 0.058\% of the communication cost of federated full fine-tuning  \citep{fan2023fatellm}.
To reduce these costs even further, \citet{xu2022training} add Noise Contrastive Estimation to reduce memory demands and \citet{xu2024fwdllm} add backpropagation-free training to improve memory and time efficiency.
Meanwhile, to address the more significant impact that non-IID client distributions can have on federated PEFT performance, \citet{babakniya2023slora} precede federated LoRA with a round of federated efficient sparse fine-tuning, reducing the performance gap while keeping combined training time low.
Differently, \citet{kim-etal-2023-client} abandon a global model altogether, instead using FL to train client-specific models that benefit from sharing data via FL but are more resilient to client drift.

Federated prompt tuning, wherein clients tune a set of continuous soft prompts appended to input prompts, has also demonstrated its effectiveness.
This technique can significantly reduce the number of model parameters that clients must train and transmit compared with full fine-tuning.
Models fine-tuned this way perform better than siloed client transfer learning~\citep{kuang2023federatedscopellm}, better than centralized training by benefiting from improved generalisation~\citep{lester2021power, zhao2023fedprompt}, and also reduce the impact of client drift due to non-IID data \citep{che2024federated, FedPromptTuning} by leveraging an adaptive optimization method.

\section{Design Principles for Federated Generative Pre-Training of LLMs}\label{sec:towards_fed_llms}

We propose a federated LLM generative pre-training paradigm to create new pre-trained models over which the previously mentioned fine-tuning techniques may be applied.
Thus, we aim to disentangle the broad, distributed community of researchers and practitioners that build upon LLMs from the willingness of large organizations to provide open-source weights.

This section proposes a series of inclusive principles necessary for such federations to be effective and incorporate them into the design of our system, \photon. The principles we chose are meant to tackle the foundational issues of federated LLM pre-training with a particular focus on data and hardware inclusivity, robustness, and efficiency. 

\paragraph{Broad Access to Data and Compute:}\label{par:broad_acess}
The ability to train an LLM should depend on the data that a participant or a group possesses rather than unrestricted access to hardware. Thus, we aim to transform organizations that right now can serve as data providers only~\citep{OpenAISpringerDeal,OpenAINewsPublishersDeal}, when their ownership of the data is respected, into active participants in the LLM training process. We believe that incorporating such contributors directly into the federated learning process and offering them an incentive to participate, obtaining a model performing well on their data, is the natural next step in the proliferation of generative AI generally and LLMs in particular.
While some data-rich organizations may be unable or unwilling to invest in computing power, we believe that \textbf{voluntarily} partnering with a compute-rich yet data-poor one in a federated training context provides an excellent avenue for their participation.
Thus, \photon aims to create a broad network of data producers and consumers where a single client may offer computing resources, data sources, or both to the training process.

\paragraph{Limited Communication Requirements:}\label{par:limited_coms}
Pre-training should be possible without the strong synchronization requirements of standard data-parallel training~\citep{FSDP_ZeRO} to accommodate geographically distributed and poorly connected participants.
Our federated learning solutions allow orders-of-magnitude reductions in communication frequency compared to centralized solutions.
While such improvements boost efficiency for all participants in the federation, they also offer particular benefits to data sources that may have been underrepresented in the past.
For example, incorporating private data silos from regions with a lower internet presence may help alleviate the challenge of NLP for low-resource languages~\citep{LowResourceLanguagesSurvey,LowResourceNMTSurvey}.

\paragraph{Broad Hardware Inclusivity:}\label{par:broad_hardware}
Organizations with valuable data must be able to participate in the federated network, even with poorly connected or limited hardware resources.
In the case of clients that may hold distributed compute nodes lacking the connectivity necessary to support the high-bandwidth \texttt{Ring AllReduce} algorithm necessary for classical data-parallel training~(e.g., Infiniband), we perform local federated training over their compute nodes and transparently aggregate the node updates before sending results to the main server.
Suppose a client suffers from either computational or communication limitations.
In that case, we account for it by adjusting the maximum round duration or allowing them to modulate the amount of local training they undertake.
Thus, our system can accommodate a broad set of potential participants with hardware varying from a few medium-powered GPUs to small data centers.

\paragraph{Scalable Local Training Pipelines:}\label{par:scalable_local_training}
Given that the participants in federated training are likely to vary in terms of model-size requirements, for example, if they prefer smaller models capable of faster training more specific to their data or more appropriate for the amount of data held by each of them, a federated training system must be effective across a wide range of model sizes. Thus, we build our system to support local training pipelines appropriate for all scales from billions-sized models with fully-shaded data parallelism~\citep{FSDP_ZeRO,FSDP_Pytorch} to simple data parallel training~\citep{PyTorchDistributed} and CPU offloading~\citep{ZeroOffload}. While this work focuses on the most challenging large-scale setting, our codebase is highly adaptable to all requirements.
\section{\photon Design}\label{sec:system_design}

\photon is a system for federated generative pre-training of large language models (LLMs), allowing organizations with \textit{limited communication capabilities} to collaboratively train a federated learning (FL) model using their private data sources.
True to FL principles, \photon guarantees that private data is never exchanged or moved without the consent of the participants, which is a necessary condition as opposed to the centralized paradigm that uses public data.
Following our principle of \textit{broad hardware inclusivity}~(\cref{par:broad_hardware}), \photon allows FL participants to execute \textit{highly scalable local training pipelines} with heterogeneous computing resources in amount and type (e.g.,~different generations of GPU), although sufficient (e.g.,~just one server with a few data center-level GPUs).
Furthermore, to offer \textit{broad access to data and compute}, \photon allows clients that do not hold both data and compute to self-organize into a distributed network of voluntary data producers and consumers while keeping their data private from all other participants.
\photon aims to satisfy the system requirements and tackle the challenges for generative pre-training of LLMs, which FL can sometimes exacerbate.
This training paradigm requires infrequent communication of unprecedentedly big payloads (for FL) through unreliable heterogeneous networks.
Every FL system is prone to performance degradation due to dynamic client availability, stragglers, hardware heterogeneity, and unexpected dropouts.
Although some of these are present even in the centralized paradigm (e.g., stragglers and dropouts), their impact on this FL application is more severe as they may result in wasting several precious GPU hours.

The features described above make \photon the first of its kind to successfully address the system challenges for federated generative pre-training of LLMs, paving the way for a more democratic training environment for federated LLMs.
Our system provides three building blocks to execute such a federated pipeline: one central server orchestrating the FL, called \photonserver; at least one distributed \photonclient supporting scalable local training; and one distributed \photondata for each participating client running on the edge.
We describe these components and their interaction in \cref{subsec:photon_architecture}, while we further detail \photon's scalability properties in \cref{subssec:communication_computation_scalability}.

\subsection{\photon Architecture}\label{subsec:photon_architecture}

\begin{figure}[ht]
    \centering
    \includegraphics[width=1.0\textwidth]{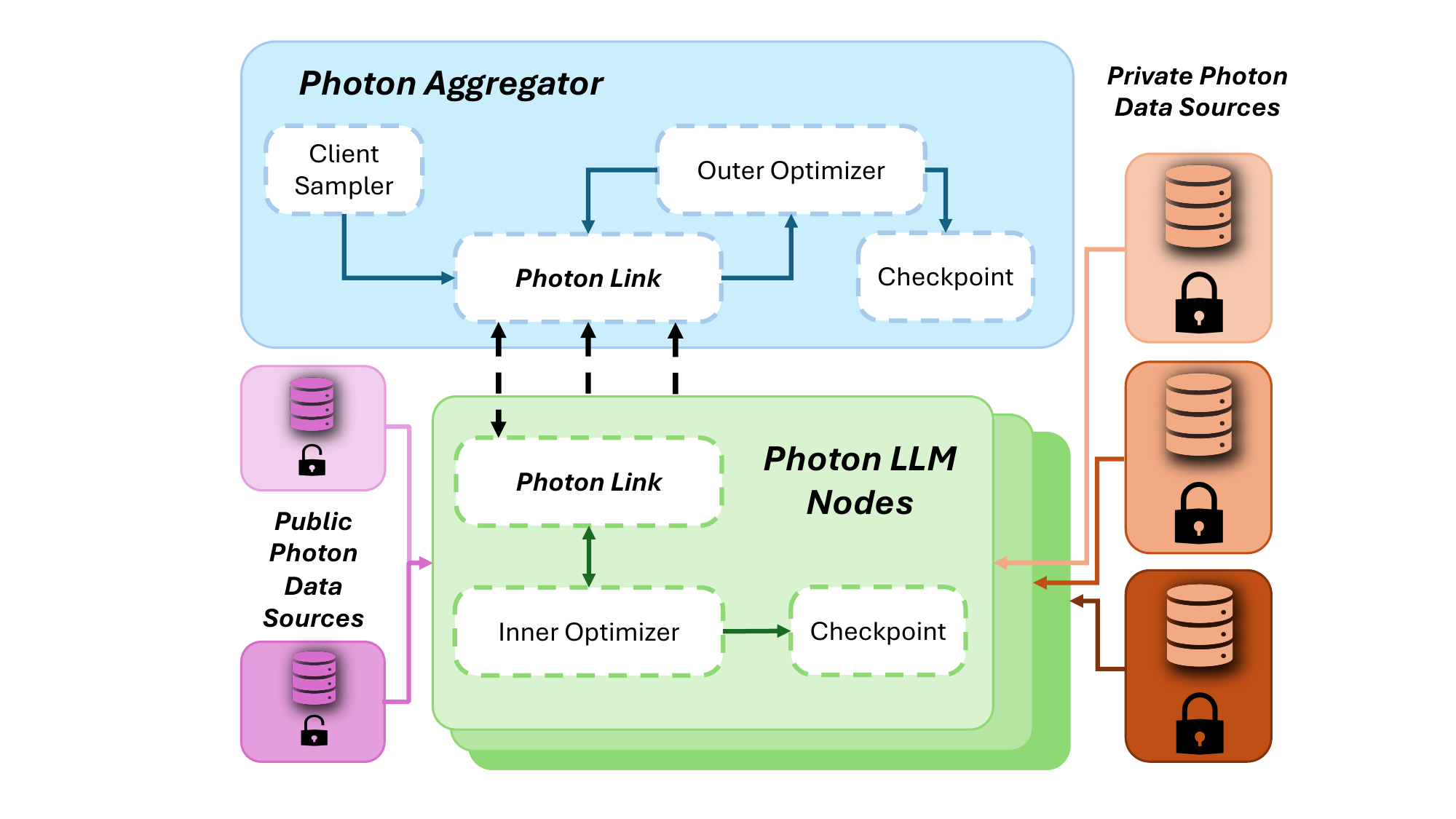}
    \caption{The diagram describes the \photon's three principal components - \colorbox{bluec}{\photonserver}, \colorbox{greenc}{\photonclient}, \colorbox{orangec}{\textit{Private} \photondatas} and \colorbox{violetc}{\textit{Public} \photondatas} - and their sub-components. Arrows describe how such elements work together or exchange messages. The \photonserver can communicate with the \photonclients only through the \photonlink. The instances responsible for storing the data samples, the \photondata, can uniquely stream to the \photonclient bonded to them.}
    \label{fig:diagram:photon}
\end{figure}

As discussed below, \photon's three core components are co-designed to account for system requirements and challenges of the specific setting featured in this work, following the principles extensively described in \cref{sec:towards_fed_llms}.
The main objective of this discussion is to promote reproducibility and further developments of the federated pre-training of LLMs.
\cref{fig:diagram:photon} presents a high-level overview of the architecture of \photon, comprehensive of its three core components and their sub-elements.

\paragraph{\photonserver:}
The central component of standard FL systems is the server responsible for \textit{orchestrating the training} procedure and maintaining the \textit{global training state} up to date.
\photonserver plays this role by using the sub-components described here.
Before starting any distributed step in the federated pipeline, the \photonserver summons the \textit{client sampler}, which assesses how many \photonclients are available and selects a number of them depending on the requirements of the optimization algorithm.
After such selection, the \photonserver broadcasts the current global model and the training instructions to the appointed \photonclients.
Any message between the \photonserver and the \photonclients travels through the \photonlink, which acts as the communication gateway for both components.
\photonlink also supports secure communication protocols, such as HTTPS and the more complex secure aggregation \citep{SecAggOG}.
The message payload includes, but is not limited to, model updates, training and evaluation instructions, client training or evaluation metrics, and side information (when considered safe to share).
The \photonlink receives the results of the distributed task executing at the \photonclients, such as model updates.
The \textit{outer optimizer} applies the gradient updates to the global model -- aggregation step -- using the appointed optimization algorithms, such as FedAvg~\citep{fedavg}, FedProx~\citep{FedProx} or FedOPT~\citep{FedOPT}.
When the method is associative, the \textit{outer optimizer} further improves its efficiency by taking advantage of asynchronous partial aggregation of the client updates.
The \photonserver guarantees robustness in case of failures by keeping the state of the FL continuously \textit{checkpointed}.
This state contains the global model parameters, the snapshot of the appointed outer optimizer's attributes, and bookkeeping metrics such as the timestamp and the elapsed time.
We assume that the \photonserver is well-connected to the \photonclients through the Internet and doesn't require particularly powerful computing capabilities, i.e.,~a commercial-level CPU is sufficient.
However, a \photonserver requires sufficient storage capacity to maintain the training state, which can be on a local disk or in the cloud.

\paragraph{\photonclient:}

The \photonclient is the distributed component of \photon, which executes the \textit{local training pipeline} of the federated optimization.
Each client in the federated population has a \photonclient that can arbitrarily connect or disconnect to \photon's runtime at any time during training.
A \photonclient leverages several sub-components to fulfill the requested training tasks robustly.
The \textit{inner optimizer} trains the latest model received by the \photonclient on the local client data using a selection of distributed optimization methods~\citep{PyTorchDistributed,FSDP_ZeRO}, which allow increasingly high memory savings at the cost of additional inter-GPU communication.
Given that our clients may range from single machines with a few GPUs to clusters of machines, the \photonclient selects the preferred local training method following the client's hardware resources, their connectivity, and the model's size.
The \photonlink can exchange model parameter updates and training instructions with \photonserver.
Given the large size of such models and the abundance of pruning techniques available, the \photonlink also handles model compression.
We do not prune the model by default and only use lossless compression.
Similarly to the \photonserver, the \photonclient has a \textit{checkpointing} component that guarantees speedy recovery in case of disruption.
In addition to the model parameters, this local state must track the optimizer and data loading index states and the number of full epochs the client has completed.
Since federated quantity skew may result in clients having very different local dataset sizes, the checkpoints save the dataset state privately without any server control that could impact the optimization's fairness.
Our assumptions for the minimal hardware capabilities of \photonclients require them to have at least a few data center-level GPUs sufficiently supported by the RAM and CPU.
We also assume the \photonclients tightly connect to their coupled \photondata through the Internet.

\paragraph{\photondata:}
FL requires strict guarantees regarding where to store data samples and how to exchange them.
To systematically satisfy such requirements, we designed \photon's data storage component, \photondata, to tie uniquely to a \photonclient.
\photondata provides the training pipeline executing on the \photonclient with \textit{continuous streaming of data samples} respecting the pipeline's throughput demands.
Exploiting this architecture, an institution possessing a large silo of data samples that may want to participate in the FL can either acquire computing resources to build a \photonclient or reach an agreement with a trusted third party with extensive computational resources.
\photon's design permits such confined collaborations that are easier to achieve than sharing global data and computing resources.

\begin{algorithm}[ht]
\caption{\photon execution pipeline} \label{alg:photon}
\small
\begin{onehalfspace}
\begin{algorithmic}[1]
\Require{Number of rounds $T$, training population $P$, number of clients per round $K$, hyperparameters $H$}
\begin{cyan}
    \Procedure{PhotonServer}{$T$, $K$, $H$, $P$}
    \State{$\theta^0 \gets \mathtt{InitModel}(H)$} \Comment{Init on the server or sample a client and extract model weights}
    \For{each round $t=1,2,3,\ldots,T$} 
    \EndFor
    \Indent
        \State{$\mathcal{C} \sim \mathcal{U}(P, K)$} \Comment{Sample $K$ clients at random from the population}
        \ForP{$k \in \mathcal{C}$} \Comment{Each sampled client in parallel executes the local training}
            \State{$\theta_k^t, \mathcal{M}^t_k \gets \Call{PhotonClient}{k, \theta^t, H}$}
            \State{$\Delta^{t}_{k} \gets \theta^t - \theta_k^t $}
        \EndForP
        \State{$\Delta^t \gets \frac{1}{|C|}\sum_{k\in C}{\Delta^{t}_{k}}$} \Comment{Aggregate pseudo-gradients $\Delta^{t}_{k}$ from clients}
        \State{$\theta^{t+1} \gets \mathrm{ServerOpt}(\theta^t, -\Delta^t, t)$}   \Comment{Apply pseudo-gradient to the global model}
        \State{$\mathcal{M}^{t+1}_k \gets \mathtt{AggMetrics}(\mathcal{M}_k^t \, \forall k \in \mathcal{C})$} \Comment{Aggregate metrics across clients}
        \State{$\mathtt{Checkpoint}(\theta_k^{t+1})$} \Comment{Checkpoint model}
    \EndIndent
    \Return{$\theta_k^{T+1}$}
    \EndProcedure
\end{cyan}
\begin{green}
\Procedure{PhotonClient}{$k, \theta^t, H$}
            \State{$\mathcal{D}_k \gets \mathtt{BindStream}(k)$} \Comment{Bind \photondatas to a merged data stream $\mathcal{D}_k$}
            \State{$I_k \gets \mathtt{GetNodes}(k)$} \Comment{Extract hardware configuration $I_k$}
            \If{$\mathtt{HasInfiniband}(I_k)$}
                \State{$B_k \gets \mathtt{CalcBatchSize}(I_k)$ } \Comment{Binary search for batch size $B_k$ with static initial guess}
                \State{$\theta^t_k, \mathcal{M}_k^t \gets \mathtt{TrainClient}(\theta^t, \mathcal{D}_k, B_k, H)$} \Comment{Use DDP or FSDP based on model size}
            \Else
                \ForP{node $i \in I$}  \Comment{In every node of the current client do FL}
                    \State{$B_k^i \gets \mathtt{CalcBatchSize}(i, I_k)$ }
                    \State{$\mathcal{D}_k^i \gets \mathtt{PartitionStream}(i, \mathcal{D}_k)$} \Comment{Split the client data into $|I|$ shards}
                    \State{$\theta^t_i, \mathcal{M}_i^t \gets \mathtt{TrainClient}(\theta^t, \mathcal{D}_k^i, B_k^i, H)$} \Comment{Use DDP or FSDP based on model size.}
                \EndForP
                 \State{$\theta^t_k \gets \frac{1}{|I|}\sum_{i\in I}{\theta^{t}_{i}}$} \Comment{Partially aggregate node models}
                 \State{$\mathcal{M}^t_k \gets \mathtt{AggMetrics}(\mathcal{M}_i^t \, \forall i \in I_k)$} \Comment{Partially aggregate metrics across nodes}\EndIf
            \State{$\mathtt{Checkpoint}(\theta_k^t, \mathcal{D}_k)$} \Comment{Checkpoint model and dataset state}
            \State{$\theta_k^t \gets \mathtt{PostProcess}(\theta_k^t, \mathcal{M}_k^t)$} \Comment{E.g.,~apply differential privacy or compress the model}
            \State \Return{$\theta_k^t, \mathcal{M}^t_k$}
\EndProcedure
\end{green}
\end{algorithmic}
\end{onehalfspace}
\end{algorithm}

\subsection{Workflow Overview}\label{subsec:workflow_overview}

We are now ready to discuss the workflow of using \photon, which is succinctly illustrated in \cref{alg:photon}.
A standard application for \photon is that of multiple independent institutions possessing siloed data sources and computational resources. 
Each of these institutions will need to instantiate a \photonclient.
The \photonserver could be one of the collaborating institutions or an appointed third party to provide further privacy protection and reliability.
Regardless of the hosting policy, the server is responsible for initializing a model or sourcing it from a participating client~(\colorbox{bluec}{\small L.$1$}), coordinating rounds and sampling participating clients~(\colorbox{bluec}{\small L.$3-4$}), of sending the model parameters to the clients and collecting results~(\colorbox{bluec}{\small L.$5-7$}), and finally building new federated models~(\colorbox{bluec}{\small L.$8-9$}).
Each institution participating with data sources must set up \photondata to supply the \photonclient with training and evaluation samples.
There might be private agreements between organizations holding data sources without possessing any computing resources and other organizations in the opposite situation.
These two organizations will collaborate to provide one coupling of \photondata and \photonclients.
Public \photondata is also permitted, which can be used for initializing the global model in a pre-training fashion, regularizing the FL, and mitigating the impact of the heterogeneity of \photondatas.

Once all the parties have set up \photon's components, the \photonserver starts the FL by initializing the global model~(\colorbox{bluec}{{\small L.$2$}}) and sampling the \photonclients~(\colorbox{bluec}{{\small L.$4$}}) participating in the first iteration.
The following steps describe such iteration, repeating until the global model converges.
The \photonserver broadcasts the model parameters to the sampled clients to execute the described local training~(\colorbox{bluec}{{\small L.$6$}}).
Then, the \photonclients start the local training using the data samples they query from the \photondata~(\colorbox{greenc}{{\small L.$13$}}).
To efficiently complete the training task, \photonclients assess the hardware resources available~(\colorbox{greenc}{{\small L.$14$}}) and choose the optimal execution strategy.
They take advantage of the fast connectivity between GPUs on different nodes if available~(\colorbox{greenc}{{\small L.$15$}}), or they execute another level of federated optimization if connectivity between nodes is poor~(\colorbox{greenc}{{\small L.$19$}}).
In the second case, an additional partial aggregation step~(\colorbox{greenc}{{\small L.$23-24$}}) between nodes is necessary to complete the local training.
At the end, clients produce model updates they send back to the \photonserver~(\colorbox{greenc}{{\small L.$25$}}). Before sending these updates~(\colorbox{greenc}{{\small L.$27$}}), clients apply local post-processing steps such as gradient clipping and noise injection in the case of differential privacy~\citep{brendan2018learning,AdaptiveClipping}, or model compression~\citep{ModelCompressionSurvey} for lower server-client communication costs.
After all models have been transmitted, the iteration comes to the last step: the aggregation.
The \photonserver produces a new version of the global model using the updates received~(\colorbox{bluec}{{\small L.$9$}}).
The next iteration will use this new version of the global model parameters.
The validation of the training procedure is always performed on a held-out split of the samples available in any \photondata.
\photondata ensures this split is preserved and streamed to the \photonclients when asked to validate the current version of the model.
As such, \photon allows for distributed evaluation of the local and global models at any time during the FL. 
The FL can also be validated with public \photondata to take advantage of well-defined benchmarks, such as perplexity on Cleaned Colossal Common Crawl~(\emph{C4})~\cite{C4}, which allows meaningfully comparing the capabilities of the FL models. 

This ability to choose between public and private data enables \photon to assess performance in a global context~(i.e.,~on public data or the average performance across client test sets) or a personalized one~(i.e.,~on one client's specific private test set). The ability to account for local performance opens the door for various fair aggregation algorithms~\citep{QFedAvg, TERM} and allows for easy evaluation of bias when accounting for AI regulation.

\subsection{Communication-computation Scalability}\label{subssec:communication_computation_scalability}

\photon is built for a setting that crucially differs from the settings of previous proposals and applications \citep{fedavg,ScaleAndSystemDesign}.
Prior system designs have mostly focused on cross-device FL, which often has opposite demands compared to the cross-silo setting discussed in this work.
While the very first FL application \citep{fedavg} has been deployed for training a model with $860$K parameters, \photon aims to train scalable LLMs of unprecedentedly large sizes~(i.e.,~several billions of parameters).
While sampling from populations of millions of devices with heterogeneous data and hardware brings many challenges \citep{LargeBatchTraining,Zhou_2024}, \photon is intended for cross-organizational use cases where the number of participants may reach the hundreds.
Furthermore, although FL is often challenged by executing ML pipelines on constrained devices (such as smartphones, laptops, or other edge devices) \cite{fl_constrained_survey}, \photon assumes clients either have data center-level hardware accelerators, though limited in number, or that some clients have such accelerators but insufficient data and may be willing to join a private voluntary partnership with a data producer which does not own such hardware.
These differences with previous works implicitly reinstate how the \photon's scalability properties impact the FL.

The communication step is considered the major bottleneck in cross-device FL settings, as local training is performed by constrained devices with small models on a few data samples.
This is not necessarily the case for \photon's use cases - we assume the \photonserver's connectivity with the \photonclients is fast and stable based on industry-level access to the Internet.
One may expect the communication steps to become even more problematic when training an LLM, as each communication has a payload size several times greater than the usual models in cross-device settings.
However, in compute-intensive tasks such as LLM training, the bottleneck introduced by external communication between client and server is negligible compared to the computational time.
Indeed, such computationally intensive tasks represent the predominant factor in the runtime for \photon's execution, as long as the number of local steps chosen is the range proposed in this work.
This does not apply to cross-device situations, where the limited hardware and data available on edge devices restrict the number of possible local steps that can produce a meaningful aggregation on the server.
This cross-silo approach differentiates \photon from previous work~\citep{DatasetGrouper}, which trained clients using $4$ local steps instead of $500$ with a batch size of $16$ rather than $512$ as \photon uses.
Given that this amount of local steps approaches FedSGD~\citep{fedavg}, which is known to be impractical in wall-clock terms even for small CNNs due to communication inefficiency, we argue that \photon's approach for cross-silo represents a far more promising avenue.

Optimizing the efficiency of the compute-intensive part of the system requires more attention than optimizing the communication part, even though the latter can still take advantage of the several optimizations the community has proposed~\citep{ModelCompressionSurvey}.

For an FL pipeline with a fixed-size model, the local training step takes a linear time with the number of local steps performed by the client, assuming that no particular system optimization has been performed and the workload for a single step fits the hardware resources.
The transmission of a model's parameters with fixed precision takes linear time in the number of parameters, assuming a constant transmission speed without chunking or compression.

\paragraph{Parallel and Sequential Computation:} Given the paramount nature of clients' computing capabilities, compute efficiency is the most important aspect to consider when evaluating the efficiency of federated learning.
To properly understand the scaling properties of FL, it is first important to distinguish between \emph{parallel} and \emph{sequential} optimization steps.
In stochastic gradient descent (SGD) context, each sequential step~(batch gradient) changes the model, causing future sequential steps to calculate gradients w.r.t.~a different set of parameters.
In contrast, samples within a batch are evaluated in parallel and thus produce a gradient w.r.t.~the same parameters before they are aggregated and applied to the model.
As such, for standard data-parallel techniques~\citep{PyTorchDistributed,FSDP_Pytorch}, the parallelization unit is a batch, and each batch completed moves the model forward.
Once federated optimization is considered, each client will perform a series of local updates in parallel and thus produce different models.
However, these models are all computed w.r.t.~the same model received at the start of the federated round.
Thus, when the server averages client updates, it averages over all the local steps of the clients, moving in the approximate mean direction of all clients and thus benefiting from access to more data and lower variance~\citep{Zhou_2024}.
Thus, FL serves to provide access to a practically larger batch size~\citep{DontUseLargeBatchesUseLocalSGD} with a far greater variety of data available, enabling us to satisfy the data demand of scaling laws~\citep{WillWeRunOutOfData} at the cost of having to train on more samples in parallel rather than sequentially. 

There is strong evidence from the cross-device FL literature~\citep{OnLargeCohortTraining,Zhou_2024}, which our evaluation supports in the case of LLMs, that training on a larger proportion of the total client population during a round has diminishing returns as long as unbiased random sampling is applied.
Thus, for a sufficiently large training population of size $P$~(e.g., between $32$ and $128$ clients) training for a fixed number of local steps, it may be possible to boost the statistical efficiency of training by $5-10 \times$ over full participation by simply sampling only $10\%$ of clients each round without hurting model quality.
While a client's data may be overlooked in a given round, it will eventually be seen and incorporated into the model.
\section{\photon Implementation}

This section discusses the details of \photon's implementation starting from the three principal components described in \cref{sec:system_design}.
As for its software infrastructure, \photon is a distributed system consisting of a set of microservices~\citep{SystematicLitReviewMicroServices}, where each microservice tackles a well-defined component of the LLM training pipeline.
We hope this description allows any group of organizations interested in the collaborative training of LLMs to replicate and build upon our design effectively.

Both \photonserver and \photonclient are microservices implemented using \texttt{Python}, \texttt{Pollen}~\cite{Pollen}, \texttt{Flower}~\citep{Flower}, and \texttt{PyTorch}~\citep{PyTorch}.
We have chosen \texttt{Flower} and \texttt{PyTorch} due to their technical maturity and ease of customization.
In particular, \texttt{Flower} provides \photon with the underlying communication layer and the basic FLOps for orchestrating the FL training.
We use the latest version of this framework, recently released and named \texttt{Flower Next}\footnote{\texttt{Flower Next} is the name given to \texttt{Flower}'s latest version (\texttt{1.9.0}) at the time of writing this manuscript.}.
We added some customizations to the framework to enable checkpointing operations and improve reproducibility, e.g., reproducible sampling.
\texttt{PyTorch} is the machine learning framework on which all the local training operations of the \photonclients are based.
We opted for \texttt{Pollen}'s edge components for managing internal communication and collaboration at \photonclient level.

The entire configuration of each given execution (i.e.,~the training session) is structured as a hierarchical set of \texttt{YAML} files, which are parsed and processed by the \photonserver and the \photonclients using \texttt{Hydra}~\citep{hydra} – an open-source \texttt{Python} framework that simplifies the development complex applications.

\photondata is implemented on a server with high storage capacity using an instance of \texttt{MinIO} \citep{minio} -- a high-performance open-source object store -- that is supported with \texttt{Amazon S3}-compatible \citep{s3} API.
For \photondata to be sufficiently compelling with the requirements of a \photon execution, the server is backed by an array of solid-state disks (SSDs), providing the best cost-random IO performance trade-off.
As various management tools and \texttt{RESTful API} client libraries are available to take full advantage of \texttt{S3}-based solutions, we built a \texttt{Python S3 client} built on top of the \texttt{boto3} \citep{boto3} library tied to the \photonclient.

The checkpointing sub-components in \photon require storing large binary objects with a frequency that depends on the setting utilized.
Such binaries are mainly composed of model parameters that cannot be lossy-compressed to maintain the consistency of the execution.
Since \texttt{MinIO} provides persistent storage of binary blobs, we also use it to store training checkpoints of the \photonserver and the \photonclients.
These components have a \texttt{Python S3 client} to communicate with the checkpoint \texttt{S3} bucket similarly to that implemented for \photondata.

\subsection{Parallel Execution within \photonclients} \label{subsec:pollen_parallel_execution}

The local execution engine used in our solution is based on \texttt{Pollen}, which can adaptively take advantage of the underlying GPU configuration at a \photonclient level.
Depending on the model size, a single computer with one GPU per \photonclient might provide insufficient hardware resources to execute the pre-training.
\texttt{Pollen} orchestrates the collaboration of the single training processes on the GPUs available with \texttt{PyTorch Distributed}~\citep{PyTorchDistributed}.
Thanks to \texttt{Pollen}, a \photonclient can dynamically set arbitrary collaboration topologies of GPUs through single or multiple \texttt{PyTorch Distributed} environments, with support for a range of optimization strategies appropriate to the hardware of a client and the size of the model.

\paragraph{Single-GPU Training} In scenarios where the model fits within the VRAM of one GPU with a sufficient device batch size to keep pace with the rest of the federation, \photonclient defaults to single-GPU training.
While this strategy may seem inappropriate for all but the smallest model sizes, FL's high parallelization potential allows the efficiency of single-GPU training to scale linearly with the total number of GPUs available in the federation.
Since FL requires no intra-round communication, this is possible even when poorly connected.
\paragraph{Multi-GPU Training} For scenarios where a single client has multiple well-connected GPUs available, \photon can exploit the \texttt{PyTorch} implementations of Distributed Data Parallel~(DDP)~\citep{PyTorchDistributed} or Fully Sharded Data Parallel~(FSDP)~\citep{FSDP_Pytorch} training.
DDP constructs replicas across GPUs, computes gradients independently w.r.t.~separated batches, and then synthesizes these gradients on a per-batch basis at a high communication cost.
For situations where a model replica is too large to fit on a single GPU with a reasonable batch size, FSDP shards the model together with its optimizer states and gradients by splitting the layers into units that are distributed across GPUs and materialized as needed at $1.5\times$ the communication cost~\citep{FSDP_ZeRO} of DDP. 

\paragraph{Multi-Machine Training} For settings where clients may own more than one GPU-equipped machine, \photonclient can optimize the communication cost further than standard \texttt{PyTorch Distributed} by adaptively deciding what technique to apply.
For very well-connected clusters, \photonclient sets up a \texttt{PyTorch Distributed} process group containing all GPUs and uses DDP or FSDP~(if the model is too large for one GPU).
However, when inter-machine connectivity is poor~(i.e., it cannot match the speed of high-bandwidth interconnection such as Infiniband NDR or RoCEv2), \photonclient creates disjoint dataset partitions of \photonclient's \photondatas.
Then, it maps each disjoint partition to one of the nodes and treats all of them as separate clients in a sub-federation.
Each node then produces a model trained on its shard.
These models are then partially aggregated by the client's lead node and transparently sent to the federated server as a single client update.
In distributed settings, islands of nodes with high-bandwidth connections or even complex hierarchical structures are common~\citep[see sec.1, Hardware Heterogeneity]{FSDP_Pytorch}.
For such scenarios, \photon applies the same procedure over islands of well-connected machines rather than single nodes, using DDP or FSDP across the nodes in the island.

The partially aggregated result is then transparently sent to the \photonserver, which combines it with the other clients' results.
Thus, while \photon is intended for federated training, it can also enhance training efficiency within single organizations when they possess multiple nodes or entire data centers that are geographically distributed and/or poorly connected.

\subsection{Streaming Data Sources}\label{subsec:streaming_data_sources}

Given the heterogeneous nature of naturally distributed data~\citep{DatasetGeography}, we built our system to support creating clients with heterogeneous data.
Our implementation, based on the underlying MosaicML \verb|StreamingDataset| abstraction, does not enforce a one-to-one mapping between clients and data sources; rather, it enables clients to draw upon arbitrary data streams with complete control over the sampling procedure within and across streams.
This flexibility enables the creation of a distributed system of data and compute providers, which extends the federated network.
Rather than having a fixed set of clients, all of which hold data and compute, the topology becomes enriched with data nodes that provide streams to compute-enabled clients in exchange for access to the final model.
To serve as data producers, clients must offer a means of accessing and locating their data buckets to the compute clients willing to use them during training.

By streaming, caching, and optionally pre-tokenizing or compressing the data, we avoid the large storage overheads of transferring it from data-producing clients to data-consuming clients.
Furthermore, we allow data sources to be arbitrarily swapped out between rounds and for newly generated data to be incorporated into the federated training process.
With pre-tokenization enabled, the computing resources of data-producing clients may also be effectively utilized in the training process, freeing up the computing clients' resources.
\section{Methods}\label{sec:methods}

The system design described so far executes real federated learning (FL) orchestration on clients holding heterogeneous hardware.
To test its efficacy, we imbue it with a series of experimental tools to enable modeling any potential federated configuration given available hardware and data using the same pipeline as a production scenario.
This section describes how we have made \photon appropriate for scientific experimentation.
Furthermore, it presents the experimental design of our evaluation meant to investigate if FL can match or even exceed the potential of centralized training.

\subsection{Reproducibility by Design}\label{subsec:reproducibility_by_design}

As discussed in \cref{sec:towards_fed_llms}, our most challenging aims are broad participation and inclusivity in the federated orchestration.
We also target perfect reproducibility of both the system and results to allow for robust and meaningful research.
The path towards democratizing the generative pre-training of LLMs inevitably involves disclosing training recipes, experimental details, and open-source and validated code.
Then, inspired by other works, such as \citet{meta_opt}, we rely on open-source resources as much as possible.

We used the popular open-source libraries maintained by Mosaic Research \citep{mosaicml} for our local training pipeline and the centralized baselines.
MosaicML NLP Team has released MPT (MosaicML Pretrained Transformer), a commercially usable, open-source combination of a language model and training pipeline to bundle the most popular software infrastructure and optimizations described in \cref{subsec:centralized_distributed_optimization}.
The standard MPT model~\cite{mpt_blogpost} is a decoder-style transformer with \num{6.7}B parameters, integrating Flash Attention~\citep{flashattention} for efficient execution, ALiBi~\citep{alibi} for context length extrapolation and general stability improvements to mitigate loss spikes.
In this work, we use MPT models of different scales and follow the best practices recommended by MosaicML for the \emph{local} training pipeline unless stated otherwise.
We seed every local training and the client selection mechanism to ensure experimental reproducibility.

\subsection{Experimental Tools}

Unlike other FL applications, the incredible computational resources and time necessary for training LLMs require constant monitoring, checkpoints, and potential manual interventions to restart training from an appropriate point.
Thus, we use the MosaicML tools to track running statistics on the norms of optimizer states, gradient updates, and activations.
These generally serve as leading indicators of model divergence in centralized settings~\citep{meta_opt}.
Our extensions allow the monitors to track per-model and per-layer statistics efficiently, a necessary addition given the added complexity of FL and the need for client-level monitoring.
Besides this, we allow for the easy integration of federated metrics that cannot be captured locally, such as the pairwise distance or cosine similarity between client models, client pseudo-gradient norms, and server-side momentum norms.
Furthermore, we ensure that all the resources controlling the local pipeline are correctly cleaned up or reused as appropriate to avoid interference between different rounds or concurrent experiments.

To enable proper experimental configuration, tracking, and resumption, we provide a set of quality-of-life improvements, such as automatic federated training resumption from the most recent round and typed experimental schemas for all federated hyperparameters.
Crucially, since federated execution may occur on various devices, we provide an efficient procedure for automatically choosing a micro-batch size appropriate for your GPU type.
The procedure produces an estimate for the micro-batch size based on the model's memory consumption and a micro-batch of $1$ by finding the power of $2$ that most closely approaches the limits of the VRAM.
It then iteratively improves upon this guess by binary searching over powers of two for the largest batch size, which does not cause an out-of-memory condition (OOM).
We only search over powers of $2$ as they are most likely to match the assumptions of both underlying algorithms and the hardware configuration of most machines, which have $1,2,4$ or $8$ GPUs.

\subsubsection{Data Partitioning}\label{subsec:data_partitioning}

To model scenarios where clients train on heterogeneous data, we provide scripts for partitioning \emph{The Pile}~\citep{ThePile} and Multilingual Cleaned Colossal Common Crawl~(\emph{mC4})~\citep{mC4} into \textbf{naturally} heterogeneous subsets, unlike previous work~\citep{DiLoCo} which relied on artificially partitioned texts within a dataset using the embeddings of a pre-trained model.
While naturally heterogeneous partitions for LLM training have been previously considered by \citet{DatasetGrouper}, their work focused on efficiently building fine-grained partitions for cross-device FL scenarios with potentially millions of clients based on individual URLs.
Given our cross-silo setting, we are far more concerned with the degrees of heterogeneity likely to be exhibited across medium-scale and large organizations.
Thus, we focus on building large and semantically meaningful heterogeneous partitions.

For example, the heterogeneous partitioning strategies we have thus far mentioned and constructed correspond to texts that differ by genre/originating website in the case of \emph{The Pile}, simulating collaborations between publishers from different domains, and texts that vary by language in the case of \emph{mC4}, simulating transnational cooperation. 

We first partition each heterogeneous dataset by category~(e.g., source dataset for \emph{The Pile} and language for \emph{mC4}) and then split each individual category into disjoint buckets.
The number of buckets we create per category equals $J \times |C|$, where $|C|$ is the number of distinct clients in the federation, while $J$ is the maximum number of categories a client may draw upon.
For example, for a federation of $8$ clients, each of which may be a combination of $3$ categories, we would create $24$ buckets per category.
Each bucket can be mapped to at most one client in the federation, ensuring that even if two clients draw from the same source, they constantly sample from disjoint data subsets.
This implementation allows us to build any arbitrarily complex topology without additional bookkeeping or runtime costs.

\subsection{Experimental Design}\label{subsec:experimental_design}

To evaluate \photon, we consider several aspects of federated LLM optimization corresponding to what we perceive as the biggest challenges to adopting federated LLM pre-training.

\paragraph{Homogeneous Data Sources:} We wish to isolate the impact of the optimization procedure itself by using a standard benchmark dataset partitioned across clients in an IID fashion.
Since parameter averaging is known to inject noise into the optimization procedure~\citep{DontUseLargeBatchesUseLocalSGD}, it may cause a change in the dynamics of LLM training.
Thus, we want to observe whether it can substitute distributed data-parallel SGD for situations where data is mainly similar from client to client~(e.g., in the same language and containing the same mix of genres). We use a version of C4~\citep{C4} randomly split into randomly constructed shards for our homogeneous data sources experiments.

\paragraph{Heterogeneous Data Sources:} We want to observe whether injecting \emph{federated statistical heterogeneity} into the system harms convergence.
We are particularly concerned with approximating a realistic cross-silo form of data heterogeneity similar to those encountered by institutions like publishers, which may specialize in different text genres.
For this purpose, we choose to use \emph{The Pile} partitioned into subsets representing general knowledge~(\emph{Wikipedia, English only}), general scientific publications~(\emph{ArXiv}), long-form books ~(\emph{Project Gutenberg, alias PG-19}), technical news/discussions~(\emph{HackerNews}), medical publications~(\emph{PubMed Central}), law~(\emph{FreeLaw}), philosophical publications~(\emph{PhilPapers}), and Q/A websites~(\emph{StackExchange}).
This coarse mapping allows us to directly determine the efficacy of federated training in reconciling different styles and sources of knowledge.

\paragraph{Partial Participation:} One aspect of federated training with broad implications for the computational costs and data efficiency of the method is partial participation.
As previously discussed in \cref{subssec:communication_computation_scalability}, while in typical~(i.e., cross-device) FL scenarios aggregating across a greater number of clients helps improve convergence by reducing the variance of client pseudo-gradients, this effect tends to saturate.
Thus, we aim to discover if similar behaviors exist in federated settings.

\subsection{Model Scaling}

We want to show that federated optimization can train a large model without access to the sprawling data centers traditionally tasked with such training.
Given the prohibitive nature of such training, we initially focus on the IID $C4$ partition, as it represents a generally agreed-upon benchmark in the community. 

We inspire from the scaling laws of \citet{TrainingComputeOptimalLLMs} to reason about what could be an equipollent discussion for federated training.
In particular, we are interested in making consistent choices regarding the number of tokens the models train on.
In centralized settings, the estimate for an optimal ratio, $\frac{\mathrm{dim}(\mathcal{D})}{\mathrm{dim}(\Theta)}$, is 20 tokens/parameter, with $\mathrm{dim}(\mathcal{D})$ being the number of tokens of the dataset $\mathcal{D}$ and $\mathrm{dim}(\Theta)$ being the number of trainable parameters for the model $\Theta$.
However, we must consider that \citet{TrainingComputeOptimalLLMs} used a tokenizer with about $32$K tokens in its vocabulary, while we intend to use $50$K.
This different tokenization impacts the size of the embedding layer only.
If we assume that the scaling laws of \citet{TrainingComputeOptimalLLMs} will hold even with a larger tokenizer, we can scale down the size of the model we train to obtain their \citet{TrainingComputeOptimalLLMs} equivalents.

While the scaling laws proposed by \citet{TrainingComputeOptimalLLMs} provide a compute-optimal way to achieve a specific loss, they do not account for inference costs after pre-training. \citet{BeyondChinchilaOptimal}, affiliated with MosaicML, proposes a longer pre-training phase to improve inference performance, which .
Very recently, \citet{he2024exploringscalinglawslocal} investigated the particular case of LocalSGD in training LLMs.
Despite that work being admittedly limited, we hope that future research in this direction will clarify optimal choices for training LLMs with this approach. Our estimates are presented in \cref{tab:scaling_laws}.

\begin{table}[ht]
\caption{Pre-training tokens and steps for a given model size $\mathrm{dim}(\theta)$ in parameters, the size in the parentheses represents the vocabulary-adjusted size to match \citet{TrainingComputeOptimalLLMs}. Where $\mathrm{dim}(\mathcal{D})|_{\Theta}$ is the compute-optimal number of tokens recommended by \citet{TrainingComputeOptimalLLMs}, $\mathrm{dim}(\mathcal{D})_{\mathrm{MPT}}|_{\Theta}$ is the number recommended by MosaicML. We use $\mathrm{dim}(\mathcal{D})_{\mathrm{SEQ} | \theta}^{*}$ as the number of sequential tokens and $\mathrm{dim}(\mathcal{D})_{\mathrm{PAR} | \theta}^*$ as the number of parallel ones, which depends on the number of clients. Given the sequence length and batch size $l$, $B$, we report the equivalent number of steps $\mathcal{T}$. }
\label{tab:scaling_laws}
\resizebox{\textwidth}{!}{%
\begin{tabular}{@{}cccccccccc@{}}
\toprule
$\mathrm{dim}(\Theta)$ & $\mathrm{dim}(\mathcal{D})|_{\Theta}$ & $\mathrm{dim}(\mathcal{D})_{\mathrm{MPT}}|_{\Theta}$ & $\mathrm{dim}(\mathcal{D})_{\mathrm{SEQ} | \theta}^{*}$ & $\mathrm{dim}(\mathcal{D})_{\mathrm{PAR} | \theta}^*$ & $l$ & $B$ & $\mathcal{T}_{\mathrm{dim}(\mathcal{D})|\Theta}$ & $\mathcal{T}_{\mathrm{dim}(\mathcal{D})_{\mathrm{MPT}}|_{\Theta}}$ & $\mathcal{T}_{\mathrm{dim}(\mathcal{D})_{\mathrm{SEQ} | \theta}^*}$ \\ \midrule
\rowcolor{lightgray} $75\mathrm{M} \, (58.54\mathrm{M})$ & $1.17 \times 10^9$ & $-$ & $5.2 \times 10^9$ & $41.9 \times 10^9$ & $1024$ & $256$ & $4463$ & $-$ & $88000$ \\
$125\mathrm{M} \, (110.89\mathrm{M})$ & $2.22 \times 10^9$ & $2.5 \times 10^9 $ & $6.6 \times 10^9$ & $52.4 \times 10^9$ & $2048$ & $256$ & $4235$ & $4800$ & $15000$ \\
\rowcolor{lightgray} $350\mathrm{M} \, (331.19\mathrm{M})$ & $6.62 \times 10^9$ & $8.0 \times 10^9$ & $10.5 \times 10^9$ & $83.9 \times 10^9$ & $2048$ & $256$ & $12627$ & $15360$ & $13500$ \\
$1.3\mathrm{B} \, (1.26\mathrm{B})$ & $25.2 \times 10^9$ & $26.0 \times 10^9$ & $7.35 \times 10^9$ & $58.8 \times 10^9$ & $2048$ & $512$ & $24033$ & $24800$ & $7000$ \\
\rowcolor{lightgray} $3\mathrm{B} \, (2.96\mathrm{B})$ & $59.2 \times 10^9$ & $54.0 \times 10^9$ & $13.1 \times 10^9$ & $52.4 \times 10^9$ & $2048$ & $512$ & $56458$ & $51500$ & $10500$ \\
$7\mathrm{B} \, (6.92\mathrm{B})$ & $138 \times 10^9$ & $134.0 \times 10^9 $ & $22.0 \times 10^9$ & $88.1 \times 10^9$ & $2048$ & $1024$ & $65804$ & $63900$ & $10500$ \\ \bottomrule
\end{tabular}%
}
\end{table}

\begin{table}[ht]
\centering
\caption{Architecture details and local training parameters for our $75$M, $125$M, $350$M, $1.3$B, $3$B, and $7$B models. We report the number of transformer blocks, hidden model dimension $d$, number of attention heads, the linear layer expansion ratio, and Adam's parameters ($\beta_1$ and $\beta_2$). We also report the vocabulary size of the tokenizer we used~\citep{eleuther_ai_tokenizer} and the sequence length $l$.}
\label{tab:model_sizes}
\begin{tabular}{@{}rccccccc@{}}
\toprule
    \textbf{Model} &
    \multirow{2}{*}{\textbf{\#Blocks}} &
    \multirow{2}{*}{$\boldsymbol{d}$} &
    \multirow{2}{*}{\textbf{\#Heads}} &
    \multirow{2}{*}{\textbf{Exp.~Ratio}} &
    \multirow{2}{*}{$\mathbf{(\boldsymbol{\beta_1},~\boldsymbol{\beta_2})}$} &
    \multirow{2}{*}{$\boldsymbol{|\textbf{Vocab}|}$} &
    \multirow{2}{*}{$\boldsymbol{l}$} \\
    \textbf{Size} &
    &
    &
    &
    &
    &
    &
    \\
\midrule
    \textbf{$\mathbf{75}$M} & 3 & 896 & 16 & 4  & $(0.9,~0.95)$ & \num{50368} & \num{1024} \\
    \textbf{$\mathbf{125}$M} & 12 & 768 & 12 & 4  & $(0.9,~0.95)$ & \num{50368} & \num{2048} \\
    \textbf{$\mathbf{350}$M} & 24 & 1024 & 16 & 4  & $(0.9,~0.95)$ & \num{50368} & \num{2048} \\
    \textbf{$\mathbf{1.3}$B} & 24 & 2048 & 16 & 4  & $(0.9,~0.95)$ & \num{50368} & \num{2048} \\
\midrule
    \textbf{$\mathbf{3}$B} & 32 & 2560 & 20 & 4  & $(0.9,~0.95)$ & \num{50368} & \num{2048} \\
    \textbf{$\mathbf{7}$B} & 32 & 4096 & 32 & 4  & $(0.9,~0.95)$ & \num{50368} & \num{2048} \\
\bottomrule
\end{tabular}
\end{table}

\subsection{Training Setup}\label{subsec:training_setup}

All experiments use \num{500} local steps per round, executed by all the clients in the federation.

Our setting respects the conditions of standard cross-silo~\citep{AdancesAndOpenProblems} FL in which the orchestrator samples the entire batch of equally capable clients in every round.
Our experiments used diverse hardware resources whose specifications allow for reasonably fast execution; as an example, we used a combination of heterogeneous servers equipped with NVIDIA A40, A100, and H100 GPUs.
Due to FL's lax synchronization requirements, these heterogeneous hardware accelerators were able to collaborate despite being located in different countries.

As shown in Table \ref{tab:model_sizes}, we trained models ranging in size from \num{75} million parameters to \num{7} billion for the causal language modeling task.
We used the tokenizer presented in \cite{eleuther_ai_tokenizer} with a vocabulary size of \num{50368}.
The local optimizer the clients use in our experiments is AdamW \citep{AdamW}, while the server optimizer is FedMom \citep{FedMOM}.
The hyperparameters we used are reported in Table \ref{tab:fl_hyperaparams}.

\begin{table}[ht]
\centering
\caption{Hyperparameters used in our experiments. The federated learning rate $\boldsymbol{\eta_s}$ and momentum $\boldsymbol{\mu_s}$~\citep{FedMOM} are applied by \photonserver. The FL is performed across the number of rounds reported. $\mathbf{S_C}$ are the parameters of the learning rate scheduler synchronized across \textbf{sequential} steps. $\alpha$ is the factor to be applied to the maximum learning rate $\eta_{max}$ to obtain the minimum learning rate for the cosine scheduler, i.e., $\eta_{min}=\alpha\times\eta_{max}$. $T$ is the duration, in steps, of the cosine scheduler. We also report the batch size used in the local training by the \photonclients.}
\label{tab:fl_hyperaparams}
\begin{tabular}{@{}rcccccc@{}}
\toprule
    \textbf{Model} &
    \multirow{2}{*}{$\boldsymbol{\eta_s}$} &
    \multirow{2}{*}{$\boldsymbol{\mu_s}$} &
    \multirow{2}{*}{$\boldsymbol{\alpha}$} &
    \multirow{2}{*}{$\boldsymbol{\eta_{max}}$} &
    \multirow{2}{*}{$\boldsymbol{T}$} &
    \textbf{Batch}  \\
    \textbf{Size} &
     &
     &
     &
     &
     &
    {\textbf{Size}}  \\
\midrule
    \textbf{$\mathbf{75}$M} & $0.7$ & $0.9$ & ${10^{-1},10^{-6}}$ & $4\times10^{-4}$ & \num{88000} & \num{256} \\
    \textbf{$\mathbf{125}$M} & $0.3, 0.5, 0.7$ & $0.9$ & ${10^{-1}, 10^{-5}}$ & $\{3.0, 6.0\}\times10^{-4}$ & \num{15000} & \num{256} \\
    \textbf{$\mathbf{350}$M} & $0.1$ & $0.9$ & $10^{-1}$ & $3\times10^{-4}$ & \num{13400} & \num{256} \\
    \textbf{$\mathbf{1.3}$B} & $0.7$ & $0.9$ & $10^{-1}$ & $2\times10^{-4}$ & \num{24800} & \num{512} \\
\midrule
    \textbf{$\mathbf{3}$B} & $0.7$ & $0.9$ & $10^{-1}$ & $1.6\times10^{-4}$ & \num{51500} & \num{512} \\
    \textbf{$\mathbf{7}$B} & $0.7$ & $0.9$ & $10^{-1}$ & $1.2\times10^{-4}$ & \num{63900} & \num{1024} \\
\bottomrule
\end{tabular}
\end{table}

\begin{table}[ht]
\centering
\caption{Hyperparameters for our federated experiments. $P$ represents the total number of clients per federations, $K$ the number of clients sampled per round, $D$ the dataset, $\tau$ the number of steps per round.}
\label{tab:fl_settings}
\begin{tabular}{@{}rccccc@{}}
\toprule
    \textbf{Model} &
    \multirow{2}{*}{\textbf{\#Rounds}} &
    \multirow{2}{*}{$\boldsymbol{P}$} &  
    \multirow{2}{*}{$\boldsymbol{K}$} &  
    \multirow{2}{*}{$\boldsymbol{D}$} &  
    \multirow{2}{*}{$\boldsymbol{\tau}$} \\  
    \textbf{Size} &
     &
     &
     &
     &
     \\
\midrule
    \textbf{$\mathbf{75}$M} & $40$ & ${8, 64}$ & ${8, 4}$ & C4~\citep{C4}, The Pile~\citep{ThePile} & $500$ \\
    \textbf{$\mathbf{125}$M} & ${10, 25}$ & ${8, 64}$ & ${8, 4}$ & C4~\citep{C4}, The Pile~\citep{ThePile} & ${250, 500}$ \\
    \textbf{$\mathbf{350}$M} & $40$ & $8$ & $8$ & C4~\citep{C4} & $500$ \\
    \textbf{$\mathbf{1.3}$B} & $14$ & $8$ & $8$ & C4~\citep{C4} & $500$ \\
\midrule
    \textbf{$\mathbf{3}$B} & $21$ & $64$ & $4$ & C4~\citep{C4} & $500$ \\
    \textbf{$\mathbf{7}$B} & $21$ & $64$ & $4$ & C4~\citep{C4} & $500$ \\
\bottomrule
\end{tabular}
\end{table}
\section{Evaluation}

We have experimentally validated the efficacy of our federated training recipe at a series of increasing model scales while investigating impactful choices for both the federated and local optimization. Thus, the results presented in this section indicate the viability of federated optimization for LLM pre-training and validate its ability to reach beyond the data and hardware constraints of centralized approaches. We push our evaluation from relatively small sizes all the way to $7$B parameters in \cref{sec:pushing_boundaries}.

\begin{figure}[ht]
    \centering
    \subfloat[]{\includegraphics[width=0.45\textwidth]{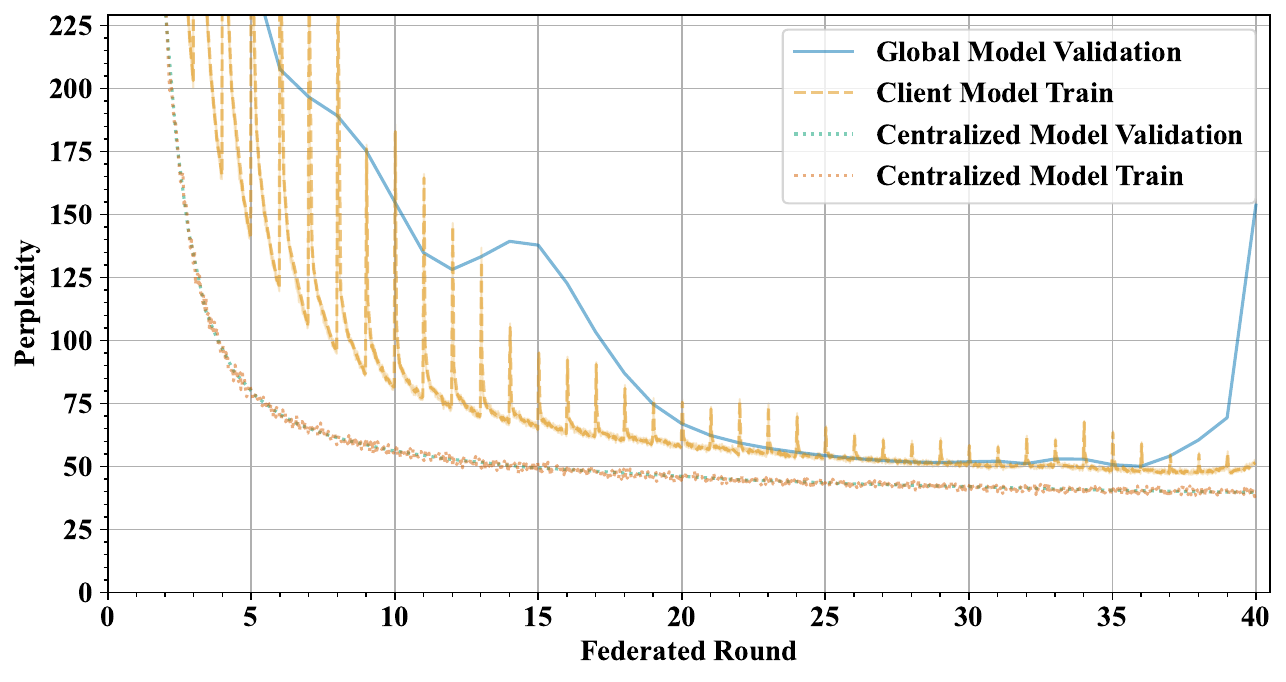}} 
    \subfloat[]{\includegraphics[width=0.45\textwidth]{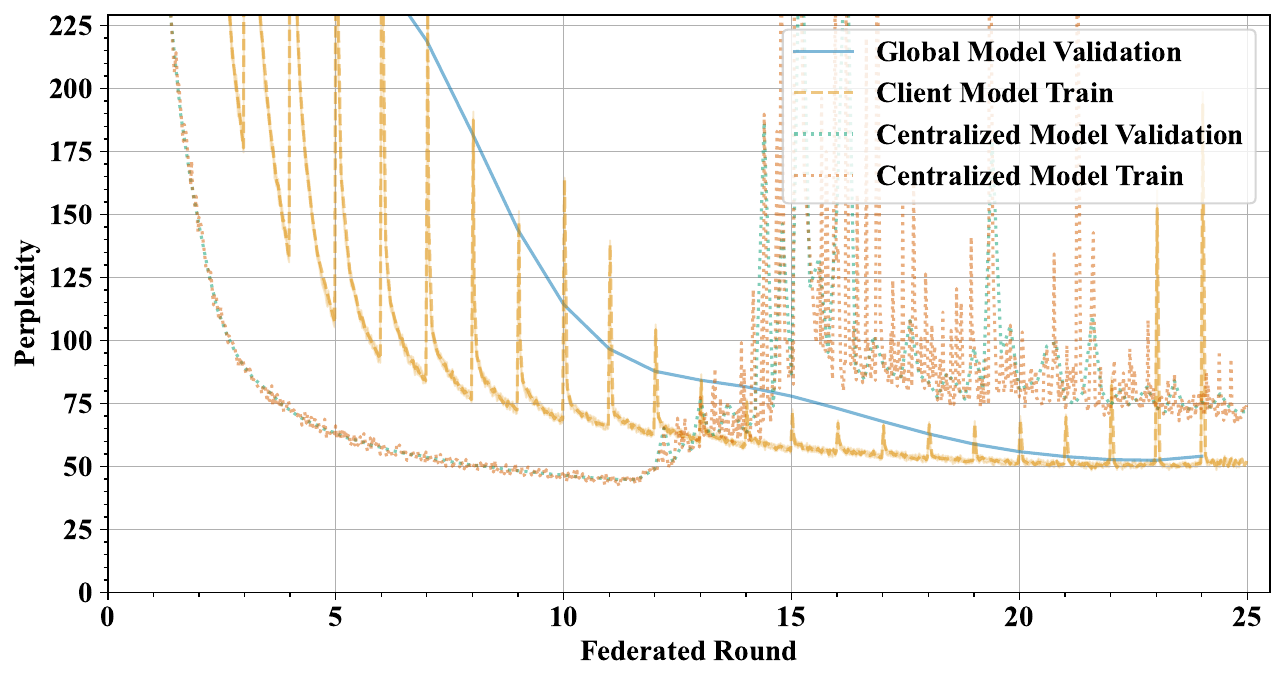}}\\
    \subfloat[]{\includegraphics[width=0.45\textwidth]{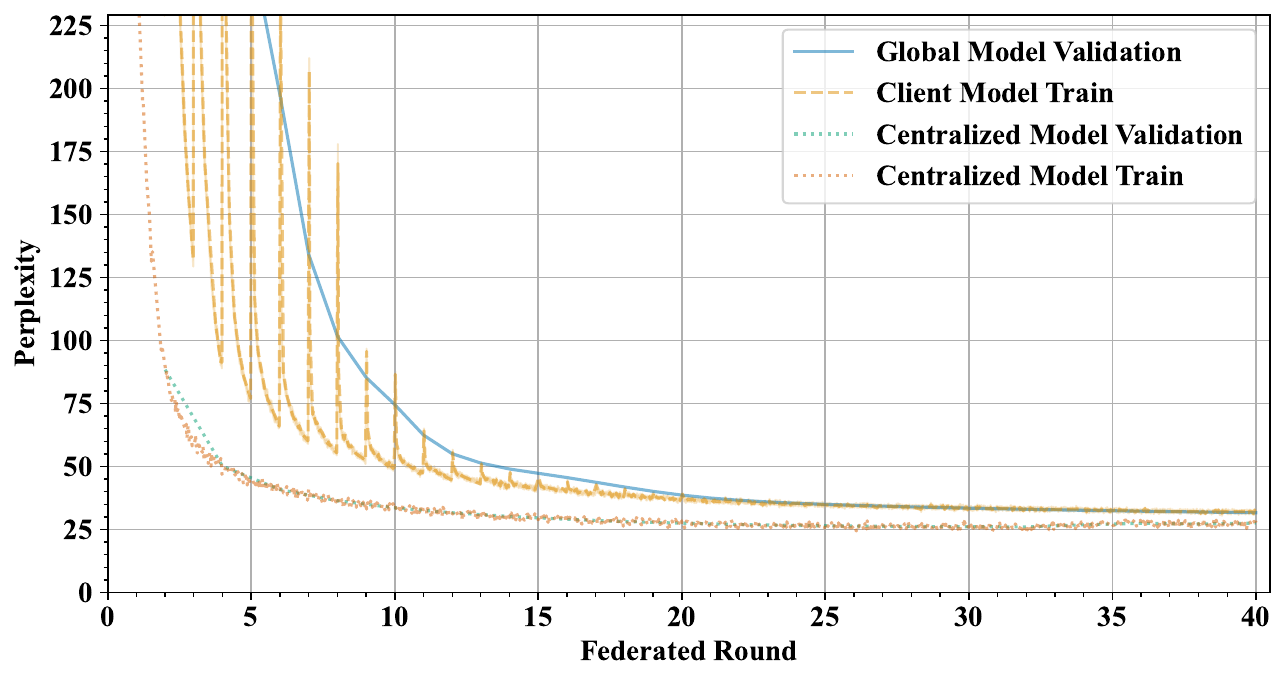}}
    \subfloat[]{\includegraphics[width=0.45\textwidth]{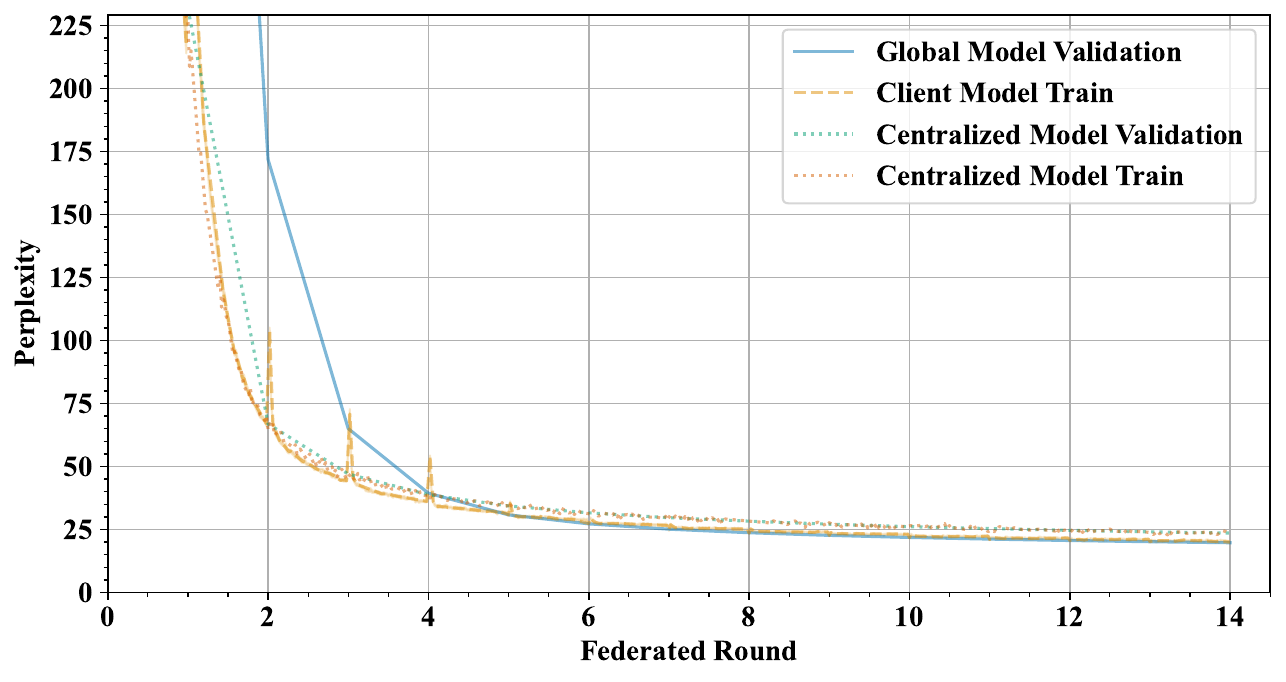}} 
    \caption{Comparison between the perplexity of the federated global model evaluated on the centralized validation set, the train perplexities of federated clients (averaged together), and the train and test perplexities for a centralized experiment.
    These metrics are reported for our $75$M~(a), $125$M~(b), $350$M~(c), and $1.3$B~(d) experiments. Crucially, the stability of federated training increases with model size. For example, the centralized model outperforms the $75$M federated model while performing near-identically for the $1.3$B models. While federated aggregation initially causes large spikes in client perplexity, these subside as the clients reach a consensus on the model parameters, which happens much quicker for larger models. Following this transitory phase, aggregation applies a regularizing effect on the model performance, allowing a better model to be trained than would be possible for a single client. The server validation perplexity is a soft upper bound for the spikes, with the gap between train and validation perplexities decreasing over time. }
    \label{fig:fed:perplexity-(generic-scale)}
\end{figure} 

\subsection{Federated Optimization is Effective and Stable across Scales:}

Our initial experiments using the IID partition of the \emph{C4} dataset reveal that federated optimization can reach a performance comparable to centralized training.
As can be seen in \cref{fig:fed:perplexity-(generic-scale)}, federated models perform comparatively well to centrally trained ones or even better at sufficiently large model sizes.
Furthermore, while their convergence curves may initially look more unstable, the federated aggregation procedure quickly resolves this once the model settles into an appropriate compromise for the clients.
Most importantly, \textbf{FL effectively trains the $\mathbf{1.3}$B parameters model}.
The gap between centralized and federated performance becomes increasingly small as the model size increases, ranging between $10$ and $2$ points on the perplexity axis. In \cref{fig:fed:perplexity_3b_and_7b}, we extend our experiments to show that this trend not only continues for $3$B and $7$B but accelerates.

As such, we believe that federated optimization can effectively replace standard data-parallel techniques in data center settings when nodes are poorly connected, which \photon handles automatically.
Furthermore, we will now argue that it is suitable for in-the-wild federated systems where data may also come from statistically heterogeneous distributions.

\subsection{Federated Optimization is Robust to Heterogeneity:}

As previously discussed, \emph{federated statistical heterogeneity} problem is inevitable for real federated learning scenarios as clients use naturally generated private data to train their models.
Thus, we investigate how much heterogeneity impacts performance using our federated partition of \emph{The Pile} composed of \emph{Wikipedia (en)}, \emph{ArXiv}, \emph{Project Gutenberg (PG-19)}, \emph{HackerNews}, \emph{PubMed Central}, \emph{FreeLaw}, \emph{PhilPapers}, and \emph{StackExchange}.
We conduct these experiments on our smaller $75$M and $125$M models.
We chose them because they are under-parameterized compared to our $1.3$B model, making reconciling heterogeneous data particularly challenging.
In \cref{tab:fl_hyperaparams}, we report the hyperparameters we use for our experiments.

\begin{figure}[ht]
    \centering
    \subfloat[]{\includegraphics[width=0.45\textwidth]{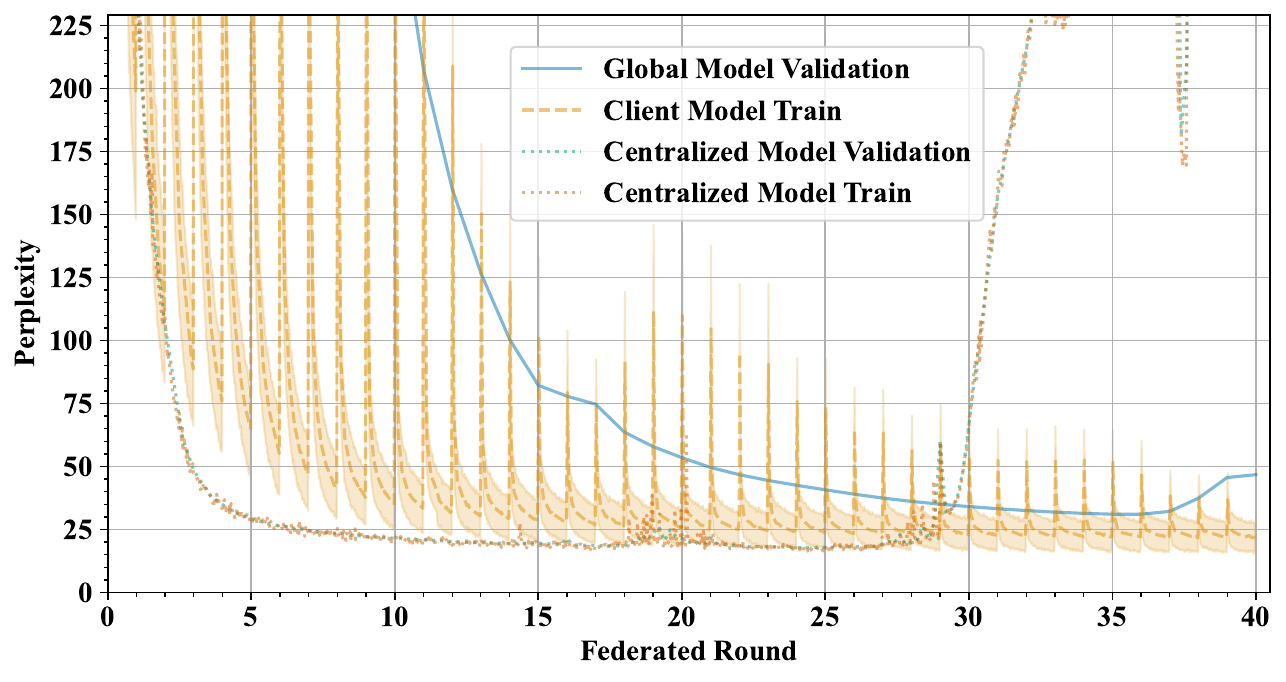}}
    \subfloat[]{\includegraphics[width=0.45\textwidth]{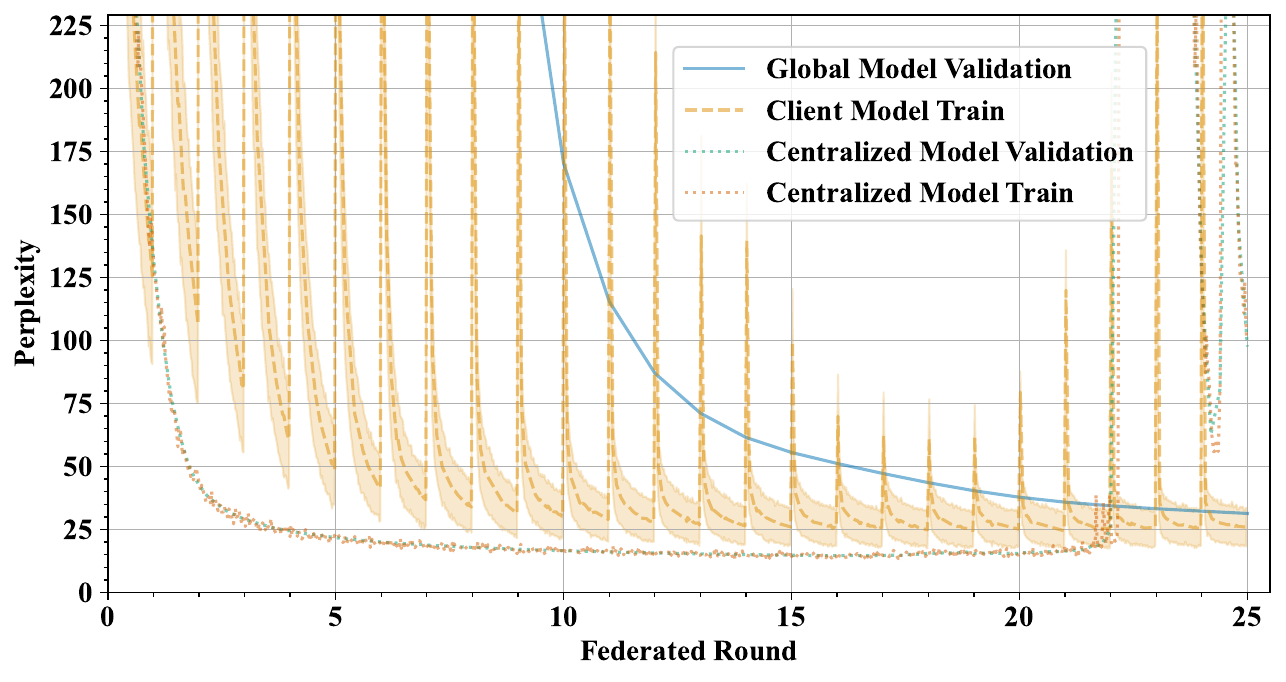}} 
    \caption{Perplexity comparison between the global model evaluated on the centralized validation set, the train and test perplexities for a centralized baseline, and the training perplexities of clients (averaged together) for our naturally heterogeneous partition of \emph{The Pile} using either a $75$M~(a) or $125$M~(b) model size. Unlike the homogeneous partition, the natural heterogeneity of the underlying datasets makes an initial consensus harder to reach for the federated model, as can be observed from the very high initial clients and server perplexities. However, like the IID partition shown in \cref{fig:fed:perplexity-(generic-scale)}, once clients reach consensus, performance becomes comparable to a centralized baseline.}
    \label{fig:fed:perplexity-pile}
\end{figure}

We observe a high degree of robustness towards this naturally heterogeneous partition.
As shown in \cref{fig:fed:perplexity-pile}, while clients show much more significant variance in their loss, the server validation loss converges with a similar trend to the IID \emph{C4} case.
Compared to the previous set of experiments, the absolute performance improves from 50 to 35 for the $75$M model and from 50 to 30 for the $125$M model mainly because of the different datasets used.
We believe that the highly effective convergence is due to the regularization properties of FL, precisely because of FL's meta-learning properties~\citep{REPTILE,PflModelAgnosticMetaLearning, lee2024fedl2p}.

\begin{figure}[ht]
    \centering
    \subfloat[]{\includegraphics[width=0.45\textwidth]{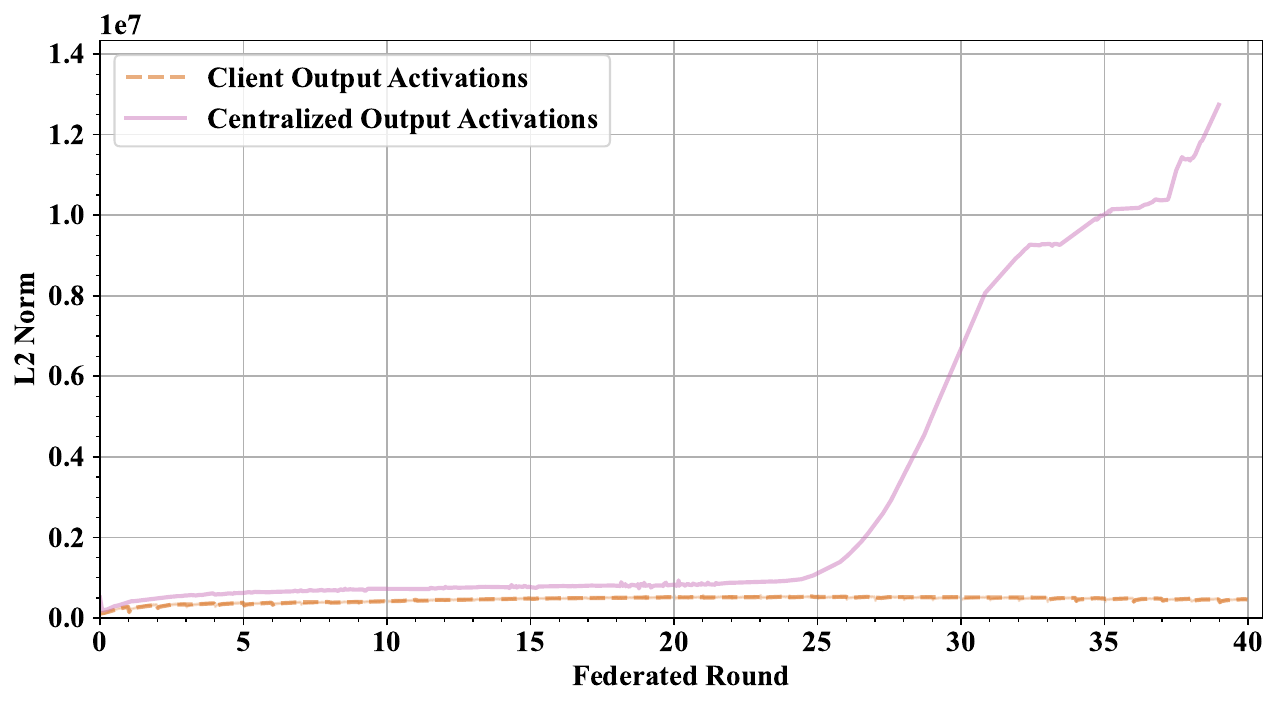}}
    \subfloat[]{\includegraphics[width=0.45\textwidth]{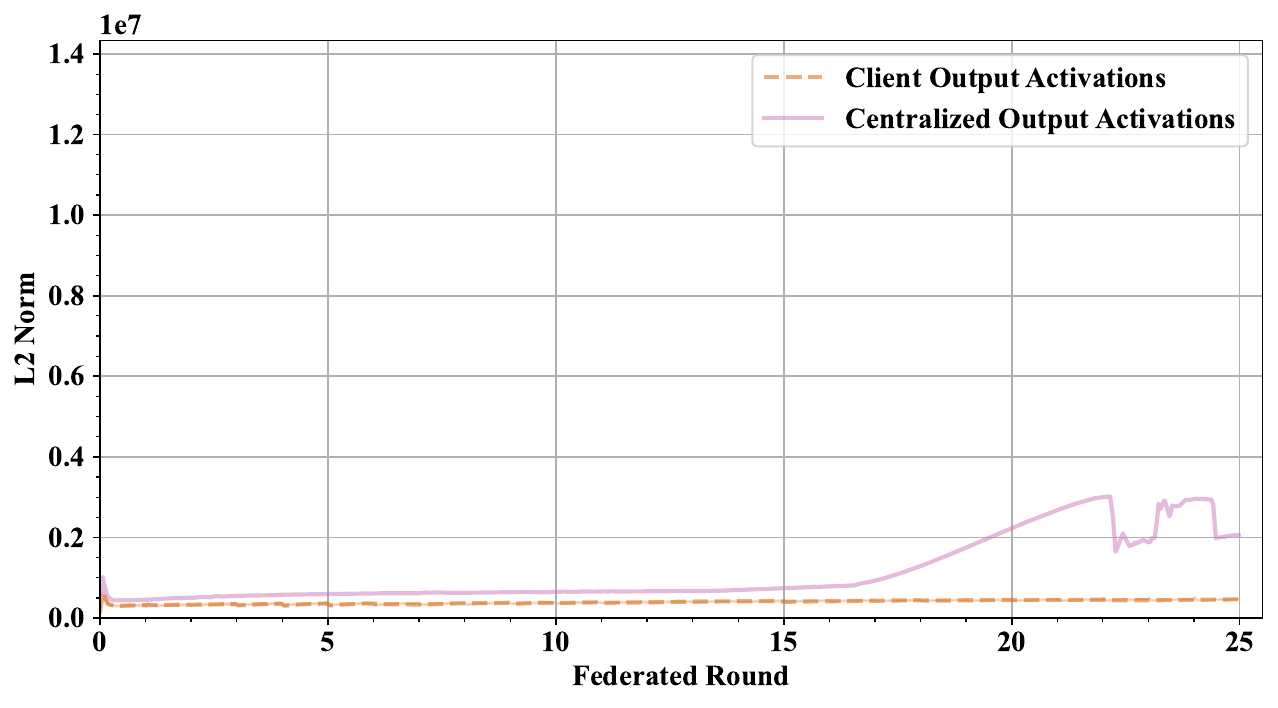}} 
    \caption{The $l_2$ norms of the output activations of our $75$M~(a) and $125$M ~(b) models trained on a naturally heterogeneous partition of \emph{The Pile}. The norm of the activations is a well-known indicator of future model divergence~\citep{meta_opt}. As can be observed, the activations of the centralized model outpace those of the federated clients right from the start and experience a massive increase towards the end of training. The aggregation procedure keeps the federated clients in check by reducing the norm of the activations round to round.}
    \label{fig:fed:act-pile}
\end{figure}

Furthermore, as can be seen in \cref{fig:fed:act-pile}, models trained in a federated fashion are not subject to the same tendency of rapidly increasing activations as centralized models are. Corroborating this against the perplexity in \cref{fig:fed:perplexity-pile} shows that while the centralized model may diverge beyond repair, the federated model can quickly recover from temporary decreases in server validation accuracy. We argue that this is not a mere artifact of the model being less trained since \cref{fig:fed:act-pile} show that activations of the centralized model start higher than those of the federated one and remain so throughout the entire duration of the experiments. We attribute this to the noise injected by the aggregation procedure, given that we observe periodic reductions in activation norms at round boundaries. This confirms the robustness properties of federated aggregation~\citep{DontUseLargeBatchesUseLocalSGD}.

\subsection{Larger Models Improve Consensus:}

Our experiments focused on scaling the model size to determine if federated optimization can train models approximating the size of those commonly used today~\citep{llama,llama2}. 

We observed that the stability of the training procedure improves directly with the model size, with federated optimization reaching a better consensus across client models.
As shown in \cref{fig:fed:perplexity-(generic-scale)}, while the centralized model's final performance is slightly greater for small model sizes, it becomes almost identical for the $1.3$B model.
Furthermore, client training perplexities showcase a shift from a transient oscillatory phase to a convergence phase where federated optimization acts as a regularizer, improving performance.
Thus, while aggregation increases local perplexity in the early stages, it decreases it at convergence.
This transition is much quicker for larger models, with the $1.3$B model spending only $4$ rounds in the pre-convergence phase, while the $75$M model requires over $20$ rounds to reach a consensus across client models.
Similarly, while the $75$M model continues to experience occasional turbulences in convergence after round $20$, the larger models, such as the $350$M model, obtain \textbf{full} client model alignment.

\subsection{Federated Optimization is Robust to Partial Participation:}

\begin{figure}[ht]
    \centering
    \subfloat[]{\includegraphics[width=0.45\textwidth]{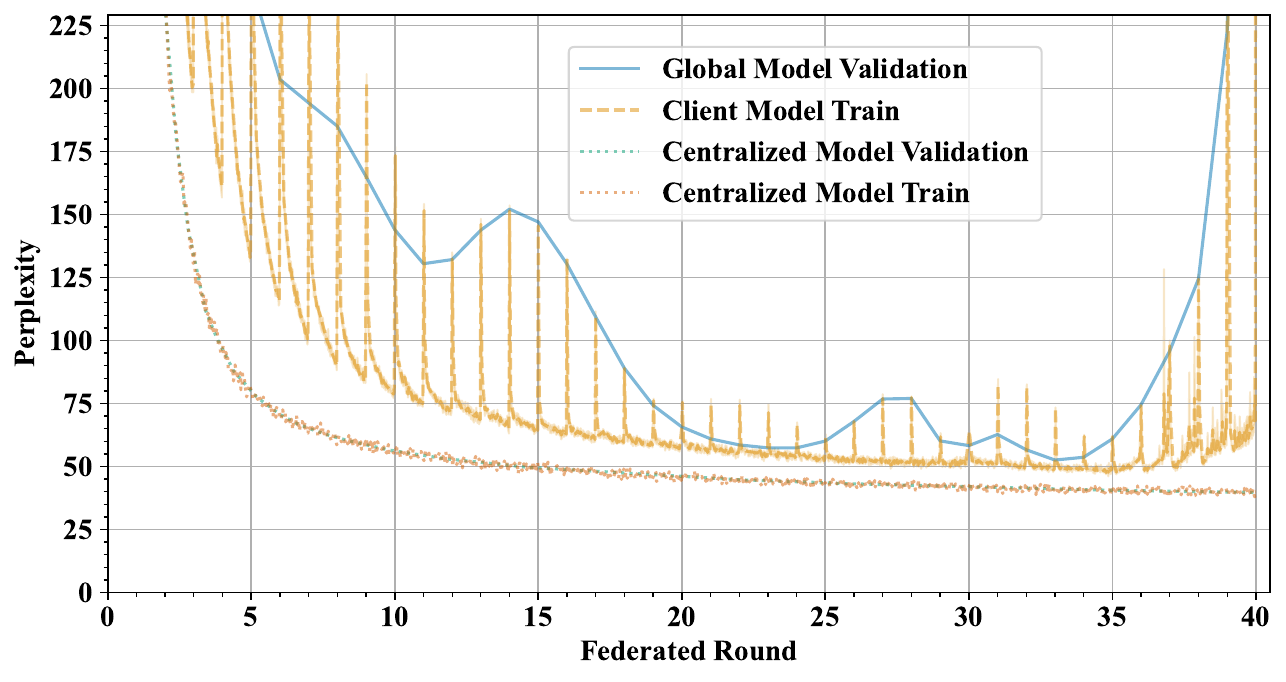}}
    \subfloat[]{\includegraphics[width=0.45\textwidth]{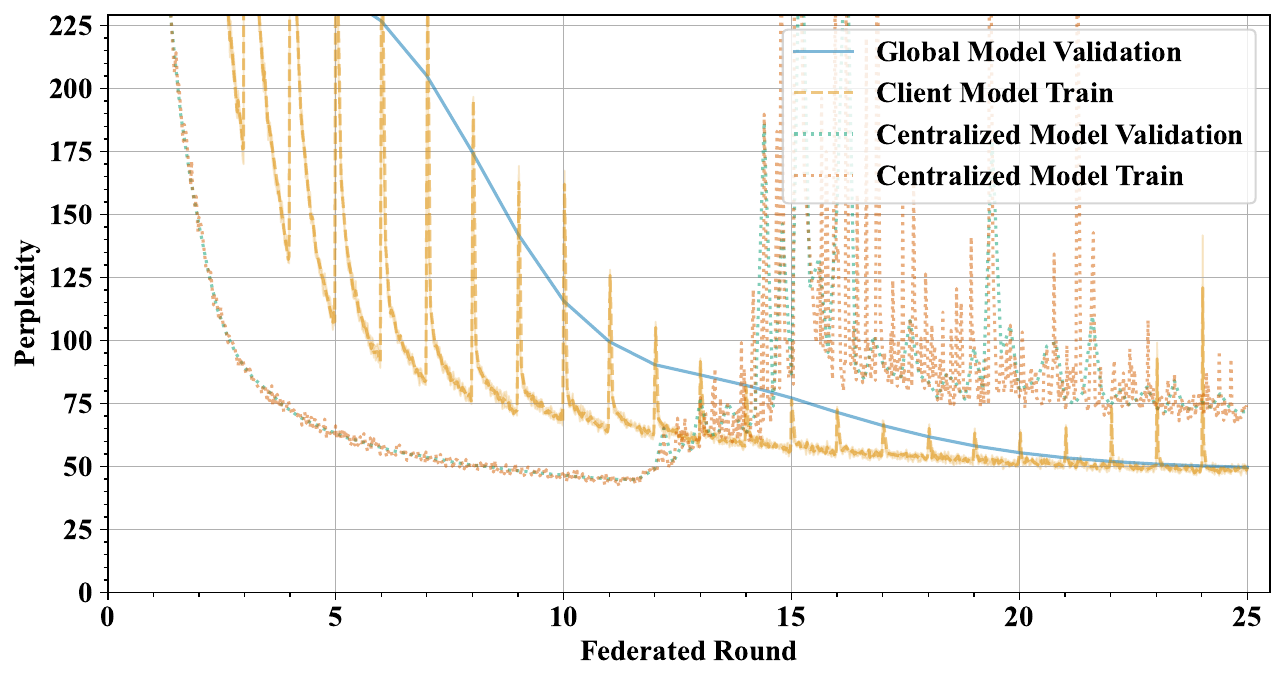}} 
    \caption{Perplexity comparison, $75$M~(a) and $125$M~(b), between the server model evaluated on the centralized validation set, the train and test perplexities for a centralized baseline, and the training perplexities of clients for our $64$-client partition of \emph{C4} with $4$ client participating per round for a $6.25\%$ participation rate. Despite using half of the parallel compute of the full participation baselines, partial participation converges to the same performance at the expense of more turbulences early in the training process. Thus, subsampling cohorts can substantially reduce hardware utilization for a sufficiently large cross-silo population, allowing for parallel training of several models.}
    \label{fig:fed:perplexity-partial}
\end{figure}

Our experiments have shown that meaningfully-sized LLMs can be trained in a federated fashion when using full client participation.
However, one particularly effective way of reducing overall training costs in terms of parallel computing while still having the regularization benefits of training on more data is to subsample a smaller portion of the entire federated population.
The experiments in \cref{fig:fed:perplexity-partial} show how subsampling only $6.25\%$ of clients from our $64$-client partition results in the same performance as our previous full-participation experiments, despite training only $4$ clients per round instead of $8$.
Such robustness to partial participation opens the door to highly efficient LLM training schedules in which the scalable addition of data, as needed, improves convergence whilst minimizing computing resource utilization.
Furthermore, it enables applications where multiple federated workloads may be executed simultaneously by sampling different clients from the population.

\subsection{Interplay of Client and Server Models:}

\begin{figure}[ht]
    \centering
    \subfloat[]{\includegraphics[width=0.45\textwidth]{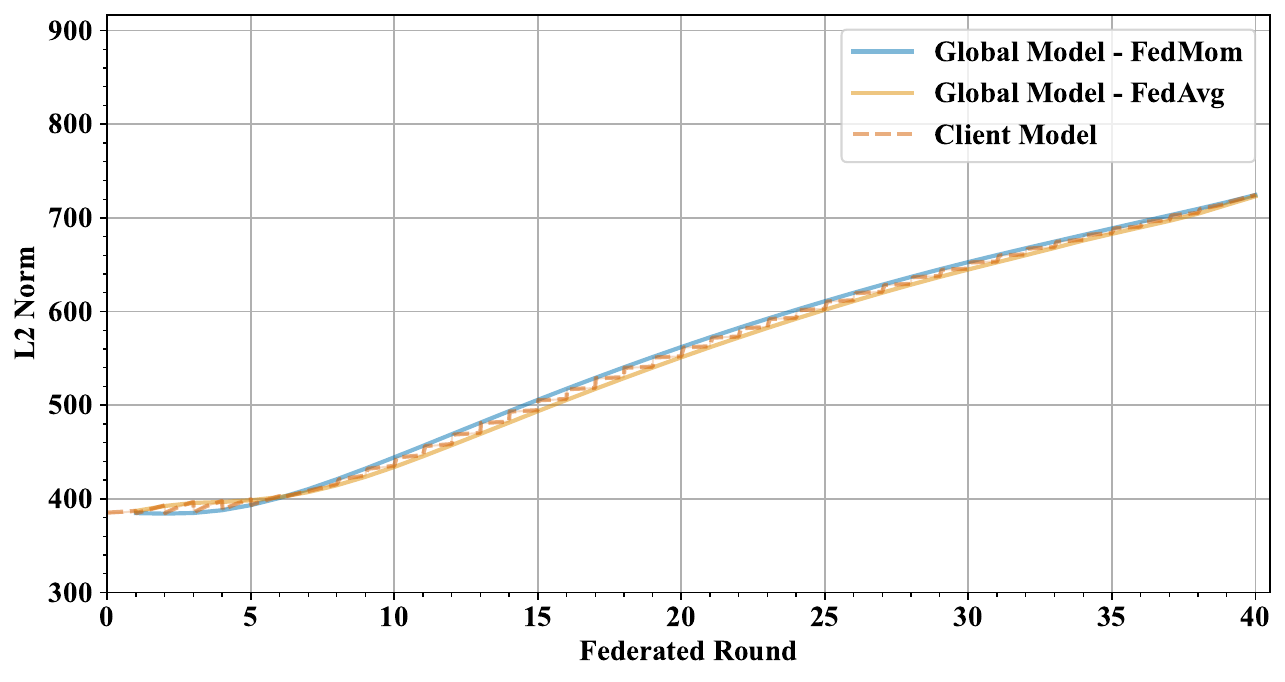}} 
    \subfloat[]{\includegraphics[width=0.45\textwidth]{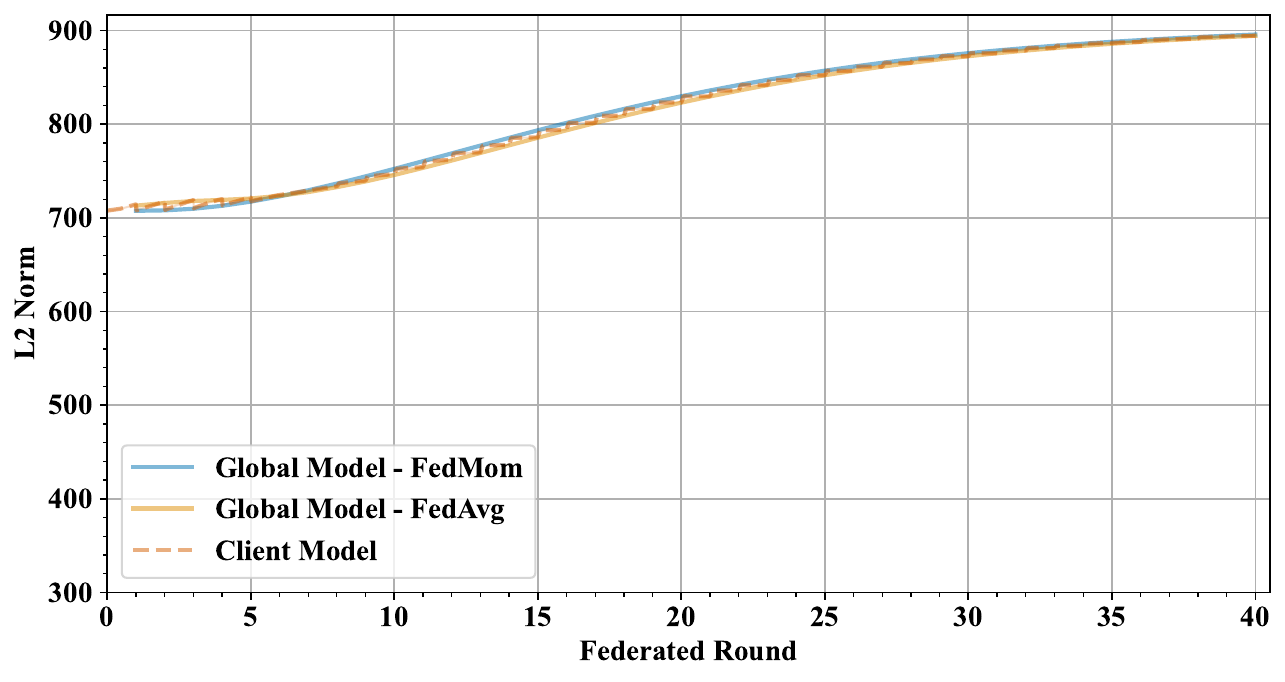}}
    \caption{The $l_2$ norms of the global model, client models, and the average of client models for our $75$M~(a) and $350$M~(b) experiments. While in the transitory early stage, the server model grows slower than the aggregate of client models, it is consistently larger in later stages. These figures reflect both clients reaching a consensus and the ability of the momentum mechanism to stabilize the optimization trajectory.}
    \label{fig:fed:norm-(generic-scale)}
\end{figure}

We further investigate the training dynamics of the models of size \num{75}M and \num{350}M parameters.
In particular, we are interested in understanding how the client and server models interplay as the FL converges.
As \cref{fig:fed:norm-(generic-scale)} shows, the server model initially ``pulls back'' the clients' model norms through the aggregation.
After a few rounds, the federated aggregation starts incrementing the norm of the averaged clients' models until the global and local models converge to the same norm.
\Cref{fig:fed:norm-(generic-scale)} showcases how the client and server optimizations interplay to agree on an initial set of global model parameters and then converge to an optimum. The same trends are present for our heterogeneous partition of The Pile~(\cref{fig:fed:norm-pile}) and for our partial-participation experiments~(\cref{fig:fed:norm-partial}.

\subsection{Federated Optimization Aligns Client Gradients:}

\begin{figure}[ht]
    \centering
    \subfloat[]{\includegraphics[width=0.45\textwidth]{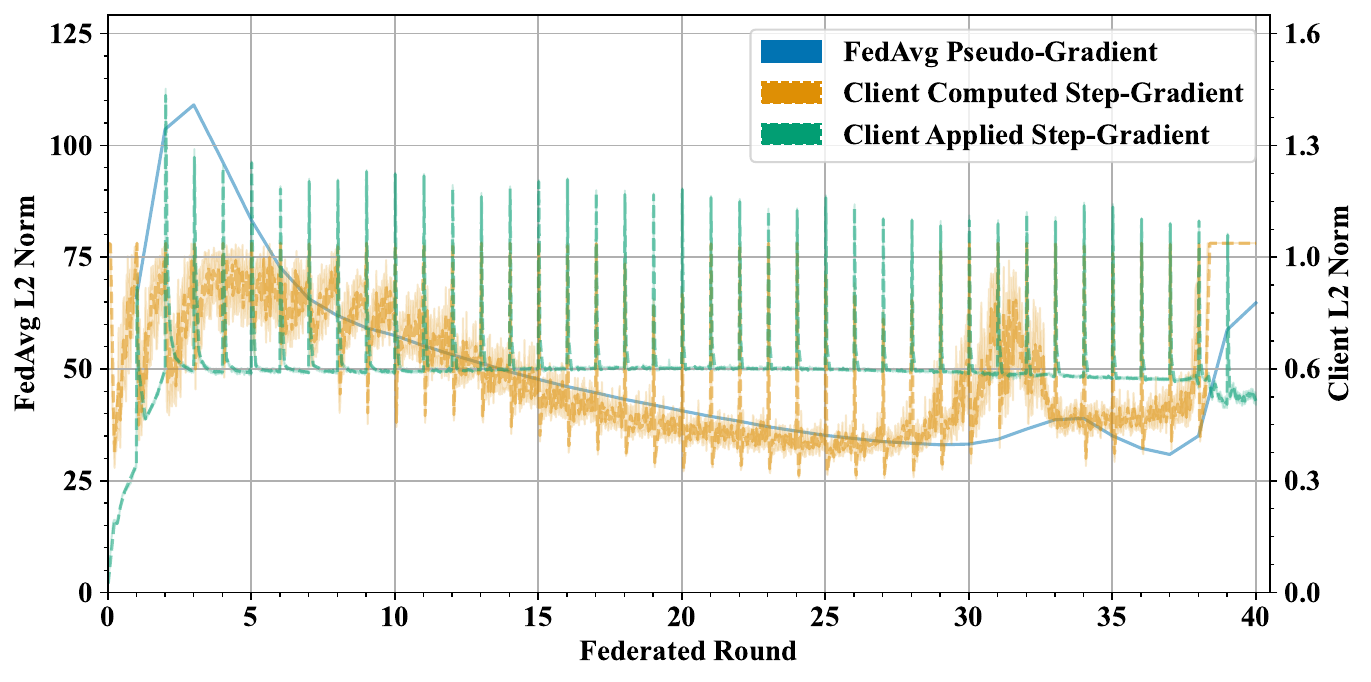}} 
    \subfloat[]{\includegraphics[width=0.45\textwidth]{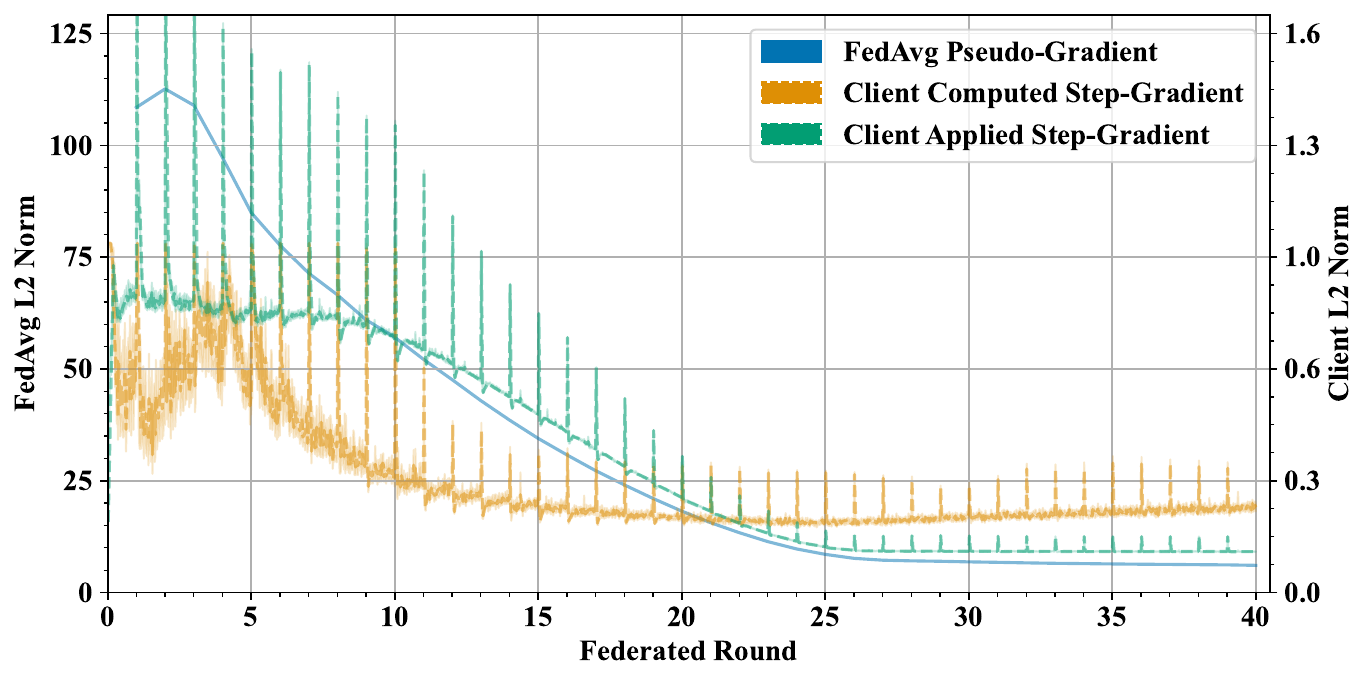}}
    \caption{The $l_2$ norms, for our $75$M~(a) and $350$M~(b) experiments, of the FedAvg Pseudo-Gradient~(average of client deltas relative to the server model of the previous round), the client model gradients computed on a per-step basis, and the client gradients applied to the model when considering learning rate, weight decay, and clipping. The decay in the pseudo-gradient is faster than the decay of the scheduler for the $75$M model, being data-driven, and comparable to that of the step-gradient for the $125$M model.}
    \label{fig:fed:pseudograd-vs-localgrad-(generic-scale)}
\end{figure}

We observe the cause of client model convergence in \cref{fig:fed:pseudograd-vs-localgrad-(generic-scale)}, which showcases the relationship between the norms of the per-round \textbf{pseudo-gradient}~(i.e., average client update) and the local client gradients applied at every SGD step. While the pseudo-gradient starts much larger than the applied local gradients, it reaches a similar or much smaller size as the client models converge. The same trends hold for our results on heterogeneous data, shown in \cref{fig:fed:pseudograd-vs-localgrad-pile}, and for our partial-participation experiments, shown in \cref{fig:fed:pseudograd-vs-localgrad-partial}.

\subsection{Pushing the Boundary on Federated LLM Size}\label{sec:pushing_boundaries}

Having established that larger models reach consensus faster and that federated pre-training can achieve comparable perplexity to centralized pre-training, we now aim to examine how this behavior extrapolates to even larger sizes.
To this end, we used \photon to pre-train $3$B and $7$B models using the same recipe.  

\begin{figure}[ht]
    \centering
    \subfloat[]{\includegraphics[width=0.45\textwidth]{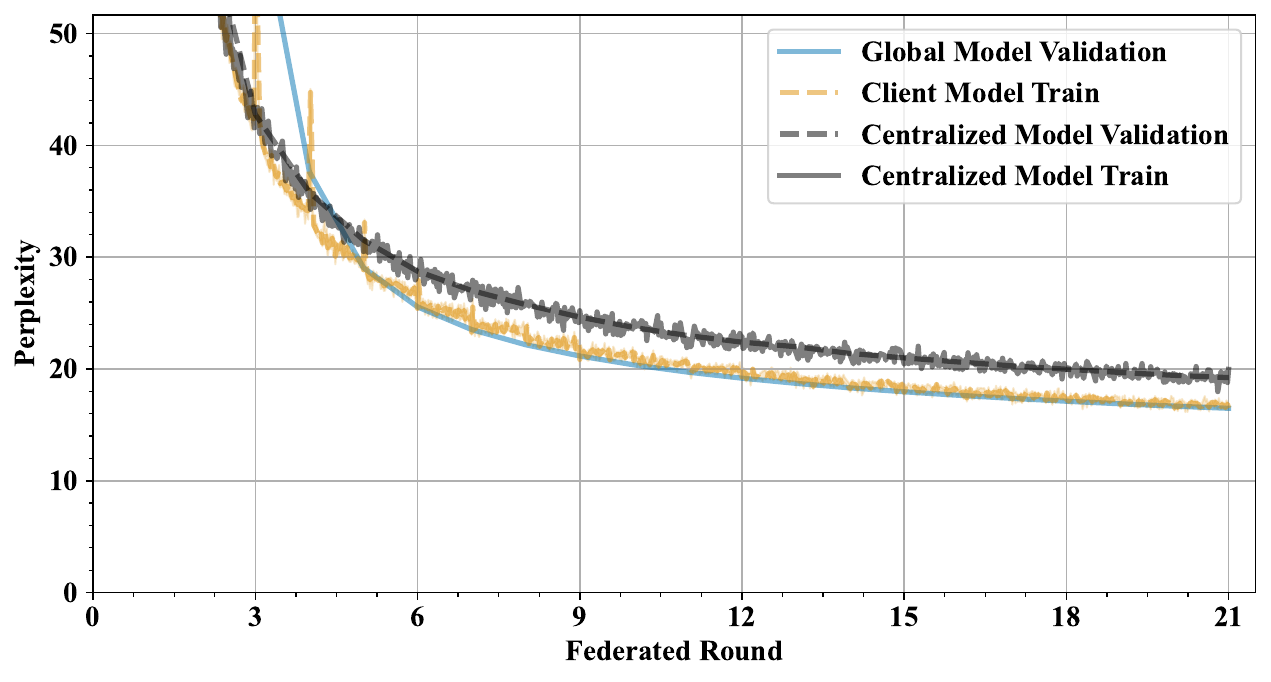}} 
    \subfloat[]{\includegraphics[width=0.45\textwidth]{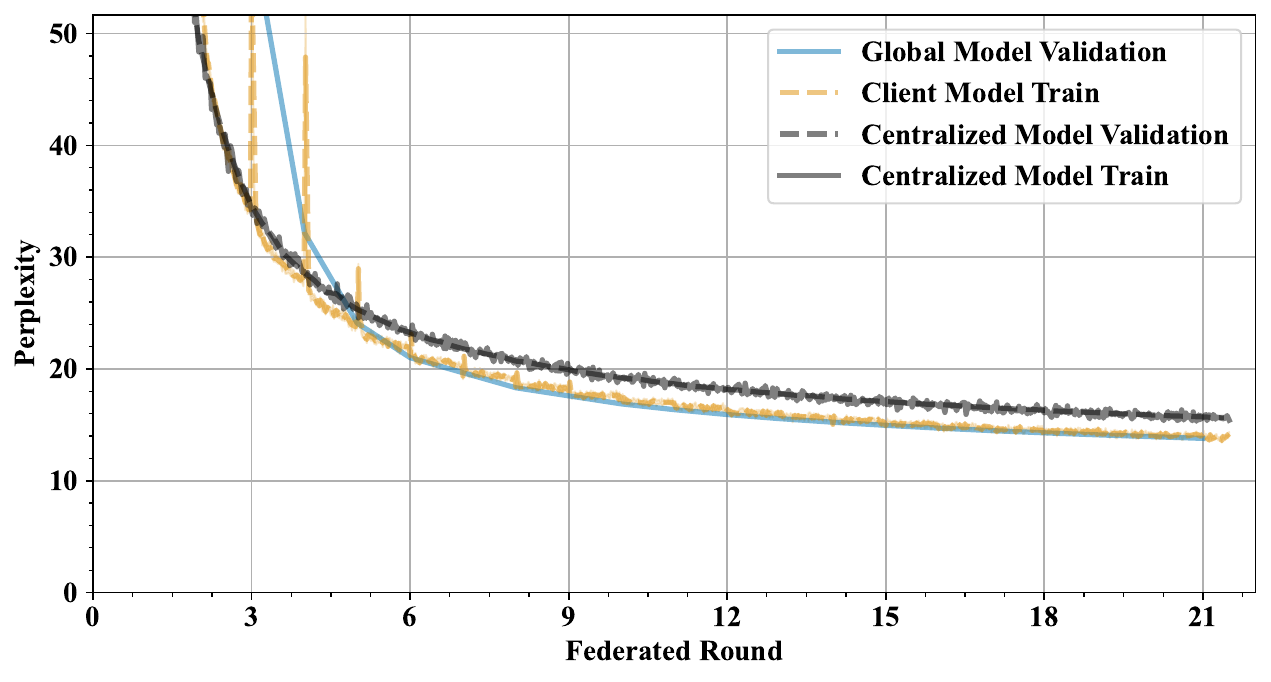}}
    \caption{Comparison between the perplexity of the federated global model evaluated on the centralized validation set, the train perplexities of federated clients (averaged together), and the train and test perplexities for a centralized experiment.
    These metrics are reported for our $3$B~(a) and $7$B~(b) experiments. These large models reach a lower perplexity than their centralized counterparts and are more robust in the face of federated aggregation, with minimal perplexity spikes after the early rounds.}
    \label{fig:fed:perplexity_3b_and_7b}
\end{figure}

As shown in \cref{fig:fed:perplexity_3b_and_7b}, these larger sizes consistently outperform their centralized counterparts regarding client training perplexity and server validation perplexity.
The gap in validation perplexity is larger than the one previously obtained for the $1.3$B, suggesting continued improvements in the effectiveness of federated optimization.

\subsection{On the choice of the outer optimizer}
\begin{figure}[ht]
    \centering
    \subfloat[]{\includegraphics[width=0.45\textwidth]{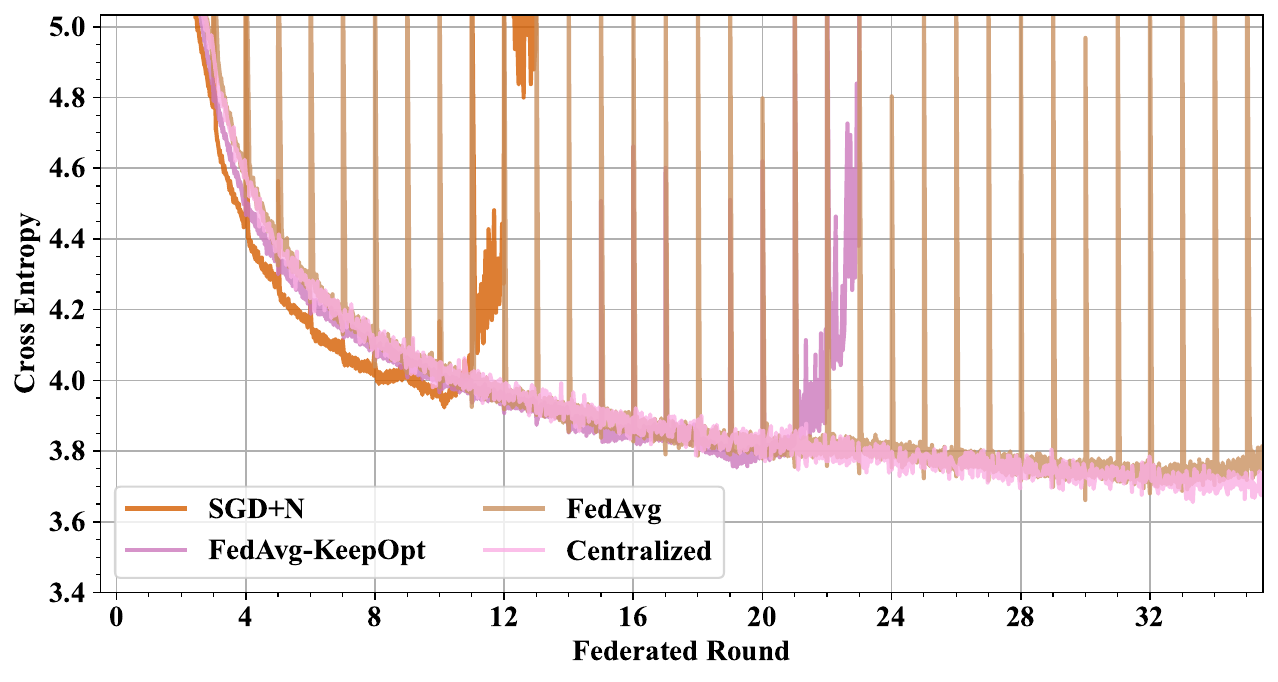}} 
    \subfloat[]{\includegraphics[width=0.45\textwidth]{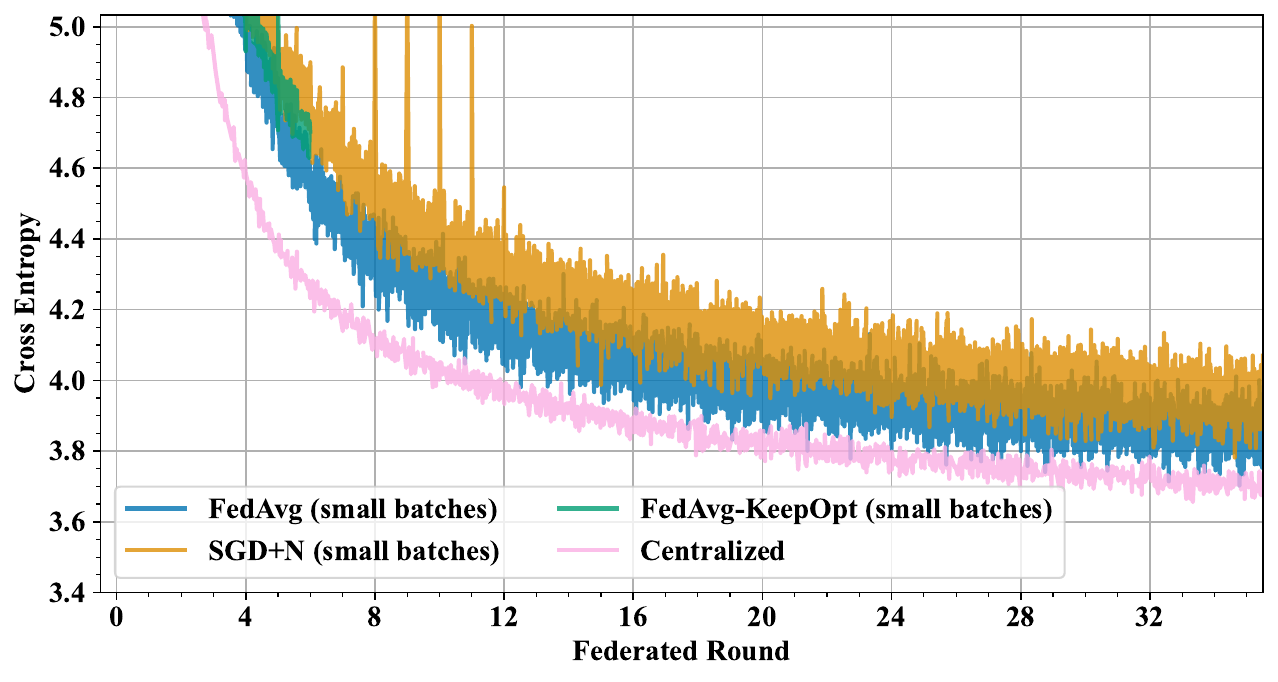}} \\
    \subfloat[]{\includegraphics[width=0.45\textwidth]{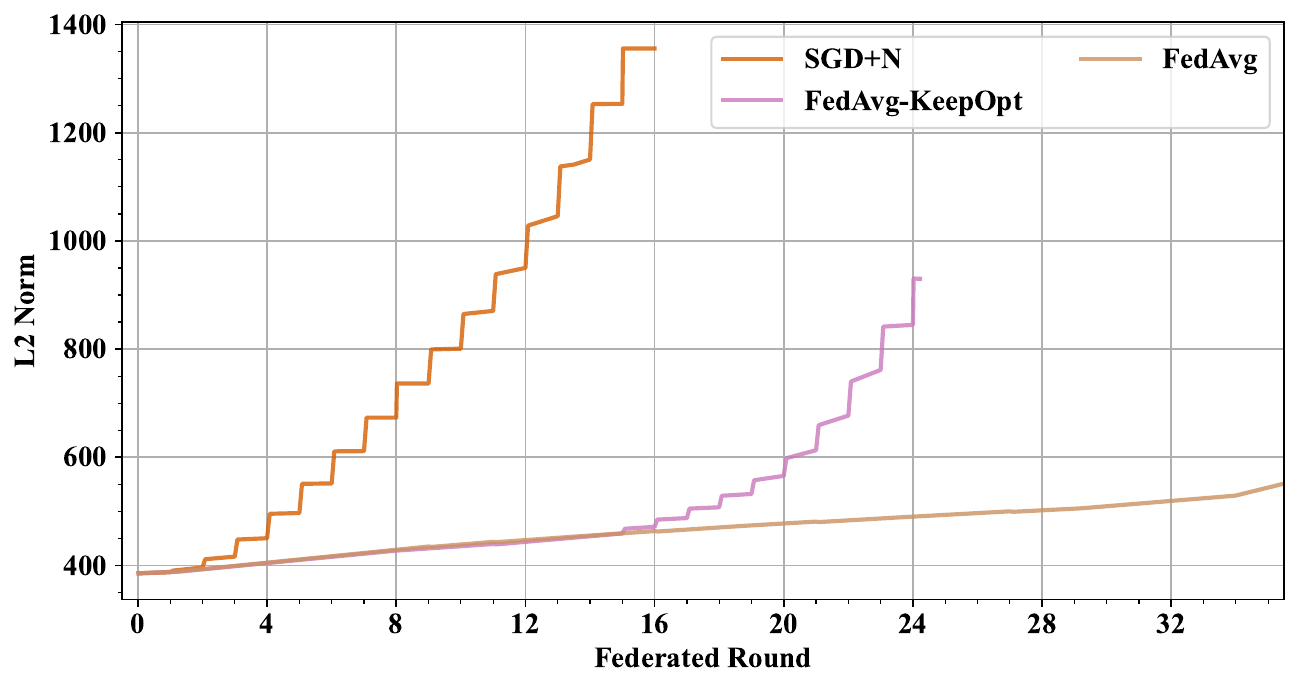}}
        \subfloat[]{\includegraphics[width=0.45\textwidth]{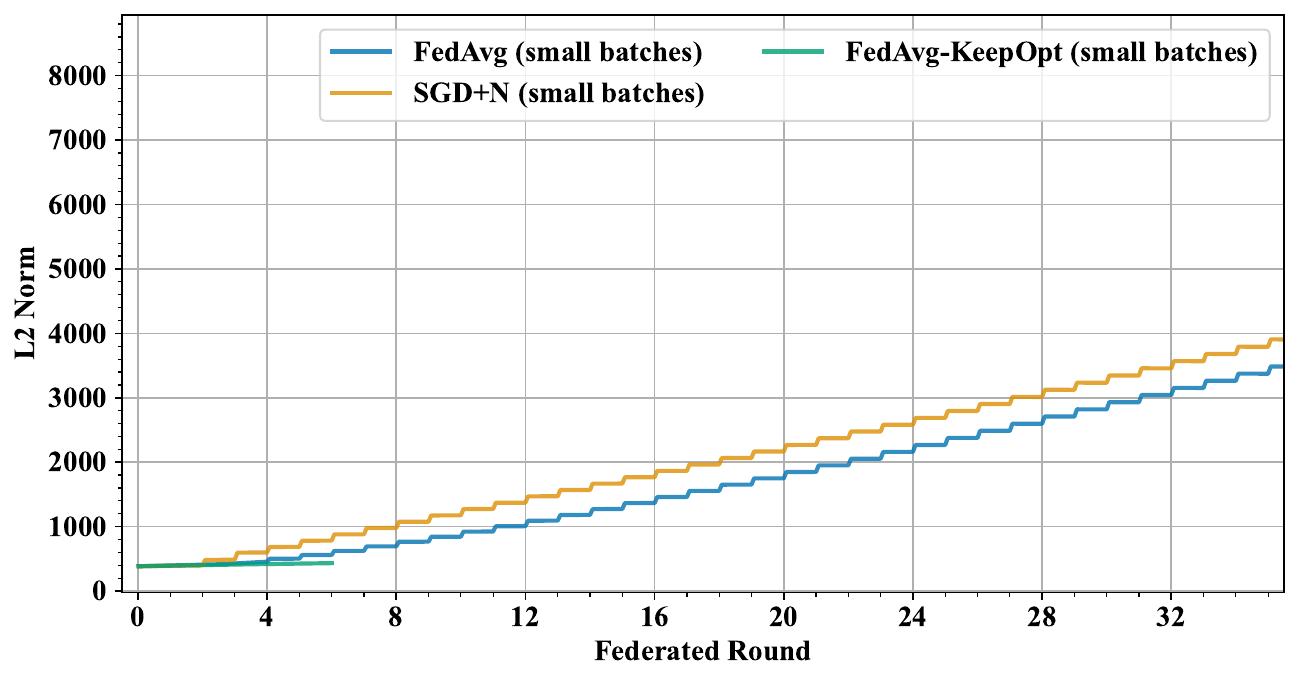}}
    \caption{Training cross-entropy comparison for (a) large batches and (b) small batch sizes. In both conditions, standard FedAvg outperforms SGD+N~(SGD + server-side Nesterov momentum) and FedAvg-KeepOpt~(FedAvg with local optimizer states saved between rounds). As can be seen in (c) and (b), the effect of server-side momentum and of keeping local optimizer states is to cause growth in the l2 norm of the model, causing the eventual divergence in cross-entropy.}
    \label{fig:fed:optimizer_choice}
\end{figure} 

Finally, it is worth considering several of the relevant choices that emerge from the interplay between local and federated optimization.
The first of these is the choice of server-side and local optimizers.
For the local optimizer, we concern ourselves with AdamW~\citep{AdamW}, assuming to preserve the most popular choice of the standard distributed approaches for our local training pipeline.
For the server-side optimizer, we consider FedAvg~\citep{fedavg} and server-side Nesterov momentum~\citep{FedMOM} as used in \citet{DiLoCo}.
A final choice worth considering when using a stateful local optimizer is whether to preserve optimizer states in between rounds. 

The second choice we are concerned with is the local batch size.
Since increasing the number of clients behaves similarly to increasing batch size~\citep{DontUseLargeBatchesUseLocalSGD}, we consider two scenarios: (a) keeping the same batch size locally as the centralized pre-training recipe, (b) decreasing the local batch size such that the effective batch size, number of clients times the local batch size, matches that of centralized pre-training.

Our results, shown in \cref{fig:fed:optimizer_choice}, indicate that standard FedAvg provides the lowest final perplexity and is the most robust to divergence without keeping local optimizer states. While both server-side momentum and keeping optimizer states can provide initial speedups, they fail to approach the perplexity of centralized training. The same trends hold when using smaller batches locally at the expense of greater communication frequency. 

Beyond performance benefits, not maintaining optimizer states also enables stateless clients, allowing, for example, partial participation without staleness, tolerating dropouts, and admitting new entrants into the federations. Furthermore, keeping optimizer states under data heterogeneity may prove cumbersome, as it increases client drift without synchronization. If optimizer states are synchronized, communication costs can grow threefold. Given these factors, we prefer and recommend stateless clients over the approach of previous methods~\citep{DiLoCo}.

\subsection{Evaluation Harness}

\begin{table}[]
\caption{In-context learning comparison between \photon models. Our biggest model wins $6$ out of $7$ comparisons in this group.}
\resizebox{\textwidth}{!}{%
\begin{tabular}{@{}lccccccc@{}}
\toprule
\textbf{Name} & \begin{tabular}[c]{@{}c@{}}\textbf{ARC-Challenge}\\ \citep{arc_challenge}\end{tabular} & \textbf{\begin{tabular}[c]{@{}c@{}}BigBench \\ QA Wikidata\\ \citep{bigbench}\end{tabular}} & \begin{tabular}[c]{@{}c@{}}\textbf{HellaSwag}\\  \citep{hellaswag}\end{tabular} & \begin{tabular}[c]{@{}c@{}}\textbf{PIQA} \\ \citep{piqa}\end{tabular} & \begin{tabular}[c]{@{}c@{}}\textbf{Winogrande} \\ \citep{winogrande}\end{tabular} & \begin{tabular}[c]{@{}c@{}}\textbf{ARC-Easy}\\ \citep{arc_challenge}\end{tabular} & \begin{tabular}[c]{@{}c@{}}\textbf{BoolQ} \\ \citep{boolq}\end{tabular} \\ \midrule
\rowcolor{lightgray} \textbf{Photon-7B} & $\mathbf{0.265}$ & $\mathbf{0.447}$ & $\mathbf{0.524}$ & $\mathbf{0.729}$ & 0.522 & $\mathbf{0.508}$ & 0.530 \\
\textbf{Photon-3B} & 0.247 & 0.360 & 0.455 & 0.705 & 0.512 & 0.461 & 0.591 \\
\rowcolor{lightgray} \textbf{Photon-1B} & 0.243 & 0.215 & 0.349 & 0.676 & 0.516 & 0.390 & $\mathbf{0.612}$ \\
\end{tabular}%
}
\label{tab:eval_gauntlet_1}
\end{table}

\begin{table}[]
\caption{In-context learning comparison between \photon models. Our biggest model wins $5$ out of $6$ comparisons in this group.}
\resizebox{\textwidth}{!}{%
\begin{tabular}{@{}lcccccc@{}}
\toprule
\textbf{Name} & \textbf{\begin{tabular}[c]{@{}c@{}}\textbf{Openbook QA} \\ \citep{openbookqa}\end{tabular}} & \textbf{\begin{tabular}[c]{@{}c@{}}\textbf{Winograd} \\ \citep{winograd}\end{tabular}} & \textbf{\begin{tabular}[c]{@{}c@{}}LAMBADA\\ (OpenAI) \citep{lambada}\end{tabular}} & \textbf{\begin{tabular}[c]{@{}c@{}}Bigbench \citep{bigbench} \\ Strategy QA\end{tabular}} & \textbf{\begin{tabular}[c]{@{}c@{}}\textbf{COPA} \\ \citep{copa}\end{tabular}} & \textbf{\begin{tabular}[c]{@{}c@{}}\textbf{MMLU} \\ \citep{mmlu}\end{tabular}} \\ \midrule
\rowcolor{lightgray}  \textbf{Photon-7B} & $\mathbf{0.358}$ & $\mathbf{0.681}$ & $\mathbf{0.457}$ & 0.466 & $\mathbf{0.710}$ & $\mathbf{0.263}$ \\
\textbf{Photon-3B} & 0.316 & 0.656 & 0.381 & 0.464 & 0.620 & 0.252 \\
\rowcolor{lightgray}  \textbf{Photon-1B} & 0.274 & 0.604 & 0.298 & 0.470 & 0.630 & 0.248 \\
\end{tabular}%
}
\label{tab:eval_gauntlet_2}
\end{table}

To evaluate the downstream task performance of our models, we test across a series of in-context learning benchmarks. Our results, shown in \cref{tab:eval_gauntlet_1,tab:eval_gauntlet_2}, demonstrate that the downstream performance of models trained with \photon scales as expected with model size, with our largest model winning $10$ out of $14$ reasons. This proves the downstream utility of \photon models even when using a pre-training dataset not optimized for downstream performance. We expect that as we continue to increase the model size and incorporate a broader and more qualitative data mixture, the downstream performance of \photon models will keep improving.

\FloatBarrier

\section{Future Work}

This work presents a complete system for generative pre-training of LLMs across a federation of clients, each participating with its own computing and data sources.
In this section, we present a summary of the future direction and possibilities this work will allow.
We plan to fine-tune these models on established benchmark tasks to assess further their performance in the broad range of downstream applications the participants may be interested in.
We also plan to scale up the federated setting presented in this work regarding the population and model size.
In addition, we will conduct further investigations into the heterogeneity of data sources, such as different languages or genres, and how this inherent property of FL impacts the capabilities of these federated models.
These investigations will illuminate the potential advantages of collaboration compared to the isolated centralized paradigm. We expect FL's benefits to grow only as we incorporate more extensive private datasets into the training procedure.
Since FL has been introduced as a privacy-preserving by-design technique, we will analyze this paradigm's potential privacy-preservation strengths and weaknesses.
As we advance the proposed system, we will delineate and develop further optimizations to mitigate the few overheads remaining in the execution runtime.
This will eventually make our proposal more efficient and effective for allowing an even broader participation of the FL.
\section{Conclusions}

We successfully demonstrate the potential of LLM  federated generative pre-training by obtaining the first federated billion-scale model fully trained in a heterogeneous FL setting and showing how federated training is robust to both data heterogeneity and partial participation.
The complete system presented in this work allows for collaborative, reproducible, and scalable pre-training of LLMs, releasing the computational and data resources spread across the planet.
Our system is built on an \textbf{open-source} framework~\citep{Flower} and will soon be made publicly available.
Furthermore, we fully disclose our training recipe to enable future development and collaborations.
Our work advances LLM pre-training by democratizing it through federated technologies.
This allows data-rich actors to pool resources together, whether extensive or limited.
Given current trends, we believe that the future of LLM pre-training lies in the data sources that FL can bring together.

\FloatBarrier

\begin{ack}
All costs for the computation used for this work was funded by Flower Labs, and the research conducted by a team of researchers from Flower Labs and The University of Cambridge. Support for university based researchers came from a variety of sources, but in particular the following funding organizations are are acknowledged: the European Research Council (REDIAL), the Royal Academy of Engineering (DANTE) and the Ministry of Education of Romania through the Credit and Scholarship Agency.
\end{ack}

\vskip 0.2in
{
\bibliographystyle{unsrtnat}
\bibliography{llm}

\begin{thebibliography}{118}
\providecommand{\natexlab}[1]{#1}
\providecommand{\url}[1]{\texttt{#1}}
\expandafter\ifx\csname urlstyle\endcsname\relax
  \providecommand{\doi}[1]{doi: #1}\else
  \providecommand{\doi}{doi: \begingroup \urlstyle{rm}\Url}\fi

\bibitem[Lane et~al.(2024)Lane, Sani, and Iacob]{flowerllm-blogpost}
Nicholas~D. Lane, Lorenzo Sani, and Alex Iacob.
\newblock {Introducing FlowerLLM}, 2024.
\newblock URL \url{https://flower.ai/blog/2024-03-14-introducing-flowerllm/}.

\bibitem[Kaplan et~al.(2020)Kaplan, McCandlish, Henighan, Brown, Chess, Child, Gray, Radford, Wu, and Amodei]{OgScalingLaws}
Jared Kaplan, Sam McCandlish, Tom Henighan, Tom~B. Brown, Benjamin Chess, Rewon Child, Scott Gray, Alec Radford, Jeffrey Wu, and Dario Amodei.
\newblock Scaling laws for neural language models.
\newblock \emph{CoRR}, abs/2001.08361, 2020.
\newblock URL \url{https://arxiv.org/abs/2001.08361}.

\bibitem[{Berriel} et~al.(2017){Berriel}, {Lopes}, {Rodrigues}, {Varejão}, and {Oliveira-Santos}]{EnergyConsumptionForecast}
R.~F. {Berriel}, A.~T. {Lopes}, A.~{Rodrigues}, F.~M. {Varejão}, and T.~{Oliveira-Santos}.
\newblock Monthly energy consumption forecast: A deep learning approach.
\newblock In \emph{2017 International Joint Conference on Neural Networks (IJCNN)}, pages 4283--4290, 2017.

\bibitem[Cao et~al.(2023)Cao, Basar, Diggavi, Eldar, Letaief, Poor, and Zhang]{comm_eff_distr_learning_survey}
Xuanyu Cao, Tamer Basar, Suhas~N. Diggavi, Yonina~C. Eldar, Khaled~B. Letaief, H.~Vincent Poor, and Junshan Zhang.
\newblock Communication-efficient distributed learning: An overview.
\newblock \emph{{IEEE} J. Sel. Areas Commun.}, 41\penalty0 (4):\penalty0 851--873, 2023.
\newblock \doi{10.1109/JSAC.2023.3242710}.
\newblock URL \url{https://doi.org/10.1109/JSAC.2023.3242710}.

\bibitem[Guerra et~al.(2023)Guerra, Wilhelmi, Miozzo, and Dini]{training_cost_distr_data}
Elia Guerra, Francesc Wilhelmi, Marco Miozzo, and Paolo Dini.
\newblock The cost of training machine learning models over distributed data sources.
\newblock \emph{{IEEE} Open J. Commun. Soc.}, 4:\penalty0 1111--1126, 2023.
\newblock \doi{10.1109/OJCOMS.2023.3274394}.
\newblock URL \url{https://doi.org/10.1109/OJCOMS.2023.3274394}.

\bibitem[Hoffmann et~al.(2022)Hoffmann, Borgeaud, Mensch, Buchatskaya, Cai, Rutherford, de~Las~Casas, Hendricks, Welbl, Clark, Hennigan, Noland, Millican, van~den Driessche, Damoc, Guy, Osindero, Simonyan, Elsen, Rae, Vinyals, and Sifre]{TrainingComputeOptimalLLMs}
Jordan Hoffmann, Sebastian Borgeaud, Arthur Mensch, Elena Buchatskaya, Trevor Cai, Eliza Rutherford, Diego de~Las~Casas, Lisa~Anne Hendricks, Johannes Welbl, Aidan Clark, Tom Hennigan, Eric Noland, Katie Millican, George van~den Driessche, Bogdan Damoc, Aurelia Guy, Simon Osindero, Karen Simonyan, Erich Elsen, Jack~W. Rae, Oriol Vinyals, and Laurent Sifre.
\newblock Training compute-optimal large language models.
\newblock \emph{CoRR}, abs/2203.15556, 2022.
\newblock \doi{10.48550/ARXIV.2203.15556}.
\newblock URL \url{https://doi.org/10.48550/arXiv.2203.15556}.

\bibitem[Grynbaum and Mac(2023)]{NytVsOpenAi}
Michael ~M Grynbaum and Ryan Mac.
\newblock The times sues openai and microsoft over a.i. use of copyrighted work, Dec 2023.
\newblock URL \url{https://www.nytimes.com/2023/12/27/business/media/new-york-times-open-ai-microsoft-lawsuit.html}.

\bibitem[Bashlovkina et~al.(2023)Bashlovkina, Kuang, Matthews, Clifford, Jun, Cohen, and Baumgartner]{trusted_sources_alignment}
Vasilisa Bashlovkina, Zhaobin Kuang, Riley Matthews, Edward Clifford, Yennie Jun, William~W. Cohen, and Simon Baumgartner.
\newblock Trusted source alignment in large language models.
\newblock \emph{CoRR}, abs/2311.06697, 2023.
\newblock \doi{10.48550/ARXIV.2311.06697}.
\newblock URL \url{https://doi.org/10.48550/arXiv.2311.06697}.

\bibitem[Shumailov et~al.(2023)Shumailov, Shumaylov, Zhao, Gal, Papernot, and Anderson]{the_curese_of_recursion}
Ilia Shumailov, Zakhar Shumaylov, Yiren Zhao, Yarin Gal, Nicolas Papernot, and Ross~J. Anderson.
\newblock The curse of recursion: Training on generated data makes models forget.
\newblock \emph{CoRR}, abs/2305.17493, 2023.
\newblock \doi{10.48550/ARXIV.2305.17493}.
\newblock URL \url{https://doi.org/10.48550/arXiv.2305.17493}.

\bibitem[Desai et~al.(2024)Desai, Pasquetto, Jacobs, and Card]{archival_pretraining_data}
Meera~A Desai, Irene~V Pasquetto, Abigail~Z Jacobs, and Dallas Card.
\newblock An archival perspective on pretraining data.
\newblock \emph{Patterns (N Y)}, 5\penalty0 (4):\penalty0 100966, March 2024.

\bibitem[Tram{\`{e}}r et~al.(2022)Tram{\`{e}}r, Kamath, and Carlini]{dp_pre_training_datasets}
Florian Tram{\`{e}}r, Gautam Kamath, and Nicholas Carlini.
\newblock Considerations for differentially private learning with large-scale public pretraining.
\newblock \emph{CoRR}, abs/2212.06470, 2022.
\newblock \doi{10.48550/ARXIV.2212.06470}.
\newblock URL \url{https://doi.org/10.48550/arXiv.2212.06470}.

\bibitem[OpenAI(2023{\natexlab{a}})]{OpenAISpringerDeal}
OpenAI, Dec 2023{\natexlab{a}}.
\newblock URL \url{https://openai.com/blog/axel-springer-partnership}.

\bibitem[Villalobos et~al.(2022)Villalobos, Sevilla, Heim, Besiroglu, Hobbhahn, and Ho]{WillWeRunOutOfData}
Pablo Villalobos, Jaime Sevilla, Lennart Heim, Tamay Besiroglu, Marius Hobbhahn, and Anson Ho.
\newblock Will we run out of data? an analysis of the limits of scaling datasets in machine learning.
\newblock \emph{CoRR}, abs/2211.04325, 2022.
\newblock \doi{10.48550/ARXIV.2211.04325}.
\newblock URL \url{https://doi.org/10.48550/arXiv.2211.04325}.

\bibitem[Abdali et~al.(2024)Abdali, Anarfi, Barberan, and He]{SecuringLLMs}
Sara Abdali, Richard Anarfi, C.~J. Barberan, and Jia He.
\newblock Securing large language models: Threats, vulnerabilities and responsible practices.
\newblock \emph{CoRR}, abs/2403.12503, 2024.
\newblock \doi{10.48550/ARXIV.2403.12503}.
\newblock URL \url{https://doi.org/10.48550/arXiv.2403.12503}.

\bibitem[Borkar(2023)]{WhatCanWeLearnDataLeakageLaw}
Jaydeep Borkar.
\newblock What can we learn from data leakage and unlearning for law?
\newblock \emph{CoRR}, abs/2307.10476, 2023.
\newblock \doi{10.48550/ARXIV.2307.10476}.
\newblock URL \url{https://doi.org/10.48550/arXiv.2307.10476}.

\bibitem[Yu et~al.(2023)Yu, Mu{\~{n}}oz, and Jannesari]{fed_FMs}
Sixing Yu, J.~Pablo Mu{\~{n}}oz, and Ali Jannesari.
\newblock Federated foundation models: Privacy-preserving and collaborative learning for large models.
\newblock \emph{CoRR}, abs/2305.11414, 2023.
\newblock \doi{10.48550/ARXIV.2305.11414}.
\newblock URL \url{https://doi.org/10.48550/arXiv.2305.11414}.

\bibitem[Stich(2019)]{LocalSGD}
Sebastian~U. Stich.
\newblock Local {SGD} converges fast and communicates little.
\newblock In \emph{7th International Conference on Learning Representations, {ICLR} 2019, New Orleans, LA, USA, May 6-9, 2019}. OpenReview.net, 2019.
\newblock URL \url{https://openreview.net/forum?id=S1g2JnRcFX}.

\bibitem[Lin et~al.(2020)Lin, Stich, Patel, and Jaggi]{DontUseLargeBatchesUseLocalSGD}
Tao Lin, Sebastian~U. Stich, Kumar~Kshitij Patel, and Martin Jaggi.
\newblock Don't use large mini-batches, use local {SGD}.
\newblock In \emph{8th International Conference on Learning Representations, {ICLR} 2020, Addis Ababa, Ethiopia, April 26-30, 2020}. OpenReview.net, 2020.
\newblock URL \url{https://openreview.net/forum?id=B1eyO1BFPr}.

\bibitem[McMahan et~al.(2017)McMahan, Moore, Ramage, Hampson, and y~Arcas]{fedavg}
Brendan McMahan, Eider Moore, Daniel Ramage, Seth Hampson, and Blaise~Aguera y~Arcas.
\newblock Communication-efficient learning of deep networks from decentralized data.
\newblock In \emph{Artificial intelligence and statistics}. PMLR, 2017.

\bibitem[Douillard et~al.(2023)Douillard, Feng, Rusu, Chhaparia, Donchev, Kuncoro, Ranzato, Szlam, and Shen]{DiLoCo}
Arthur Douillard, Qixuang Feng, Andrei~A. Rusu, Rachita Chhaparia, Yani Donchev, Adhiguna Kuncoro, Marc'Aurelio Ranzato, Arthur Szlam, and Jiajun Shen.
\newblock Diloco: Distributed low-communication training of language models.
\newblock \emph{CoRR}, abs/2311.08105, 2023.
\newblock \doi{10.48550/ARXIV.2311.08105}.
\newblock URL \url{https://doi.org/10.48550/arXiv.2311.08105}.

\bibitem[Liu et~al.(2024)Liu, Chhaparia, Douillard, Kale, Rusu, Shen, Szlam, and Ranzato]{asyncDiLoCo}
Bo~Liu, Rachita Chhaparia, Arthur Douillard, Satyen Kale, Andrei~A. Rusu, Jiajun Shen, Arthur Szlam, and Marc'Aurelio Ranzato.
\newblock Asynchronous local-sgd training for language modeling.
\newblock \emph{CoRR}, abs/2401.09135, 2024.
\newblock \doi{10.48550/ARXIV.2401.09135}.
\newblock URL \url{https://doi.org/10.48550/arXiv.2401.09135}.

\bibitem[Douillard et~al.(2024)Douillard, Feng, Rusu, Kuncoro, Donchev, Chhaparia, Gog, Ranzato, Shen, and Szlam]{DiPaCo}
Arthur Douillard, Qixuan Feng, Andrei~A. Rusu, Adhiguna Kuncoro, Yani Donchev, Rachita Chhaparia, Ionel Gog, Marc'Aurelio Ranzato, Jiajun Shen, and Arthur Szlam.
\newblock Dipaco: Distributed path composition.
\newblock \emph{CoRR}, abs/2403.10616, 2024.
\newblock \doi{10.48550/ARXIV.2403.10616}.
\newblock URL \url{https://doi.org/10.48550/arXiv.2403.10616}.

\bibitem[Beutel et~al.(2022)Beutel, Topal, Mathur, Qiu, Fernandez-Marques, Gao, Sani, Li, Parcollet, de~Gusmão, and Lane]{Flower}
Daniel~J. Beutel, Taner Topal, Akhil Mathur, Xinchi Qiu, Javier Fernandez-Marques, Yan Gao, Lorenzo Sani, Kwing~Hei Li, Titouan Parcollet, Pedro Porto~Buarque de~Gusmão, and Nicholas~D. Lane.
\newblock Flower: A friendly federated learning research framework.
\newblock \emph{CoRR}, abs/2007.14390, 2022.
\newblock URL \url{https://arxiv.org/abs/2007.14390}.

\bibitem[Sani et~al.(2023)Sani, de~Gusm{\~{a}}o, Iacob, Zhao, Qiu, Gao, Fern{\'{a}}ndez{-}Marqu{\'{a}}s, and Lane]{Pollen}
Lorenzo Sani, Pedro Porto~Buarque de~Gusm{\~{a}}o, Alex Iacob, Wanru Zhao, Xinchi Qiu, Yan Gao, Javier Fern{\'{a}}ndez{-}Marqu{\'{a}}s, and Nicholas~Donald Lane.
\newblock High-throughput simulation of federated learning via resource-aware client placement.
\newblock \emph{CoRR}, abs/2306.17453, 2023.
\newblock \doi{10.48550/ARXIV.2306.17453}.
\newblock URL \url{https://doi.org/10.48550/arXiv.2306.17453}.

\bibitem[Brown et~al.(2020)Brown, Mann, Ryder, Subbiah, Kaplan, Dhariwal, Neelakantan, Shyam, Sastry, Askell, Agarwal, Herbert-Voss, Krueger, Henighan, Child, Ramesh, Ziegler, Wu, Winter, Hesse, Chen, Sigler, Litwin, Gray, Chess, Clark, Berner, McCandlish, Radford, Sutskever, and Amodei]{gpt3}
Tom~B. Brown, Benjamin Mann, Nick Ryder, Melanie Subbiah, Jared Kaplan, Prafulla Dhariwal, Arvind Neelakantan, Pranav Shyam, Girish Sastry, Amanda Askell, Sandhini Agarwal, Ariel Herbert-Voss, Gretchen Krueger, Tom Henighan, Rewon Child, Aditya Ramesh, Daniel~M. Ziegler, Jeffrey Wu, Clemens Winter, Christopher Hesse, Mark Chen, Eric Sigler, Mateusz Litwin, Scott Gray, Benjamin Chess, Jack Clark, Christopher Berner, Sam McCandlish, Alec Radford, Ilya Sutskever, and Dario Amodei.
\newblock Language models are few-shot learners, 2020.

\bibitem[OpenAI(2023{\natexlab{b}})]{gpt4}
OpenAI.
\newblock {GPT-4} technical report.
\newblock \emph{CoRR}, abs/2303.08774, 2023{\natexlab{b}}.
\newblock \doi{10.48550/ARXIV.2303.08774}.
\newblock URL \url{https://doi.org/10.48550/arXiv.2303.08774}.

\bibitem[Anil et~al.(2023)Anil, Borgeaud, Wu, Alayrac, Yu, Soricut, Schalkwyk, Dai, Hauth, Millican, Silver, Petrov, Johnson, Antonoglou, Schrittwieser, Glaese, Chen, Pitler, Lillicrap, Lazaridou, Firat, Molloy, Isard, Barham, Hennigan, Lee, Viola, Reynolds, Xu, Doherty, Collins, Meyer, Rutherford, Moreira, Ayoub, Goel, Tucker, Piqueras, Krikun, Barr, Savinov, Danihelka, Roelofs, White, Andreassen, von Glehn, Yagati, Kazemi, Gonzalez, Khalman, Sygnowski, and et~al.]{GEMINI}
Rohan Anil, Sebastian Borgeaud, Yonghui Wu, Jean{-}Baptiste Alayrac, Jiahui Yu, Radu Soricut, Johan Schalkwyk, Andrew~M. Dai, Anja Hauth, Katie Millican, David Silver, Slav Petrov, Melvin Johnson, Ioannis Antonoglou, Julian Schrittwieser, Amelia Glaese, Jilin Chen, Emily Pitler, Timothy~P. Lillicrap, Angeliki Lazaridou, Orhan Firat, James Molloy, Michael Isard, Paul~Ronald Barham, Tom Hennigan, Benjamin Lee, Fabio Viola, Malcolm Reynolds, Yuanzhong Xu, Ryan Doherty, Eli Collins, Clemens Meyer, Eliza Rutherford, Erica Moreira, Kareem Ayoub, Megha Goel, George Tucker, Enrique Piqueras, Maxim Krikun, Iain Barr, Nikolay Savinov, Ivo Danihelka, Becca Roelofs, Ana{\"{\i}}s White, Anders Andreassen, Tamara von Glehn, Lakshman Yagati, Mehran Kazemi, Lucas Gonzalez, Misha Khalman, Jakub Sygnowski, and et~al.
\newblock Gemini: {A} family of highly capable multimodal models.
\newblock \emph{CoRR}, abs/2312.11805, 2023.
\newblock \doi{10.48550/ARXIV.2312.11805}.
\newblock URL \url{https://doi.org/10.48550/arXiv.2312.11805}.

\bibitem[Scao et~al.(2022)Scao, Fan, Akiki, Pavlick, Ilic, Hesslow, Castagn{\'{e}}, Luccioni, Yvon, Gall{\'{e}}, Tow, Rush, Biderman, Webson, Ammanamanchi, Wang, Sagot, Muennighoff, del Moral, Ruwase, Bawden, Bekman, McMillan{-}Major, Beltagy, Nguyen, Saulnier, Tan, Suarez, Sanh, Lauren{\c{c}}on, Jernite, Launay, Mitchell, Raffel, Gokaslan, Simhi, Soroa, Aji, Alfassy, Rogers, Nitzav, Xu, Mou, Emezue, Klamm, Leong, van Strien, Adelani, and et~al.]{BLOOM}
Teven~Le Scao, Angela Fan, Christopher Akiki, Ellie Pavlick, Suzana Ilic, Daniel Hesslow, Roman Castagn{\'{e}}, Alexandra~Sasha Luccioni, Fran{\c{c}}ois Yvon, Matthias Gall{\'{e}}, Jonathan Tow, Alexander~M. Rush, Stella Biderman, Albert Webson, Pawan~Sasanka Ammanamanchi, Thomas Wang, Beno{\^{\i}}t Sagot, Niklas Muennighoff, Albert~Villanova del Moral, Olatunji Ruwase, Rachel Bawden, Stas Bekman, Angelina McMillan{-}Major, Iz~Beltagy, Huu Nguyen, Lucile Saulnier, Samson Tan, Pedro~Ortiz Suarez, Victor Sanh, Hugo Lauren{\c{c}}on, Yacine Jernite, Julien Launay, Margaret Mitchell, Colin Raffel, Aaron Gokaslan, Adi Simhi, Aitor Soroa, Alham~Fikri Aji, Amit Alfassy, Anna Rogers, Ariel~Kreisberg Nitzav, Canwen Xu, Chenghao Mou, Chris Emezue, Christopher Klamm, Colin Leong, Daniel van Strien, David~Ifeoluwa Adelani, and et~al.
\newblock {BLOOM:} {A} 176b-parameter open-access multilingual language model.
\newblock \emph{CoRR}, abs/2211.05100, 2022.
\newblock \doi{10.48550/ARXIV.2211.05100}.
\newblock URL \url{https://doi.org/10.48550/arXiv.2211.05100}.

\bibitem[Touvron et~al.(2023{\natexlab{a}})Touvron, Lavril, Izacard, Martinet, Lachaux, Lacroix, Rozière, Goyal, Hambro, Azhar, Rodriguez, Joulin, Grave, and Lample]{llama}
Hugo Touvron, Thibaut Lavril, Gautier Izacard, Xavier Martinet, Marie-Anne Lachaux, Timothée Lacroix, Baptiste Rozière, Naman Goyal, Eric Hambro, Faisal Azhar, Aurelien Rodriguez, Armand Joulin, Edouard Grave, and Guillaume Lample.
\newblock Llama: Open and efficient foundation language models, 2023{\natexlab{a}}.

\bibitem[Touvron et~al.(2023{\natexlab{b}})Touvron, Martin, Stone, Albert, Almahairi, Babaei, Bashlykov, Batra, Bhargava, Bhosale, Bikel, Blecher, Ferrer, Chen, Cucurull, Esiobu, Fernandes, Fu, Fu, Fuller, Gao, Goswami, Goyal, Hartshorn, Hosseini, Hou, Inan, Kardas, Kerkez, Khabsa, Kloumann, Korenev, Koura, Lachaux, Lavril, Lee, Liskovich, Lu, Mao, Martinet, Mihaylov, Mishra, Molybog, Nie, Poulton, Reizenstein, Rungta, Saladi, Schelten, Silva, Smith, Subramanian, Tan, Tang, Taylor, Williams, Kuan, Xu, Yan, Zarov, Zhang, Fan, Kambadur, Narang, Rodriguez, Stojnic, Edunov, and Scialom]{llama2}
Hugo Touvron, Louis Martin, Kevin Stone, Peter Albert, Amjad Almahairi, Yasmine Babaei, Nikolay Bashlykov, Soumya Batra, Prajjwal Bhargava, Shruti Bhosale, Dan Bikel, Lukas Blecher, Cristian~Canton Ferrer, Moya Chen, Guillem Cucurull, David Esiobu, Jude Fernandes, Jeremy Fu, Wenyin Fu, Brian Fuller, Cynthia Gao, Vedanuj Goswami, Naman Goyal, Anthony Hartshorn, Saghar Hosseini, Rui Hou, Hakan Inan, Marcin Kardas, Viktor Kerkez, Madian Khabsa, Isabel Kloumann, Artem Korenev, Punit~Singh Koura, Marie-Anne Lachaux, Thibaut Lavril, Jenya Lee, Diana Liskovich, Yinghai Lu, Yuning Mao, Xavier Martinet, Todor Mihaylov, Pushkar Mishra, Igor Molybog, Yixin Nie, Andrew Poulton, Jeremy Reizenstein, Rashi Rungta, Kalyan Saladi, Alan Schelten, Ruan Silva, Eric~Michael Smith, Ranjan Subramanian, Xiaoqing~Ellen Tan, Binh Tang, Ross Taylor, Adina Williams, Jian~Xiang Kuan, Puxin Xu, Zheng Yan, Iliyan Zarov, Yuchen Zhang, Angela Fan, Melanie Kambadur, Sharan Narang, Aurelien Rodriguez, Robert Stojnic, Sergey Edunov, and Thomas
  Scialom.
\newblock Llama 2: Open foundation and fine-tuned chat models, 2023{\natexlab{b}}.

\bibitem[Penedo et~al.(2023)Penedo, Malartic, Hesslow, Cojocaru, Cappelli, Alobeidli, Pannier, Almazrouei, and Launay]{falcon}
Guilherme Penedo, Quentin Malartic, Daniel Hesslow, Ruxandra Cojocaru, Alessandro Cappelli, Hamza Alobeidli, Baptiste Pannier, Ebtesam Almazrouei, and Julien Launay.
\newblock The refinedweb dataset for falcon llm: Outperforming curated corpora with web data, and web data only, 2023.

\bibitem[McCandlish et~al.(2018)McCandlish, Kaplan, Amodei, and Team]{LargeBatchTraining}
Sam McCandlish, Jared Kaplan, Dario Amodei, and OpenAI~Dota Team.
\newblock An empirical model of large-batch training.
\newblock \emph{CoRR}, abs/1812.06162, 2018.
\newblock URL \url{http://arxiv.org/abs/1812.06162}.

\bibitem[Berner et~al.(2019)Berner, Brockman, Chan, Cheung, Debiak, Dennison, Farhi, Fischer, Hashme, Hesse, J{\'{o}}zefowicz, Gray, Olsson, Pachocki, Petrov, de~Oliveira~Pinto, Raiman, Salimans, Schlatter, Schneider, Sidor, Sutskever, Tang, Wolski, and Zhang]{OpenAIDota}
Christopher Berner, Greg Brockman, Brooke Chan, Vicki Cheung, Przemyslaw Debiak, Christy Dennison, David Farhi, Quirin Fischer, Shariq Hashme, Christopher Hesse, Rafal J{\'{o}}zefowicz, Scott Gray, Catherine Olsson, Jakub Pachocki, Michael Petrov, Henrique~Pond{\'{e}} de~Oliveira~Pinto, Jonathan Raiman, Tim Salimans, Jeremy Schlatter, Jonas Schneider, Szymon Sidor, Ilya Sutskever, Jie Tang, Filip Wolski, and Susan Zhang.
\newblock Dota 2 with large scale deep reinforcement learning.
\newblock \emph{CoRR}, abs/1912.06680, 2019.
\newblock URL \url{http://arxiv.org/abs/1912.06680}.

\bibitem[Li et~al.(2020{\natexlab{a}})Li, Zhao, Varma, Salpekar, Noordhuis, Li, Paszke, Smith, Vaughan, Damania, and Chintala]{PyTorchDistributed}
Shen Li, Yanli Zhao, Rohan Varma, Omkar Salpekar, Pieter Noordhuis, Teng Li, Adam Paszke, Jeff Smith, Brian Vaughan, Pritam Damania, and Soumith Chintala.
\newblock Pytorch distributed: Experiences on accelerating data parallel training.
\newblock \emph{Proc. VLDB Endow.}, 13\penalty0 (12):\penalty0 3005–3018, aug 2020{\natexlab{a}}.
\newblock ISSN 2150-8097.
\newblock \doi{10.14778/3415478.3415530}.
\newblock URL \url{https://doi.org/10.14778/3415478.3415530}.

\bibitem[Sergeev and Balso(2018)]{Horovod}
Alexander Sergeev and Mike~Del Balso.
\newblock Horovod: fast and easy distributed deep learning in tensorflow.
\newblock \emph{CoRR}, abs/1802.05799, 2018.
\newblock URL \url{http://arxiv.org/abs/1802.05799}.

\bibitem[Shoeybi et~al.(2019)Shoeybi, Patwary, Puri, LeGresley, Casper, and Catanzaro]{ModelParallelism}
Mohammad Shoeybi, Mostofa Patwary, Raul Puri, Patrick LeGresley, Jared Casper, and Bryan Catanzaro.
\newblock Megatron-lm: Training multi-billion parameter language models using model parallelism.
\newblock \emph{CoRR}, abs/1909.08053, 2019.
\newblock URL \url{http://arxiv.org/abs/1909.08053}.

\bibitem[Shazeer et~al.(2018)Shazeer, Cheng, Parmar, Tran, Vaswani, Koanantakool, Hawkins, Lee, Hong, Young, Sepassi, and Hechtman]{MeshTensorflow}
Noam Shazeer, Youlong Cheng, Niki Parmar, Dustin Tran, Ashish Vaswani, Penporn Koanantakool, Peter Hawkins, HyoukJoong Lee, Mingsheng Hong, Cliff Young, Ryan Sepassi, and Blake~A. Hechtman.
\newblock Mesh-tensorflow: Deep learning for supercomputers.
\newblock In Samy Bengio, Hanna~M. Wallach, Hugo Larochelle, Kristen Grauman, Nicol{\`{o}} Cesa{-}Bianchi, and Roman Garnett, editors, \emph{Advances in Neural Information Processing Systems 31: Annual Conference on Neural Information Processing Systems 2018, NeurIPS 2018, December 3-8, 2018, Montr{\'{e}}al, Canada}, pages 10435--10444, 2018.
\newblock URL \url{https://proceedings.neurips.cc/paper/2018/hash/3a37abdeefe1dab1b30f7c5c7e581b93-Abstract.html}.

\bibitem[Rajbhandari et~al.(2020)Rajbhandari, Rasley, Ruwase, and He]{FSDP_ZeRO}
Samyam Rajbhandari, Jeff Rasley, Olatunji Ruwase, and Yuxiong He.
\newblock Zero: memory optimizations toward training trillion parameter models.
\newblock In Christine Cuicchi, Irene Qualters, and William~T. Kramer, editors, \emph{Proceedings of the International Conference for High Performance Computing, Networking, Storage and Analysis, {SC} 2020, Virtual Event / Atlanta, Georgia, USA, November 9-19, 2020}, page~20. {IEEE/ACM}, 2020.
\newblock \doi{10.1109/SC41405.2020.00024}.
\newblock URL \url{https://doi.org/10.1109/SC41405.2020.00024}.

\bibitem[Zhao et~al.(2023{\natexlab{a}})Zhao, Gu, Varma, Luo, Huang, Xu, Wright, Shojanazeri, Ott, Shleifer, Desmaison, Balioglu, Damania, Nguyen, Chauhan, Hao, Mathews, and Li]{FSDP_Pytorch}
Yanli Zhao, Andrew Gu, Rohan Varma, Liang Luo, Chien{-}Chin Huang, Min Xu, Less Wright, Hamid Shojanazeri, Myle Ott, Sam Shleifer, Alban Desmaison, Can Balioglu, Pritam Damania, Bernard Nguyen, Geeta Chauhan, Yuchen Hao, Ajit Mathews, and Shen Li.
\newblock Pytorch {FSDP:} experiences on scaling fully sharded data parallel.
\newblock \emph{Proc. {VLDB} Endow.}, 16\penalty0 (12):\penalty0 3848--3860, 2023{\natexlab{a}}.
\newblock \doi{10.14778/3611540.3611569}.
\newblock URL \url{https://www.vldb.org/pvldb/vol16/p3848-huang.pdf}.

\bibitem[Chen et~al.(2016)Chen, Xu, Zhang, and Guestrin]{ActivationCheckpointing}
Tianqi Chen, Bing Xu, Chiyuan Zhang, and Carlos Guestrin.
\newblock Training deep nets with sublinear memory cost.
\newblock \emph{CoRR}, abs/1604.06174, 2016.
\newblock URL \url{http://arxiv.org/abs/1604.06174}.

\bibitem[Ren et~al.(2021)Ren, Rajbhandari, Aminabadi, Ruwase, Yang, Zhang, Li, and He]{ZeroOffload}
Jie Ren, Samyam Rajbhandari, Reza~Yazdani Aminabadi, Olatunji Ruwase, Shuangyan Yang, Minjia Zhang, Dong Li, and Yuxiong He.
\newblock Zero-offload: Democratizing billion-scale model training.
\newblock In Irina Calciu and Geoff Kuenning, editors, \emph{2021 {USENIX} Annual Technical Conference, {USENIX} {ATC} 2021, July 14-16, 2021}, pages 551--564. {USENIX} Association, 2021.
\newblock URL \url{https://www.usenix.org/conference/atc21/presentation/ren-jie}.

\bibitem[Lu et~al.(2023)Lu, Wang, and Wei]{MLForSyntheticData}
Yingzhou Lu, Huazheng Wang, and Wenqi Wei.
\newblock Machine learning for synthetic data generation: a review.
\newblock \emph{CoRR}, abs/2302.04062, 2023.
\newblock \doi{10.48550/ARXIV.2302.04062}.
\newblock URL \url{https://doi.org/10.48550/arXiv.2302.04062}.

\bibitem[Javaness(2023)]{llmcost}
La~Javaness.
\newblock \url{https://lajavaness.medium.com/llm-large-language-model-cost-analysis-d5022bb43e9e}, 2023.

\bibitem[Ma et~al.(2024)Ma, Wang, Ma, Wang, Wang, Huang, Dong, Wang, Xue, and Wei]{Quantized1bitLLMs}
Shuming Ma, Hongyu Wang, Lingxiao Ma, Lei Wang, Wenhui Wang, Shaohan Huang, Li~Dong, Ruiping Wang, Jilong Xue, and Furu Wei.
\newblock The era of 1-bit llms: All large language models are in 1.58 bits.
\newblock \emph{CoRR}, abs/2402.17764, 2024.
\newblock \doi{10.48550/ARXIV.2402.17764}.
\newblock URL \url{https://doi.org/10.48550/arXiv.2402.17764}.

\bibitem[Hu et~al.(2022)Hu, Shen, Wallis, Allen{-}Zhu, Li, Wang, Wang, and Chen]{hu2021lora}
Edward~J. Hu, Yelong Shen, Phillip Wallis, Zeyuan Allen{-}Zhu, Yuanzhi Li, Shean Wang, Lu~Wang, and Weizhu Chen.
\newblock Lora: Low-rank adaptation of large language models.
\newblock In \emph{The Tenth International Conference on Learning Representations, {ICLR} 2022, Virtual Event, April 25-29, 2022}. OpenReview.net, 2022.
\newblock URL \url{https://openreview.net/forum?id=nZeVKeeFYf9}.

\bibitem[Borzunov et~al.(2023)Borzunov, Ryabinin, Chumachenko, Baranchuk, Dettmers, Belkada, Samygin, and Raffel]{DistributedInferenceAndFineTuningLargeModelsOverTheInternet}
Alexander Borzunov, Max Ryabinin, Artem Chumachenko, Dmitry Baranchuk, Tim Dettmers, Younes Belkada, Pavel Samygin, and Colin Raffel.
\newblock Distributed inference and fine-tuning of large language models over the internet, 2023.

\bibitem[Huo et~al.(2020)Huo, Yang, Gu, Carin, and Huang]{FedMOM}
Zhouyuan Huo, Qian Yang, Bin Gu, Lawrence Carin, and Heng Huang.
\newblock Faster on-device training using new federated momentum algorithm.
\newblock \emph{CoRR}, abs/2002.02090, 2020.
\newblock URL \url{https://arxiv.org/abs/2002.02090}.

\bibitem[McMahan et~al.(2018)McMahan, Ramage, Talwar, and Zhang]{brendan2018learning}
H.~Brendan McMahan, Daniel Ramage, Kunal Talwar, and Li~Zhang.
\newblock Learning differentially private recurrent language models.
\newblock In \emph{6th International Conference on Learning Representations, {ICLR} 2018, Vancouver, BC, Canada, April 30 - May 3, 2018, Conference Track Proceedings}. OpenReview.net, 2018.
\newblock URL \url{https://openreview.net/forum?id=BJ0hF1Z0b}.

\bibitem[Bonawitz et~al.(2016{\natexlab{a}})Bonawitz, Ivanov, Kreuter, Marcedone, McMahan, Patel, Ramage, Segal, and Seth]{45808}
K.~A. Bonawitz, Vladimir Ivanov, Ben Kreuter, Antonio Marcedone, H.~Brendan McMahan, Sarvar Patel, Daniel Ramage, Aaron Segal, and Karn Seth.
\newblock Practical secure aggregation for federated learning on user-held data.
\newblock In \emph{NIPS Workshop on Private Multi-Party Machine Learning}, 2016{\natexlab{a}}.
\newblock URL \url{https://arxiv.org/abs/1611.04482}.

\bibitem[Kairouz et~al.(2021)Kairouz, McMahan, Avent, Bellet, Bennis, Bhagoji, Bonawitz, Charles, Cormode, Cummings, D'Oliveira, Eichner, Rouayheb, Evans, Gardner, Garrett, Gasc{\'{o}}n, Ghazi, Gibbons, Gruteser, Harchaoui, He, He, Huo, Hutchinson, Hsu, Jaggi, Javidi, Joshi, Khodak, Kone{\v{c}}n{\'y}, Korolova, Koushanfar, Koyejo, Lepoint, Liu, Mittal, Mohri, Nock, {\"{O}}zg{\"{u}}r, Pagh, Qi, Ramage, Raskar, Raykova, Song, Song, Stich, Sun, Suresh, Tram{\`{e}}r, Vepakomma, Wang, Xiong, Xu, Yang, Yu, Yu, and Zhao]{AdancesAndOpenProblems}
Peter Kairouz, H.~Brendan McMahan, Brendan Avent, Aur{\'{e}}lien Bellet, Mehdi Bennis, Arjun~Nitin Bhagoji, Kallista~A. Bonawitz, Zachary Charles, Graham Cormode, Rachel Cummings, Rafael G.~L. D'Oliveira, Hubert Eichner, Salim~El Rouayheb, David Evans, Josh Gardner, Zachary Garrett, Adri{\`{a}} Gasc{\'{o}}n, Badih Ghazi, Phillip~B. Gibbons, Marco Gruteser, Za{\"{\i}}d Harchaoui, Chaoyang He, Lie He, Zhouyuan Huo, Ben Hutchinson, Justin Hsu, Martin Jaggi, Tara Javidi, Gauri Joshi, Mikhail Khodak, Jakub Kone{\v{c}}n{\'y}, Aleksandra Korolova, Farinaz Koushanfar, Sanmi Koyejo, Tancr{\`{e}}de Lepoint, Yang Liu, Prateek Mittal, Mehryar Mohri, Richard Nock, Ayfer {\"{O}}zg{\"{u}}r, Rasmus Pagh, Hang Qi, Daniel Ramage, Ramesh Raskar, Mariana Raykova, Dawn Song, Weikang Song, Sebastian~U. Stich, Ziteng Sun, Ananda~Theertha Suresh, Florian Tram{\`{e}}r, Praneeth Vepakomma, Jianyu Wang, Li~Xiong, Zheng Xu, Qiang Yang, Felix~X. Yu, Han Yu, and Sen Zhao.
\newblock Advances and open problems in federated learning.
\newblock \emph{Found. Trends Mach. Learn.}, 14\penalty0 (1-2):\penalty0 1--210, 2021.
\newblock \doi{10.1561/2200000083}.
\newblock URL \url{https://doi.org/10.1561/2200000083}.

\bibitem[Bonawitz et~al.(2019)Bonawitz, Eichner, Grieskamp, Huba, Ingerman, Ivanov, Kiddon, Kone{\v{c}}n{\'y}, Mazzocchi, McMahan, Overveldt, Petrou, Ramage, and Roselander]{ScaleAndSystemDesign}
Kallista~A. Bonawitz, Hubert Eichner, Wolfgang Grieskamp, Dzmitry Huba, Alex Ingerman, Vladimir Ivanov, Chlo{\'{e}} Kiddon, Jakub Kone{\v{c}}n{\'y}, Stefano Mazzocchi, Brendan McMahan, Timon~Van Overveldt, David Petrou, Daniel Ramage, and Jason Roselander.
\newblock Towards federated learning at scale: System design.
\newblock In Ameet Talwalkar, Virginia Smith, and Matei Zaharia, editors, \emph{Proceedings of Machine Learning and Systems 2019, MLSys 2019, Stanford, CA, USA, March 31 - April 2, 2019}. mlsys.org, 2019.
\newblock URL \url{https://proceedings.mlsys.org/book/271.pdf}.

\bibitem[Ortiz et~al.(2021)Ortiz, Frankle, Rabbat, Morcos, and Ballas]{LocalSGD_Trade_Offs_At_Scale}
Jose Javier~Gonzalez Ortiz, Jonathan Frankle, Mike Rabbat, Ari~S. Morcos, and Nicolas Ballas.
\newblock Trade-offs of local {SGD} at scale: An empirical study.
\newblock \emph{CoRR}, abs/2110.08133, 2021.
\newblock URL \url{https://arxiv.org/abs/2110.08133}.

\bibitem[Hilmkil et~al.(2021)Hilmkil, Callh, Barbieri, S{\"{u}}tfeld, Zec, and Mogren]{hilmkil2021scaling}
Agrin Hilmkil, Sebastian Callh, Matteo Barbieri, Leon~Ren{\'{e}} S{\"{u}}tfeld, Edvin~Listo Zec, and Olof Mogren.
\newblock Scaling federated learning for fine-tuning of large language models.
\newblock In Elisabeth M{\'{e}}tais, Farid Meziane, Helmut Horacek, and Epaminondas Kapetanios, editors, \emph{Natural Language Processing and Information Systems - 26th International Conference on Applications of Natural Language to Information Systems, {NLDB} 2021, Saarbr{\"{u}}cken, Germany, June 23-25, 2021, Proceedings}, volume 12801 of \emph{Lecture Notes in Computer Science}, pages 15--23. Springer, 2021.
\newblock \doi{10.1007/978-3-030-80599-9\_2}.
\newblock URL \url{https://doi.org/10.1007/978-3-030-80599-9\_2}.

\bibitem[Lan et~al.(2020)Lan, Chen, Goodman, Gimpel, Sharma, and Soricut]{ALBERT}
Zhenzhong Lan, Mingda Chen, Sebastian Goodman, Kevin Gimpel, Piyush Sharma, and Radu Soricut.
\newblock {ALBERT:} {A} lite {BERT} for self-supervised learning of language representations.
\newblock In \emph{8th International Conference on Learning Representations, {ICLR} 2020, Addis Ababa, Ethiopia, April 26-30, 2020}. OpenReview.net, 2020.
\newblock URL \url{https://openreview.net/forum?id=H1eA7AEtvS}.

\bibitem[Devlin et~al.(2019)Devlin, Chang, Lee, and Toutanova]{BERT}
Jacob Devlin, Ming{-}Wei Chang, Kenton Lee, and Kristina Toutanova.
\newblock {BERT:} pre-training of deep bidirectional transformers for language understanding.
\newblock In Jill Burstein, Christy Doran, and Thamar Solorio, editors, \emph{Proceedings of the 2019 Conference of the North American Chapter of the Association for Computational Linguistics: Human Language Technologies, {NAACL-HLT} 2019, Minneapolis, MN, USA, June 2-7, 2019, Volume 1 (Long and Short Papers)}, pages 4171--4186. Association for Computational Linguistics, 2019.
\newblock \doi{10.18653/V1/N19-1423}.
\newblock URL \url{https://doi.org/10.18653/v1/n19-1423}.

\bibitem[Riedel et~al.(2023)Riedel, Reichert, von Schwerin, Hafner, Schaudt, and Singh]{riedel2023}
Pascal Riedel, Manfred Reichert, Reinhold von Schwerin, Alexander Hafner, Daniel Schaudt, and Gaurav Singh.
\newblock Performance analysis of federated learning algorithms for multilingual protest news detection using pre-trained distilbert and {BERT}.
\newblock \emph{{IEEE} Access}, 11:\penalty0 134009--134022, 2023.
\newblock \doi{10.1109/ACCESS.2023.3334910}.
\newblock URL \url{https://doi.org/10.1109/ACCESS.2023.3334910}.

\bibitem[Wang et~al.(2023)Wang, Zhang, Cao, Li, McMahan, Oh, Xu, and Zaheer]{wang2024public}
Boxin Wang, Jacky~Yibo Zhang, Yuan Cao, Bo~Li, H.~Brendan McMahan, Sewoong Oh, Zheng Xu, and Manzil Zaheer.
\newblock Can public large language models help private cross-device federated learning?
\newblock \emph{CoRR}, abs/2305.12132, 2023.
\newblock \doi{10.48550/ARXIV.2305.12132}.
\newblock URL \url{https://doi.org/10.48550/arXiv.2305.12132}.

\bibitem[Weller et~al.(2022)Weller, Marone, Braverman, Lawrie, and Durme]{weller2022}
Orion Weller, Marc Marone, Vladimir Braverman, Dawn~J. Lawrie, and Benjamin~Van Durme.
\newblock Pretrained models for multilingual federated learning.
\newblock In Marine Carpuat, Marie{-}Catherine de~Marneffe, and Iv{\'{a}}n Vladimir~Meza Ru{\'{\i}}z, editors, \emph{Proceedings of the 2022 Conference of the North American Chapter of the Association for Computational Linguistics: Human Language Technologies, {NAACL} 2022, Seattle, WA, United States, July 10-15, 2022}, pages 1413--1421. Association for Computational Linguistics, 2022.
\newblock \doi{10.18653/V1/2022.NAACL-MAIN.101}.
\newblock URL \url{https://doi.org/10.18653/v1/2022.naacl-main.101}.

\bibitem[Zhang et~al.(2023)Zhang, Vahidian, Kuo, Li, Zhang, Wang, and Chen]{zhang2024building}
Jianyi Zhang, Saeed Vahidian, Martin Kuo, Chunyuan Li, Ruiyi Zhang, Guoyin Wang, and Yiran Chen.
\newblock Towards building the federated {GPT:} federated instruction tuning.
\newblock \emph{CoRR}, abs/2305.05644, 2023.
\newblock \doi{10.48550/ARXIV.2305.05644}.
\newblock URL \url{https://doi.org/10.48550/arXiv.2305.05644}.

\bibitem[Fan et~al.(2023)Fan, Kang, Ma, Chen, Wei, Fan, and Yang]{fan2023fatellm}
Tao Fan, Yan Kang, Guoqiang Ma, Weijing Chen, Wenbin Wei, Lixin Fan, and Qiang Yang.
\newblock {FATE-LLM:} {A} industrial grade federated learning framework for large language models.
\newblock \emph{CoRR}, abs/2310.10049, 2023.
\newblock \doi{10.48550/ARXIV.2310.10049}.
\newblock URL \url{https://doi.org/10.48550/arXiv.2310.10049}.

\bibitem[Kuang et~al.(2023)Kuang, Qian, Li, Chen, Gao, Pan, Xie, Li, Ding, and Zhou]{kuang2023federatedscopellm}
Weirui Kuang, Bingchen Qian, Zitao Li, Daoyuan Chen, Dawei Gao, Xuchen Pan, Yuexiang Xie, Yaliang Li, Bolin Ding, and Jingren Zhou.
\newblock Federatedscope-llm: {A} comprehensive package for fine-tuning large language models in federated learning.
\newblock \emph{CoRR}, abs/2309.00363, 2023.
\newblock \doi{10.48550/ARXIV.2309.00363}.
\newblock URL \url{https://doi.org/10.48550/arXiv.2309.00363}.

\bibitem[Jiang et~al.(2023)Jiang, Liu, and Fan]{jiang2023lowparameter}
Jingang Jiang, Xiangyang Liu, and Chenyou Fan.
\newblock Low-parameter federated learning with large language models.
\newblock \emph{CoRR}, abs/2307.13896, 2023.
\newblock \doi{10.48550/ARXIV.2307.13896}.
\newblock URL \url{https://doi.org/10.48550/arXiv.2307.13896}.

\bibitem[Malaviya et~al.(2023)Malaviya, Shukla, and Lodha]{pmlr-v232-malaviya23a}
Shubham Malaviya, Manish Shukla, and Sachin Lodha.
\newblock Reducing communication overhead in federated learning for pre-trained language models using parameter-efficient finetuning.
\newblock In Sarath Chandar, Razvan Pascanu, Hanie Sedghi, and Doina Precup, editors, \emph{Conference on Lifelong Learning Agents, 22-25 August 2023, McGill University, Montr{\'{e}}al, Qu{\'{e}}bec, Canada}, volume 232 of \emph{Proceedings of Machine Learning Research}, pages 456--469. {PMLR}, 2023.
\newblock URL \url{https://proceedings.mlr.press/v232/malaviya23a.html}.

\bibitem[Xu et~al.(2023)Xu, Song, Tian, Agrawal, Granqvist, van Dalen, Zhang, Argueta, Han, Deng, Liu, Walia, and Jin]{xu2022training}
Mingbin Xu, Congzheng Song, Ye~Tian, Neha Agrawal, Filip Granqvist, Rogier~C. van Dalen, Xiao Zhang, Arturo Argueta, Shiyi Han, Yaqiao Deng, Leo Liu, Anmol Walia, and Alex Jin.
\newblock Training large-vocabulary neural language models by private federated learning for resource-constrained devices.
\newblock In \emph{{IEEE} International Conference on Acoustics, Speech and Signal Processing {ICASSP} 2023, Rhodes Island, Greece, June 4-10, 2023}, pages 1--5. {IEEE}, 2023.
\newblock \doi{10.1109/ICASSP49357.2023.10096570}.
\newblock URL \url{https://doi.org/10.1109/ICASSP49357.2023.10096570}.

\bibitem[Xu et~al.(2024)Xu, Cai, Wu, Li, and Wang]{xu2024fwdllm}
Mengwei Xu, Dongqi Cai, Yaozong Wu, Xiang Li, and Shangguang Wang.
\newblock Fwdllm: Efficient fedllm using forward gradient, 2024.

\bibitem[Babakniya et~al.(2023)Babakniya, Elkordy, Ezzeldin, Liu, Song, El{-}Khamy, and Avestimehr]{babakniya2023slora}
Sara Babakniya, Ahmed~Roushdy Elkordy, Yahya~H. Ezzeldin, Qingfeng Liu, Kee{-}Bong Song, Mostafa El{-}Khamy, and Salman Avestimehr.
\newblock Slora: Federated parameter efficient fine-tuning of language models.
\newblock \emph{CoRR}, abs/2308.06522, 2023.
\newblock \doi{10.48550/ARXIV.2308.06522}.
\newblock URL \url{https://doi.org/10.48550/arXiv.2308.06522}.

\bibitem[Kim et~al.(2023)Kim, Kim, Mok, Park, and Lee]{kim-etal-2023-client}
Yeachan Kim, Junho Kim, Wing{-}Lam Mok, Jun{-}Hyung Park, and SangKeun Lee.
\newblock Client-customized adaptation for parameter-efficient federated learning.
\newblock In Anna Rogers, Jordan~L. Boyd{-}Graber, and Naoaki Okazaki, editors, \emph{Findings of the Association for Computational Linguistics: {ACL} 2023, Toronto, Canada, July 9-14, 2023}, pages 1159--1172. Association for Computational Linguistics, 2023.
\newblock \doi{10.18653/V1/2023.FINDINGS-ACL.75}.
\newblock URL \url{https://doi.org/10.18653/v1/2023.findings-acl.75}.

\bibitem[Lester et~al.(2021)Lester, Al{-}Rfou, and Constant]{lester2021power}
Brian Lester, Rami Al{-}Rfou, and Noah Constant.
\newblock The power of scale for parameter-efficient prompt tuning.
\newblock In Marie{-}Francine Moens, Xuanjing Huang, Lucia Specia, and Scott~Wen{-}tau Yih, editors, \emph{Proceedings of the 2021 Conference on Empirical Methods in Natural Language Processing, {EMNLP} 2021, Virtual Event / Punta Cana, Dominican Republic, 7-11 November, 2021}, pages 3045--3059. Association for Computational Linguistics, 2021.
\newblock \doi{10.18653/V1/2021.EMNLP-MAIN.243}.
\newblock URL \url{https://doi.org/10.18653/v1/2021.emnlp-main.243}.

\bibitem[Zhao et~al.(2023{\natexlab{b}})Zhao, Du, Li, Li, and Liu]{zhao2023fedprompt}
Haodong Zhao, Wei Du, Fangqi Li, Peixuan Li, and Gongshen Liu.
\newblock Fedprompt: Communication-efficient and privacy-preserving prompt tuning in federated learning.
\newblock In \emph{{IEEE} International Conference on Acoustics, Speech and Signal Processing {ICASSP} 2023, Rhodes Island, Greece, June 4-10, 2023}, pages 1--5. {IEEE}, 2023{\natexlab{b}}.
\newblock \doi{10.1109/ICASSP49357.2023.10095356}.
\newblock URL \url{https://doi.org/10.1109/ICASSP49357.2023.10095356}.

\bibitem[Che et~al.(2023)Che, Liu, Zhou, Ren, Zhou, Sheng, Dai, and Dou]{che2024federated}
Tianshi Che, Ji~Liu, Yang Zhou, Jiaxiang Ren, Jiwen Zhou, Victor~S. Sheng, Huaiyu Dai, and Dejing Dou.
\newblock Federated learning of large language models with parameter-efficient prompt tuning and adaptive optimization.
\newblock In Houda Bouamor, Juan Pino, and Kalika Bali, editors, \emph{Proceedings of the 2023 Conference on Empirical Methods in Natural Language Processing, {EMNLP} 2023, Singapore, December 6-10, 2023}, pages 7871--7888. Association for Computational Linguistics, 2023.
\newblock \doi{10.18653/V1/2023.EMNLP-MAIN.488}.
\newblock URL \url{https://doi.org/10.18653/v1/2023.emnlp-main.488}.

\bibitem[Zhao et~al.(2024)Zhao, Lee, Chen, Qiu, Gao, Fan, and Lane]{FedPromptTuning}
Wanru Zhao, Royson Lee, Yihong Chen, Xinchi Qiu, Yan Gao, Hongxiang Fan, and Nicholas~Donald Lane.
\newblock Breaking physical and linguistic borders: Multilingual federated prompt tuning for low-resource languages.
\newblock In \emph{The Twelfth International Conference on Learning Representations}, 2024.

\bibitem[Patel and Palazzolo(2024)]{OpenAINewsPublishersDeal}
Sahil Patel and Stephanie Palazzolo.
\newblock {OpenAI offers publishers as little as \$1 million a year — the information}, Jan 2024.
\newblock URL \url{https://www.theinformation.com/articles/openai-offers-publishers-as-little-as-1-million-a-year}.

\bibitem[Magueresse et~al.(2020)Magueresse, Carles, and Heetderks]{LowResourceLanguagesSurvey}
Alexandre Magueresse, Vincent Carles, and Evan Heetderks.
\newblock Low-resource languages: {A} review of past work and future challenges.
\newblock \emph{CoRR}, abs/2006.07264, 2020.
\newblock URL \url{https://arxiv.org/abs/2006.07264}.

\bibitem[Ranathunga et~al.(2023)Ranathunga, Lee, Skenduli, Shekhar, Alam, and Kaur]{LowResourceNMTSurvey}
Surangika Ranathunga, En{-}Shiun~Annie Lee, Marjana~Prifti Skenduli, Ravi Shekhar, Mehreen Alam, and Rishemjit Kaur.
\newblock Neural machine translation for low-resource languages: {A} survey.
\newblock \emph{{ACM} Comput. Surv.}, 55\penalty0 (11):\penalty0 229:1--229:37, 2023.
\newblock \doi{10.1145/3567592}.
\newblock URL \url{https://doi.org/10.1145/3567592}.

\bibitem[Bonawitz et~al.(2016{\natexlab{b}})Bonawitz, Ivanov, Kreuter, Marcedone, McMahan, Patel, Ramage, Segal, and Seth]{SecAggOG}
Kallista~A. Bonawitz, Vladimir Ivanov, Ben Kreuter, Antonio Marcedone, H.~Brendan McMahan, Sarvar Patel, Daniel Ramage, Aaron Segal, and Karn Seth.
\newblock Practical secure aggregation for federated learning on user-held data.
\newblock \emph{CoRR}, abs/1611.04482, 2016{\natexlab{b}}.
\newblock URL \url{http://arxiv.org/abs/1611.04482}.

\bibitem[Li et~al.(2020{\natexlab{b}})Li, Sahu, Zaheer, Sanjabi, Talwalkar, and Smith]{FedProx}
Tian Li, Anit~Kumar Sahu, Manzil Zaheer, Maziar Sanjabi, Ameet Talwalkar, and Virginia Smith.
\newblock Federated optimization in heterogeneous networks.
\newblock In Inderjit~S. Dhillon, Dimitris~S. Papailiopoulos, and Vivienne Sze, editors, \emph{Proceedings of Machine Learning and Systems 2020, MLSys 2020, Austin, TX, USA, March 2-4, 2020}. mlsys.org, 2020{\natexlab{b}}.
\newblock URL \url{https://proceedings.mlsys.org/book/316.pdf}.

\bibitem[Reddi et~al.(2021)Reddi, Charles, Zaheer, Garrett, Rush, Kone{\v{c}}n{\'y}, Kumar, and McMahan]{FedOPT}
Sashank~J. Reddi, Zachary Charles, Manzil Zaheer, Zachary Garrett, Keith Rush, Jakub Kone{\v{c}}n{\'y}, Sanjiv Kumar, and Hugh~Brendan McMahan.
\newblock Adaptive federated optimization.
\newblock In \emph{9th International Conference on Learning Representations, {ICLR} 2021, Virtual Event, Austria, May 3-7, 2021}. OpenReview.net, 2021.
\newblock URL \url{https://openreview.net/forum?id=LkFG3lB13U5}.

\bibitem[Andrew et~al.(2021)Andrew, Thakkar, McMahan, and Ramaswamy]{AdaptiveClipping}
Galen Andrew, Om~Thakkar, Brendan McMahan, and Swaroop Ramaswamy.
\newblock Differentially private learning with adaptive clipping.
\newblock In Marc'Aurelio Ranzato, Alina Beygelzimer, Yann~N. Dauphin, Percy Liang, and Jennifer~Wortman Vaughan, editors, \emph{Advances in Neural Information Processing Systems 34: Annual Conference on Neural Information Processing Systems 2021, NeurIPS 2021, December 6-14, 2021, virtual}, pages 17455--17466, 2021.
\newblock URL \url{https://proceedings.neurips.cc/paper/2021/hash/91cff01af640a24e7f9f7a5ab407889f-Abstract.html}.

\bibitem[Deng et~al.(2020)Deng, Li, Han, Shi, and Xie]{ModelCompressionSurvey}
Lei Deng, Guoqi Li, Song Han, Luping Shi, and Yuan Xie.
\newblock Model compression and hardware acceleration for neural networks: {A} comprehensive survey.
\newblock \emph{Proc. {IEEE}}, 108\penalty0 (4):\penalty0 485--532, 2020.
\newblock \doi{10.1109/JPROC.2020.2976475}.
\newblock URL \url{https://doi.org/10.1109/JPROC.2020.2976475}.

\bibitem[Raffel et~al.(2020)Raffel, Shazeer, Roberts, Lee, Narang, Matena, Zhou, Li, and Liu]{C4}
Colin Raffel, Noam Shazeer, Adam Roberts, Katherine Lee, Sharan Narang, Michael Matena, Yanqi Zhou, Wei Li, and Peter~J. Liu.
\newblock Exploring the limits of transfer learning with a unified text-to-text transformer.
\newblock \emph{J. Mach. Learn. Res.}, 21:\penalty0 140:1--140:67, 2020.
\newblock URL \url{http://jmlr.org/papers/v21/20-074.html}.

\bibitem[Li et~al.(2020{\natexlab{c}})Li, Sanjabi, Beirami, and Smith]{QFedAvg}
Tian Li, Maziar Sanjabi, Ahmad Beirami, and Virginia Smith.
\newblock Fair resource allocation in federated learning.
\newblock In \emph{8th International Conference on Learning Representations, {ICLR} 2020, Addis Ababa, Ethiopia, April 26-30, 2020}. OpenReview.net, 2020{\natexlab{c}}.
\newblock URL \url{https://openreview.net/forum?id=ByexElSYDr}.

\bibitem[Li et~al.(2021)Li, Beirami, Sanjabi, and Smith]{TERM}
Tian Li, Ahmad Beirami, Maziar Sanjabi, and Virginia Smith.
\newblock Tilted empirical risk minimization.
\newblock In \emph{9th International Conference on Learning Representations, {ICLR} 2021, Virtual Event, Austria, May 3-7, 2021}. OpenReview.net, 2021.
\newblock URL \url{https://openreview.net/forum?id=K5YasWXZT3O}.

\bibitem[Zhou et~al.(2024)Zhou, Lin, Zhang, and Tsang]{Zhou_2024}
Tailin Zhou, Zehong Lin, Jun Zhang, and Danny~H.K. Tsang.
\newblock Understanding and improving model averaging in federated learning on heterogeneous data.
\newblock \emph{IEEE Transactions on Mobile Computing}, page 1–16, 2024.
\newblock ISSN 2161-9875.
\newblock \doi{10.1109/tmc.2024.3406554}.
\newblock URL \url{http://dx.doi.org/10.1109/TMC.2024.3406554}.

\bibitem[Pfeiffer et~al.(2023)Pfeiffer, Rapp, Khalili, and Henkel]{fl_constrained_survey}
Kilian Pfeiffer, Martin Rapp, Ramin Khalili, and J{\"{o}}rg Henkel.
\newblock Federated learning for computationally constrained heterogeneous devices: {A} survey.
\newblock \emph{{ACM} Comput. Surv.}, 55\penalty0 (14s):\penalty0 334:1--334:27, 2023.
\newblock \doi{10.1145/3596907}.
\newblock URL \url{https://doi.org/10.1145/3596907}.

\bibitem[Charles et~al.(2023)Charles, Mitchell, Pillutla, Reneer, and Garrett]{DatasetGrouper}
Zachary Charles, Nicole Mitchell, Krishna Pillutla, Michael Reneer, and Zachary Garrett.
\newblock Towards federated foundation models: Scalable dataset pipelines for group-structured learning.
\newblock In Alice Oh, Tristan Naumann, Amir Globerson, Kate Saenko, Moritz Hardt, and Sergey Levine, editors, \emph{Advances in Neural Information Processing Systems 36: Annual Conference on Neural Information Processing Systems 2023, NeurIPS 2023, New Orleans, LA, USA, December 10 - 16, 2023}, 2023.
\newblock URL \url{http://papers.nips.cc/paper\_files/paper/2023/hash/662bb9c4dcc96aeaac8e7cd3fc6a0add-Abstract-Datasets\_and\_Benchmarks.html}.

\bibitem[Charles et~al.(2021)Charles, Garrett, Huo, Shmulyian, and Smith]{OnLargeCohortTraining}
Zachary Charles, Zachary Garrett, Zhouyuan Huo, Sergei Shmulyian, and Virginia Smith.
\newblock On large-cohort training for federated learning.
\newblock In Marc'Aurelio Ranzato, Alina Beygelzimer, Yann~N. Dauphin, Percy Liang, and Jennifer~Wortman Vaughan, editors, \emph{Advances in Neural Information Processing Systems 34: Annual Conference on Neural Information Processing Systems 2021, NeurIPS 2021, December 6-14, 2021, virtual}, pages 20461--20475, 2021.
\newblock URL \url{https://proceedings.neurips.cc/paper/2021/hash/ab9ebd57177b5106ad7879f0896685d4-Abstract.html}.

\bibitem[Vural et~al.(2017)Vural, Koyuncu, and Guney]{SystematicLitReviewMicroServices}
Hulya Vural, Murat Koyuncu, and Sinem Guney.
\newblock A systematic literature review on microservices.
\newblock In Osvaldo Gervasi, Beniamino Murgante, Sanjay Misra, Giuseppe Borruso, Carmelo~Maria Torre, Ana Maria A.~C. Rocha, David Taniar, Bernady~O. Apduhan, Elena~N. Stankova, and Alfredo Cuzzocrea, editors, \emph{Computational Science and Its Applications - {ICCSA} 2017 - 17th International Conference, Trieste, Italy, July 3-6, 2017, Proceedings, Part {VI}}, volume 10409 of \emph{Lecture Notes in Computer Science}, pages 203--217. Springer, 2017.
\newblock \doi{10.1007/978-3-319-62407-5\_14}.
\newblock URL \url{https://doi.org/10.1007/978-3-319-62407-5\_14}.

\bibitem[Paszke et~al.(2019)Paszke, Gross, Massa, Lerer, Bradbury, Chanan, Killeen, Lin, Gimelshein, Antiga, Desmaison, K{\"{o}}pf, Yang, DeVito, Raison, Tejani, Chilamkurthy, Steiner, Fang, Bai, and Chintala]{PyTorch}
Adam Paszke, Sam Gross, Francisco Massa, Adam Lerer, James Bradbury, Gregory Chanan, Trevor Killeen, Zeming Lin, Natalia Gimelshein, Luca Antiga, Alban Desmaison, Andreas K{\"{o}}pf, Edward~Z. Yang, Zachary DeVito, Martin Raison, Alykhan Tejani, Sasank Chilamkurthy, Benoit Steiner, Lu~Fang, Junjie Bai, and Soumith Chintala.
\newblock Pytorch: An imperative style, high-performance deep learning library.
\newblock In Hanna~M. Wallach, Hugo Larochelle, Alina Beygelzimer, Florence d'Alch{\'{e}}{-}Buc, Emily~B. Fox, and Roman Garnett, editors, \emph{Advances in Neural Information Processing Systems 32: Annual Conference on Neural Information Processing Systems 2019, NeurIPS 2019, December 8-14, 2019, Vancouver, BC, Canada}, pages 8024--8035, 2019.
\newblock URL \url{https://proceedings.neurips.cc/paper/2019/hash/bdbca288fee7f92f2bfa9f7012727740-Abstract.html}.

\bibitem[Yadan(2019)]{hydra}
Omry Yadan.
\newblock Hydra - a framework for elegantly configuring complex applications.
\newblock Github, 2019.
\newblock URL \url{https://github.com/facebookresearch/hydra}.

\bibitem[MinIO(2024)]{minio}
Inc. MinIO.
\newblock {The Object Store for AI Data Infrastructure}, 2024.
\newblock URL \url{https://min.io/}.

\bibitem[Amazon(2024)]{s3}
Amazon.
\newblock {Amazon S3}, 2024.
\newblock URL \url{https://aws.amazon.com/s3/}.

\bibitem[the~boto project(2024)]{boto3}
the~boto project.
\newblock {Boto3 - The AWS SDK for Python}, 2024.
\newblock URL \url{https://github.com/boto/boto3}.

\bibitem[Faisal et~al.(2022)Faisal, Wang, and Anastasopoulos]{DatasetGeography}
Fahim Faisal, Yinkai Wang, and Antonios Anastasopoulos.
\newblock Dataset geography: Mapping language data to language users.
\newblock In Smaranda Muresan, Preslav Nakov, and Aline Villavicencio, editors, \emph{Proceedings of the 60th Annual Meeting of the Association for Computational Linguistics (Volume 1: Long Papers), {ACL} 2022, Dublin, Ireland, May 22-27, 2022}, pages 3381--3411. Association for Computational Linguistics, 2022.
\newblock \doi{10.18653/V1/2022.ACL-LONG.239}.
\newblock URL \url{https://doi.org/10.18653/v1/2022.acl-long.239}.

\bibitem[Zhang et~al.(2022)Zhang, Roller, Goyal, Artetxe, Chen, Chen, Dewan, Diab, Li, Lin, Mihaylov, Ott, Shleifer, Shuster, Simig, Koura, Sridhar, Wang, and Zettlemoyer]{meta_opt}
Susan Zhang, Stephen Roller, Naman Goyal, Mikel Artetxe, Moya Chen, Shuohui Chen, Christopher Dewan, Mona~T. Diab, Xian Li, Xi~Victoria Lin, Todor Mihaylov, Myle Ott, Sam Shleifer, Kurt Shuster, Daniel Simig, Punit~Singh Koura, Anjali Sridhar, Tianlu Wang, and Luke Zettlemoyer.
\newblock {OPT:} open pre-trained transformer language models.
\newblock \emph{CoRR}, abs/2205.01068, 2022.
\newblock \doi{10.48550/ARXIV.2205.01068}.
\newblock URL \url{https://doi.org/10.48550/arXiv.2205.01068}.

\bibitem[Databricks(2024)]{mosaicml}
Databricks.
\newblock {mosaic research}, 2024.
\newblock URL \url{https://www.databricks.com/research/mosaic}.

\bibitem[Team(2023)]{mpt_blogpost}
MosaicML~NLP Team.
\newblock {Introducing MPT-7B: A New Standard for Open-Source, Commercially Usable LLMs}, 2023.
\newblock URL \url{https://www.databricks.com/blog/mpt-7b}.

\bibitem[Dao et~al.(2022)Dao, Fu, Ermon, Rudra, and R{\'e}]{flashattention}
Tri Dao, Daniel~Y. Fu, Stefano Ermon, Atri Rudra, and Christopher R{\'e}.
\newblock Flash{A}ttention: Fast and memory-efficient exact attention with {IO}-awareness.
\newblock In \emph{Advances in Neural Information Processing Systems}, 2022.

\bibitem[Press et~al.(2022)Press, Smith, and Lewis]{alibi}
Ofir Press, Noah Smith, and Mike Lewis.
\newblock Train short, test long: Attention with linear biases enables input length extrapolation.
\newblock In \emph{International Conference on Learning Representations}, 2022.
\newblock URL \url{https://openreview.net/forum?id=R8sQPpGCv0}.

\bibitem[Gao et~al.(2021)Gao, Biderman, Black, Golding, Hoppe, Foster, Phang, He, Thite, Nabeshima, Presser, and Leahy]{ThePile}
Leo Gao, Stella Biderman, Sid Black, Laurence Golding, Travis Hoppe, Charles Foster, Jason Phang, Horace He, Anish Thite, Noa Nabeshima, Shawn Presser, and Connor Leahy.
\newblock The pile: An 800gb dataset of diverse text for language modeling.
\newblock \emph{CoRR}, abs/2101.00027, 2021.
\newblock URL \url{https://arxiv.org/abs/2101.00027}.

\bibitem[Xue et~al.(2021)Xue, Constant, Roberts, Kale, Al{-}Rfou, Siddhant, Barua, and Raffel]{mC4}
Linting Xue, Noah Constant, Adam Roberts, Mihir Kale, Rami Al{-}Rfou, Aditya Siddhant, Aditya Barua, and Colin Raffel.
\newblock mt5: {A} massively multilingual pre-trained text-to-text transformer.
\newblock In Kristina Toutanova, Anna Rumshisky, Luke Zettlemoyer, Dilek Hakkani{-}T{\"{u}}r, Iz~Beltagy, Steven Bethard, Ryan Cotterell, Tanmoy Chakraborty, and Yichao Zhou, editors, \emph{Proceedings of the 2021 Conference of the North American Chapter of the Association for Computational Linguistics: Human Language Technologies, {NAACL-HLT} 2021, Online, June 6-11, 2021}, pages 483--498. Association for Computational Linguistics, 2021.
\newblock \doi{10.18653/V1/2021.NAACL-MAIN.41}.
\newblock URL \url{https://doi.org/10.18653/v1/2021.naacl-main.41}.

\bibitem[Sardana et~al.(2024)Sardana, Portes, Doubov, and Frankle]{BeyondChinchilaOptimal}
Nikhil Sardana, Jacob Portes, Sasha Doubov, and Jonathan Frankle.
\newblock Beyond chinchilla-optimal: Accounting for inference in language model scaling laws.
\newblock In \emph{{ICML}}. OpenReview.net, 2024.

\bibitem[He et~al.(2024)He, Zhuang, and Wu]{he2024exploringscalinglawslocal}
Qiaozhi He, Xiaomin Zhuang, and Zhihua Wu.
\newblock Exploring scaling laws for local sgd in large language model training, 2024.
\newblock URL \url{https://arxiv.org/abs/2409.13198}.

\bibitem[Black et~al.(2022)Black, Biderman, Hallahan, Anthony, Gao, Golding, He, Leahy, McDonell, Phang, Pieler, Prashanth, Purohit, Reynolds, Tow, Wang, and Weinbach]{eleuther_ai_tokenizer}
Sid Black, Stella Biderman, Eric Hallahan, Quentin Anthony, Leo Gao, Laurence Golding, Horace He, Connor Leahy, Kyle McDonell, Jason Phang, Michael Pieler, USVSN~Sai Prashanth, Shivanshu Purohit, Laria Reynolds, Jonathan Tow, Ben Wang, and Samuel Weinbach.
\newblock Gpt-neox-20b: An open-source autoregressive language model.
\newblock \emph{CoRR}, abs/2204.06745, 2022.
\newblock \doi{10.48550/ARXIV.2204.06745}.
\newblock URL \url{https://doi.org/10.48550/arXiv.2204.06745}.

\bibitem[Loshchilov and Hutter(2019)]{AdamW}
Ilya Loshchilov and Frank Hutter.
\newblock Decoupled weight decay regularization.
\newblock In \emph{7th International Conference on Learning Representations, {ICLR} 2019, New Orleans, LA, USA, May 6-9, 2019}. OpenReview.net, 2019.
\newblock URL \url{https://openreview.net/forum?id=Bkg6RiCqY7}.

\bibitem[Nichol et~al.(2018)Nichol, Achiam, and Schulman]{REPTILE}
Alex Nichol, Joshua Achiam, and John Schulman.
\newblock On first-order meta-learning algorithms.
\newblock \emph{CoRR}, abs/1803.02999, 2018.
\newblock URL \url{http://arxiv.org/abs/1803.02999}.

\bibitem[Fallah et~al.(2020)Fallah, Mokhtari, and Ozdaglar]{PflModelAgnosticMetaLearning}
Alireza Fallah, Aryan Mokhtari, and Asuman~E. Ozdaglar.
\newblock Personalized federated learning with theoretical guarantees: {A} model-agnostic meta-learning approach.
\newblock In Hugo Larochelle, Marc'Aurelio Ranzato, Raia Hadsell, Maria{-}Florina Balcan, and Hsuan{-}Tien Lin, editors, \emph{Advances in Neural Information Processing Systems 33: Annual Conference on Neural Information Processing Systems 2020, NeurIPS 2020, December 6-12, 2020, virtual}, 2020.
\newblock URL \url{https://proceedings.neurips.cc/paper/2020/hash/24389bfe4fe2eba8bf9aa9203a44cdad-Abstract.html}.

\bibitem[Lee et~al.(2023)Lee, Kim, Li, Qiu, Hospedales, Huszar, and Lane]{lee2024fedl2p}
Royson Lee, Minyoung Kim, Da~Li, Xinchi Qiu, Timothy~M. Hospedales, Ferenc Huszar, and Nicholas~D. Lane.
\newblock Fedl2p: Federated learning to personalize.
\newblock In Alice Oh, Tristan Naumann, Amir Globerson, Kate Saenko, Moritz Hardt, and Sergey Levine, editors, \emph{Advances in Neural Information Processing Systems 36: Annual Conference on Neural Information Processing Systems 2023, NeurIPS 2023, New Orleans, LA, USA, December 10 - 16, 2023}, 2023.
\newblock URL \url{http://papers.nips.cc/paper\_files/paper/2023/hash/2fb57276bfbaf1b832d7bfcba36bb41c-Abstract-Conference.html}.

\bibitem[Clark et~al.(2018)Clark, Cowhey, Etzioni, Khot, Sabharwal, Schoenick, and Tafjord]{arc_challenge}
Peter Clark, Isaac Cowhey, Oren Etzioni, Tushar Khot, Ashish Sabharwal, Carissa Schoenick, and Oyvind Tafjord.
\newblock Think you have solved question answering? try arc, the {AI2} reasoning challenge.
\newblock \emph{CoRR}, abs/1803.05457, 2018.
\newblock URL \url{http://arxiv.org/abs/1803.05457}.

\bibitem[Srivastava et~al.(2023)Srivastava, Rastogi, Rao, Shoeb, Abid, Fisch, Brown, Santoro, Gupta, Garriga{-}Alonso, Kluska, Lewkowycz, Agarwal, Power, Ray, Warstadt, Kocurek, Safaya, Tazarv, Xiang, Parrish, Nie, Hussain, Askell, Dsouza, Slone, Rahane, Iyer, Andreassen, Madotto, Santilli, Stuhlm{\"{u}}ller, Dai, La, Lampinen, Zou, Jiang, Chen, Vuong, Gupta, Gottardi, Norelli, Venkatesh, Gholamidavoodi, Tabassum, Menezes, Kirubarajan, Mullokandov, Sabharwal, Herrick, Efrat, Erdem, Karakas, Roberts, Loe, Zoph, Bojanowski, {\"{O}}zyurt, Hedayatnia, Neyshabur, Inden, Stein, Ekmekci, Lin, Howald, Orinion, Diao, Dour, Stinson, Argueta, Ram{\'{\i}}rez, Singh, Rathkopf, Meng, Baral, Wu, Callison{-}Burch, Waites, Voigt, Manning, Potts, Ramirez, Rivera, Siro, Raffel, Ashcraft, Garbacea, Sileo, Garrette, Hendrycks, Kilman, Roth, Freeman, Khashabi, Levy, Gonz{\'{a}}lez, Perszyk, Hernandez, Chen, Ippolito, Gilboa, Dohan, Drakard, Jurgens, Datta, Ganguli, Emelin, Kleyko, Yuret, Chen, Tam, Hupkes, Misra, Buzan, Mollo,
  Yang, Lee, Schrader, Shutova, Cubuk, Segal, Hagerman, Barnes, Donoway, Pavlick, Rodol{\`{a}}, Lam, Chu, Tang, Erdem, Chang, Chi, Dyer, Jerzak, Kim, Manyasi, Zheltonozhskii, Xia, Siar, Mart{\'{\i}}nez{-}Plumed, Happ{\'{e}}, Chollet, Rong, Mishra, Winata, de~Melo, Kruszewski, Parascandolo, Mariani, Wang, Jaimovitch{-}L{\'{o}}pez, Betz, Gur{-}Ari, Galijasevic, Kim, Rashkin, Hajishirzi, Mehta, Bogar, Shevlin, Sch{\"{u}}tze, Yakura, Zhang, Wong, Ng, Noble, Jumelet, Geissinger, Kernion, Hilton, Lee, Fisac, Simon, Koppel, Zheng, Zou, Kocon, Thompson, Wingfield, Kaplan, Radom, Sohl{-}Dickstein, Phang, Wei, Yosinski, Novikova, Bosscher, Marsh, Kim, Taal, Engel, Alabi, Xu, Song, Tang, Waweru, Burden, Miller, Balis, Batchelder, Berant, Frohberg, Rozen, Hern{\'{a}}ndez{-}Orallo, Boudeman, Guerr, Jones, Tenenbaum, Rule, Chua, Kanclerz, Livescu, Krauth, Gopalakrishnan, Ignatyeva, Markert, Dhole, Gimpel, Omondi, Mathewson, Chiafullo, Shkaruta, Shridhar, McDonell, Richardson, Reynolds, Gao, Zhang, Dugan, Qin, Ochando,
  Morency, Moschella, Lam, Noble, Schmidt, He, Col{\'{o}}n, Metz, Senel, Bosma, Sap, ter Hoeve, Farooqi, Faruqui, Mazeika, Baturan, Marelli, Maru, Ram{\'{\i}}rez{-}Quintana, Tolkiehn, Giulianelli, Lewis, Potthast, Leavitt, Hagen, Schubert, Baitemirova, Arnaud, McElrath, Yee, Cohen, Gu, Ivanitskiy, Starritt, Strube, Swedrowski, Bevilacqua, Yasunaga, Kale, Cain, Xu, Suzgun, Walker, Tiwari, Bansal, Aminnaseri, Geva, Gheini, T., Peng, Chi, Lee, Krakover, Cameron, Roberts, Doiron, Martinez, Nangia, Deckers, Muennighoff, Keskar, Iyer, Constant, Fiedel, Wen, Zhang, Agha, Elbaghdadi, Levy, Evans, Casares, Doshi, Fung, Liang, Vicol, Alipoormolabashi, Liao, Liang, Chang, Eckersley, Htut, Hwang, Milkowski, Patil, Pezeshkpour, Oli, Mei, Lyu, Chen, Banjade, Rudolph, Gabriel, Habacker, Risco, Milli{\`{e}}re, Garg, Barnes, Saurous, Arakawa, Raymaekers, Frank, Sikand, Novak, Sitelew, LeBras, Liu, Jacobs, Zhang, Salakhutdinov, Chi, Lee, Stovall, Teehan, Yang, Singh, Mohammad, Anand, Dillavou, Shleifer, Wiseman, Gruetter,
  Bowman, Schoenholz, Han, Kwatra, Rous, Ghazarian, Ghosh, Casey, Bischoff, Gehrmann, Schuster, Sadeghi, Hamdan, Zhou, Srivastava, Shi, Singh, Asaadi, Gu, Pachchigar, Toshniwal, Upadhyay, Debnath, Shakeri, Thormeyer, Melzi, Reddy, Makini, Lee, Torene, Hatwar, Dehaene, Divic, Ermon, Biderman, Lin, Prasad, Piantadosi, Shieber, Misherghi, Kiritchenko, Mishra, Linzen, Schuster, Li, Yu, Ali, Hashimoto, Wu, Desbordes, Rothschild, Phan, Wang, Nkinyili, Schick, Kornev, Tunduny, Gerstenberg, Chang, Neeraj, Khot, Shultz, Shaham, Misra, Demberg, Nyamai, Raunak, Ramasesh, Prabhu, Padmakumar, Srikumar, Fedus, Saunders, Zhang, Vossen, Ren, Tong, Zhao, Wu, Shen, Yaghoobzadeh, Lakretz, Song, Bahri, Choi, Yang, Hao, Chen, Belinkov, Hou, Hou, Bai, Seid, Zhao, Wang, Wang, Wang, and Wu]{bigbench}
Aarohi Srivastava, Abhinav Rastogi, Abhishek Rao, Abu Awal~Md Shoeb, Abubakar Abid, Adam Fisch, Adam~R. Brown, Adam Santoro, Aditya Gupta, Adri{\`{a}} Garriga{-}Alonso, Agnieszka Kluska, Aitor Lewkowycz, Akshat Agarwal, Alethea Power, Alex Ray, Alex Warstadt, Alexander~W. Kocurek, Ali Safaya, Ali Tazarv, Alice Xiang, Alicia Parrish, Allen Nie, Aman Hussain, Amanda Askell, Amanda Dsouza, Ambrose Slone, Ameet Rahane, Anantharaman~S. Iyer, Anders Andreassen, Andrea Madotto, Andrea Santilli, Andreas Stuhlm{\"{u}}ller, Andrew~M. Dai, Andrew La, Andrew~K. Lampinen, Andy Zou, Angela Jiang, Angelica Chen, Anh Vuong, Animesh Gupta, Anna Gottardi, Antonio Norelli, Anu Venkatesh, Arash Gholamidavoodi, Arfa Tabassum, Arul Menezes, Arun Kirubarajan, Asher Mullokandov, Ashish Sabharwal, Austin Herrick, Avia Efrat, Aykut Erdem, Ayla Karakas, B.~Ryan Roberts, Bao~Sheng Loe, Barret Zoph, Bartlomiej Bojanowski, Batuhan {\"{O}}zyurt, Behnam Hedayatnia, Behnam Neyshabur, Benjamin Inden, Benno Stein, Berk Ekmekci, Bill~Yuchen
  Lin, Blake Howald, Bryan Orinion, Cameron Diao, Cameron Dour, Catherine Stinson, Cedrick Argueta, C{\`{e}}sar~Ferri Ram{\'{\i}}rez, Chandan Singh, Charles Rathkopf, Chenlin Meng, Chitta Baral, Chiyu Wu, Chris Callison{-}Burch, Chris Waites, Christian Voigt, Christopher~D. Manning, Christopher Potts, Cindy Ramirez, Clara~E. Rivera, Clemencia Siro, Colin Raffel, Courtney Ashcraft, Cristina Garbacea, Damien Sileo, Dan Garrette, Dan Hendrycks, Dan Kilman, Dan Roth, Daniel Freeman, Daniel Khashabi, Daniel Levy, Daniel~Mosegu{\'{\i}} Gonz{\'{a}}lez, Danielle Perszyk, Danny Hernandez, Danqi Chen, Daphne Ippolito, Dar Gilboa, David Dohan, David Drakard, David Jurgens, Debajyoti Datta, Deep Ganguli, Denis Emelin, Denis Kleyko, Deniz Yuret, Derek Chen, Derek Tam, Dieuwke Hupkes, Diganta Misra, Dilyar Buzan, Dimitri~Coelho Mollo, Diyi Yang, Dong{-}Ho Lee, Dylan Schrader, Ekaterina Shutova, Ekin~Dogus Cubuk, Elad Segal, Eleanor Hagerman, Elizabeth Barnes, Elizabeth Donoway, Ellie Pavlick, Emanuele Rodol{\`{a}}, Emma
  Lam, Eric Chu, Eric Tang, Erkut Erdem, Ernie Chang, Ethan~A. Chi, Ethan Dyer, Ethan~J. Jerzak, Ethan Kim, Eunice~Engefu Manyasi, Evgenii Zheltonozhskii, Fanyue Xia, Fatemeh Siar, Fernando Mart{\'{\i}}nez{-}Plumed, Francesca Happ{\'{e}}, Fran{\c{c}}ois Chollet, Frieda Rong, Gaurav Mishra, Genta~Indra Winata, Gerard de~Melo, Germ{\'{a}}n Kruszewski, Giambattista Parascandolo, Giorgio Mariani, Gloria Wang, Gonzalo Jaimovitch{-}L{\'{o}}pez, Gregor Betz, Guy Gur{-}Ari, Hana Galijasevic, Hannah Kim, Hannah Rashkin, Hannaneh Hajishirzi, Harsh Mehta, Hayden Bogar, Henry Shevlin, Hinrich Sch{\"{u}}tze, Hiromu Yakura, Hongming Zhang, Hugh~Mee Wong, Ian Ng, Isaac Noble, Jaap Jumelet, Jack Geissinger, Jackson Kernion, Jacob Hilton, Jaehoon Lee, Jaime~Fern{\'{a}}ndez Fisac, James~B. Simon, James Koppel, James Zheng, James Zou, Jan Kocon, Jana Thompson, Janelle Wingfield, Jared Kaplan, Jarema Radom, Jascha Sohl{-}Dickstein, Jason Phang, Jason Wei, Jason Yosinski, Jekaterina Novikova, Jelle Bosscher, Jennifer Marsh,
  Jeremy Kim, Jeroen Taal, Jesse~H. Engel, Jesujoba Alabi, Jiacheng Xu, Jiaming Song, Jillian Tang, Joan Waweru, John Burden, John Miller, John~U. Balis, Jonathan Batchelder, Jonathan Berant, J{\"{o}}rg Frohberg, Jos Rozen, Jos{\'{e}} Hern{\'{a}}ndez{-}Orallo, Joseph Boudeman, Joseph Guerr, Joseph Jones, Joshua~B. Tenenbaum, Joshua~S. Rule, Joyce Chua, Kamil Kanclerz, Karen Livescu, Karl Krauth, Karthik Gopalakrishnan, Katerina Ignatyeva, Katja Markert, Kaustubh~D. Dhole, Kevin Gimpel, Kevin Omondi, Kory Mathewson, Kristen Chiafullo, Ksenia Shkaruta, Kumar Shridhar, Kyle McDonell, Kyle Richardson, Laria Reynolds, Leo Gao, Li~Zhang, Liam Dugan, Lianhui Qin, Lidia~Contreras Ochando, Louis{-}Philippe Morency, Luca Moschella, Lucas Lam, Lucy Noble, Ludwig Schmidt, Luheng He, Luis~Oliveros Col{\'{o}}n, Luke Metz, L{\"{u}}tfi~Kerem Senel, Maarten Bosma, Maarten Sap, Maartje ter Hoeve, Maheen Farooqi, Manaal Faruqui, Mantas Mazeika, Marco Baturan, Marco Marelli, Marco Maru, Mar{\'{\i}}a~Jos{\'{e}}
  Ram{\'{\i}}rez{-}Quintana, Marie Tolkiehn, Mario Giulianelli, Martha Lewis, Martin Potthast, Matthew~L. Leavitt, Matthias Hagen, M{\'{a}}ty{\'{a}}s Schubert, Medina Baitemirova, Melody Arnaud, Melvin McElrath, Michael~A. Yee, Michael Cohen, Michael Gu, Michael~I. Ivanitskiy, Michael Starritt, Michael Strube, Michal Swedrowski, Michele Bevilacqua, Michihiro Yasunaga, Mihir Kale, Mike Cain, Mimee Xu, Mirac Suzgun, Mitch Walker, Mo~Tiwari, Mohit Bansal, Moin Aminnaseri, Mor Geva, Mozhdeh Gheini, Mukund~Varma T., Nanyun Peng, Nathan~A. Chi, Nayeon Lee, Neta~Gur{-}Ari Krakover, Nicholas Cameron, Nicholas Roberts, Nick Doiron, Nicole Martinez, Nikita Nangia, Niklas Deckers, Niklas Muennighoff, Nitish~Shirish Keskar, Niveditha Iyer, Noah Constant, Noah Fiedel, Nuan Wen, Oliver Zhang, Omar Agha, Omar Elbaghdadi, Omer Levy, Owain Evans, Pablo Antonio~Moreno Casares, Parth Doshi, Pascale Fung, Paul~Pu Liang, Paul Vicol, Pegah Alipoormolabashi, Peiyuan Liao, Percy Liang, Peter Chang, Peter Eckersley, Phu~Mon Htut,
  Pinyu Hwang, Piotr Milkowski, Piyush Patil, Pouya Pezeshkpour, Priti Oli, Qiaozhu Mei, Qing Lyu, Qinlang Chen, Rabin Banjade, Rachel~Etta Rudolph, Raefer Gabriel, Rahel Habacker, Ramon Risco, Rapha{\"{e}}l Milli{\`{e}}re, Rhythm Garg, Richard Barnes, Rif~A. Saurous, Riku Arakawa, Robbe Raymaekers, Robert Frank, Rohan Sikand, Roman Novak, Roman Sitelew, Ronan LeBras, Rosanne Liu, Rowan Jacobs, Rui Zhang, Ruslan Salakhutdinov, Ryan Chi, Ryan Lee, Ryan Stovall, Ryan Teehan, Rylan Yang, Sahib Singh, Saif~M. Mohammad, Sajant Anand, Sam Dillavou, Sam Shleifer, Sam Wiseman, Samuel Gruetter, Samuel~R. Bowman, Samuel~S. Schoenholz, Sanghyun Han, Sanjeev Kwatra, Sarah~A. Rous, Sarik Ghazarian, Sayan Ghosh, Sean Casey, Sebastian Bischoff, Sebastian Gehrmann, Sebastian Schuster, Sepideh Sadeghi, Shadi Hamdan, Sharon Zhou, Shashank Srivastava, Sherry Shi, Shikhar Singh, Shima Asaadi, Shixiang~Shane Gu, Shubh Pachchigar, Shubham Toshniwal, Shyam Upadhyay, Shyamolima~(Shammie) Debnath, Siamak Shakeri, Simon Thormeyer,
  Simone Melzi, Siva Reddy, Sneha~Priscilla Makini, Soo{-}Hwan Lee, Spencer Torene, Sriharsha Hatwar, Stanislas Dehaene, Stefan Divic, Stefano Ermon, Stella Biderman, Stephanie Lin, Stephen Prasad, Steven~T. Piantadosi, Stuart~M. Shieber, Summer Misherghi, Svetlana Kiritchenko, Swaroop Mishra, Tal Linzen, Tal Schuster, Tao Li, Tao Yu, Tariq Ali, Tatsu Hashimoto, Te{-}Lin Wu, Th{\'{e}}o Desbordes, Theodore Rothschild, Thomas Phan, Tianle Wang, Tiberius Nkinyili, Timo Schick, Timofei Kornev, Titus Tunduny, Tobias Gerstenberg, Trenton Chang, Trishala Neeraj, Tushar Khot, Tyler Shultz, Uri Shaham, Vedant Misra, Vera Demberg, Victoria Nyamai, Vikas Raunak, Vinay~V. Ramasesh, Vinay~Uday Prabhu, Vishakh Padmakumar, Vivek Srikumar, William Fedus, William Saunders, William Zhang, Wout Vossen, Xiang Ren, Xiaoyu Tong, Xinran Zhao, Xinyi Wu, Xudong Shen, Yadollah Yaghoobzadeh, Yair Lakretz, Yangqiu Song, Yasaman Bahri, Yejin Choi, Yichi Yang, Yiding Hao, Yifu Chen, Yonatan Belinkov, Yu~Hou, Yufang Hou, Yuntao Bai,
  Zachary Seid, Zhuoye Zhao, Zijian Wang, Zijie~J. Wang, Zirui Wang, and Ziyi Wu.
\newblock Beyond the imitation game: Quantifying and extrapolating the capabilities of language models.
\newblock \emph{Trans. Mach. Learn. Res.}, 2023, 2023.

\bibitem[Zellers et~al.(2019)Zellers, Holtzman, Bisk, Farhadi, and Choi]{hellaswag}
Rowan Zellers, Ari Holtzman, Yonatan Bisk, Ali Farhadi, and Yejin Choi.
\newblock Hellaswag: Can a machine really finish your sentence?
\newblock In Anna Korhonen, David~R. Traum, and Llu{\'{\i}}s M{\`{a}}rquez, editors, \emph{Proceedings of the 57th Conference of the Association for Computational Linguistics, {ACL} 2019, Florence, Italy, July 28- August 2, 2019, Volume 1: Long Papers}, pages 4791--4800. Association for Computational Linguistics, 2019.
\newblock \doi{10.18653/V1/P19-1472}.
\newblock URL \url{https://doi.org/10.18653/v1/p19-1472}.

\bibitem[Bisk et~al.(2020)Bisk, Zellers, Bras, Gao, and Choi]{piqa}
Yonatan Bisk, Rowan Zellers, Ronan~Le Bras, Jianfeng Gao, and Yejin Choi.
\newblock {PIQA:} reasoning about physical commonsense in natural language.
\newblock In \emph{The Thirty-Fourth {AAAI} Conference on Artificial Intelligence, {AAAI} 2020, The Thirty-Second Innovative Applications of Artificial Intelligence Conference, {IAAI} 2020, The Tenth {AAAI} Symposium on Educational Advances in Artificial Intelligence, {EAAI} 2020, New York, NY, USA, February 7-12, 2020}, pages 7432--7439. {AAAI} Press, 2020.
\newblock \doi{10.1609/AAAI.V34I05.6239}.
\newblock URL \url{https://doi.org/10.1609/aaai.v34i05.6239}.

\bibitem[Sakaguchi et~al.(2020)Sakaguchi, Bras, Bhagavatula, and Choi]{winogrande}
Keisuke Sakaguchi, Ronan~Le Bras, Chandra Bhagavatula, and Yejin Choi.
\newblock Winogrande: An adversarial winograd schema challenge at scale.
\newblock In \emph{The Thirty-Fourth {AAAI} Conference on Artificial Intelligence, {AAAI} 2020, The Thirty-Second Innovative Applications of Artificial Intelligence Conference, {IAAI} 2020, The Tenth {AAAI} Symposium on Educational Advances in Artificial Intelligence, {EAAI} 2020, New York, NY, USA, February 7-12, 2020}, pages 8732--8740. {AAAI} Press, 2020.
\newblock \doi{10.1609/AAAI.V34I05.6399}.
\newblock URL \url{https://doi.org/10.1609/aaai.v34i05.6399}.

\bibitem[Clark et~al.(2019)Clark, Lee, Chang, Kwiatkowski, Collins, and Toutanova]{boolq}
Christopher Clark, Kenton Lee, Ming{-}Wei Chang, Tom Kwiatkowski, Michael Collins, and Kristina Toutanova.
\newblock Boolq: Exploring the surprising difficulty of natural yes/no questions.
\newblock In Jill Burstein, Christy Doran, and Thamar Solorio, editors, \emph{Proceedings of the 2019 Conference of the North American Chapter of the Association for Computational Linguistics: Human Language Technologies, {NAACL-HLT} 2019, Minneapolis, MN, USA, June 2-7, 2019, Volume 1 (Long and Short Papers)}, pages 2924--2936. Association for Computational Linguistics, 2019.
\newblock \doi{10.18653/V1/N19-1300}.
\newblock URL \url{https://doi.org/10.18653/v1/n19-1300}.

\bibitem[Mihaylov et~al.(2018)Mihaylov, Clark, Khot, and Sabharwal]{openbookqa}
Todor Mihaylov, Peter Clark, Tushar Khot, and Ashish Sabharwal.
\newblock Can a suit of armor conduct electricity? {A} new dataset for open book question answering.
\newblock In Ellen Riloff, David Chiang, Julia Hockenmaier, and Jun'ichi Tsujii, editors, \emph{Proceedings of the 2018 Conference on Empirical Methods in Natural Language Processing, Brussels, Belgium, October 31 - November 4, 2018}, pages 2381--2391. Association for Computational Linguistics, 2018.
\newblock \doi{10.18653/V1/D18-1260}.
\newblock URL \url{https://doi.org/10.18653/v1/d18-1260}.

\bibitem[Lo et~al.(2023)Lo, Sadrzadeh, and Mansfield]{winograd}
Kin~Ian Lo, Mehrnoosh Sadrzadeh, and Shane Mansfield.
\newblock Generalised winograd schema and its contextuality.
\newblock In Shane Mansfield, Beno{\^{\i}}t Valiron, and Vladimir Zamdzhiev, editors, \emph{Proceedings of the Twentieth International Conference on Quantum Physics and Logic, {QPL} 2023, Paris, France, 17-21st July 2023}, volume 384 of \emph{{EPTCS}}, pages 187--202, 2023.
\newblock \doi{10.4204/EPTCS.384.11}.
\newblock URL \url{https://doi.org/10.4204/EPTCS.384.11}.

\bibitem[Paperno et~al.(2016)Paperno, Kruszewski, Lazaridou, Pham, Bernardi, Pezzelle, Baroni, Boleda, and Fern{\'{a}}ndez]{lambada}
Denis Paperno, Germ{\'{a}}n Kruszewski, Angeliki Lazaridou, Quan~Ngoc Pham, Raffaella Bernardi, Sandro Pezzelle, Marco Baroni, Gemma Boleda, and Raquel Fern{\'{a}}ndez.
\newblock The {LAMBADA} dataset: Word prediction requiring a broad discourse context.
\newblock In \emph{Proceedings of the 54th Annual Meeting of the Association for Computational Linguistics, {ACL} 2016, August 7-12, 2016, Berlin, Germany, Volume 1: Long Papers}. The Association for Computer Linguistics, 2016.
\newblock \doi{10.18653/V1/P16-1144}.
\newblock URL \url{https://doi.org/10.18653/v1/p16-1144}.

\bibitem[Roemmele et~al.(2011)Roemmele, Bejan, and Gordon]{copa}
Melissa Roemmele, Cosmin~Adrian Bejan, and Andrew~S. Gordon.
\newblock Choice of plausible alternatives: An evaluation of commonsense causal reasoning.
\newblock In \emph{Logical Formalizations of Commonsense Reasoning, Papers from the 2011 {AAAI} Spring Symposium, Technical Report SS-11-06, Stanford, California, USA, March 21-23, 2011}. {AAAI}, 2011.
\newblock URL \url{http://www.aaai.org/ocs/index.php/SSS/SSS11/paper/view/2418}.

\bibitem[Hendrycks et~al.(2021)Hendrycks, Burns, Basart, Zou, Mazeika, Song, and Steinhardt]{mmlu}
Dan Hendrycks, Collin Burns, Steven Basart, Andy Zou, Mantas Mazeika, Dawn Song, and Jacob Steinhardt.
\newblock Measuring massive multitask language understanding.
\newblock In \emph{{ICLR}}. OpenReview.net, 2021.

\end{thebibliography}
}

\newpage
\appendix


\section{Additional Results}

\begin{figure}[ht]
    \centering
    \subfloat[]{\includegraphics[width=0.45\textwidth]{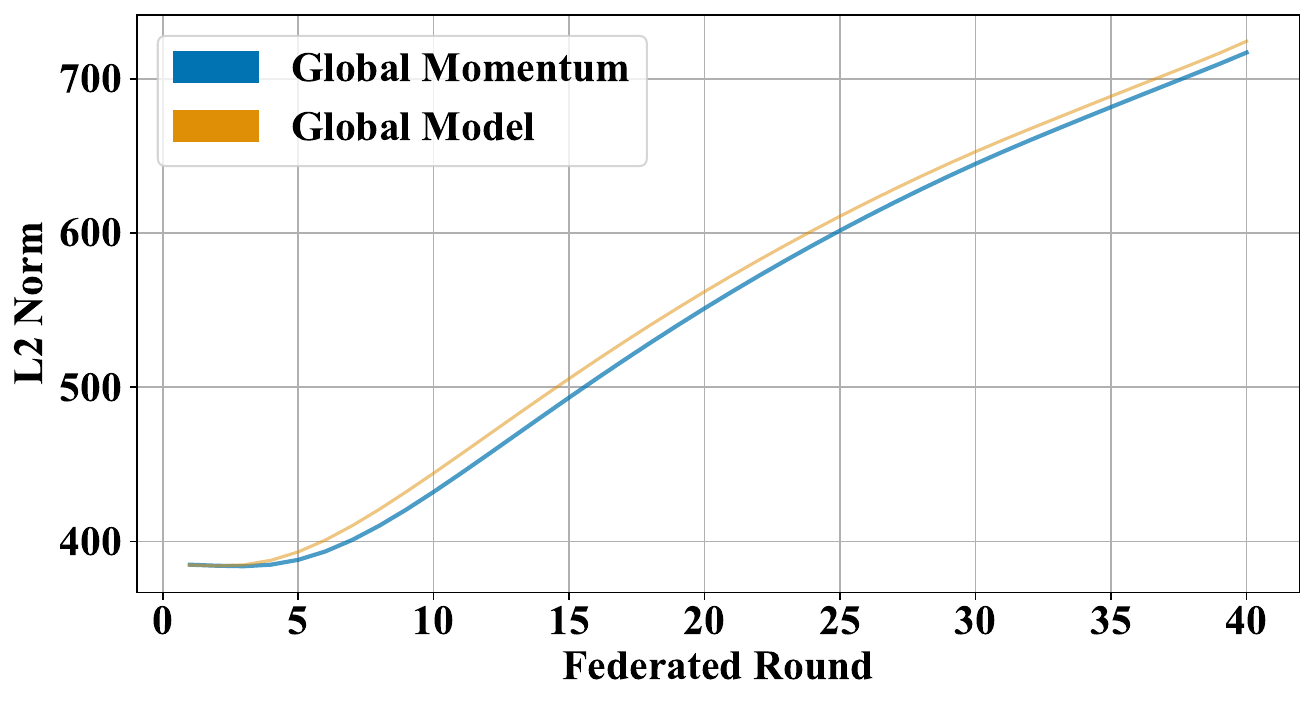}} 
    \subfloat[]{\includegraphics[width=0.45\textwidth]{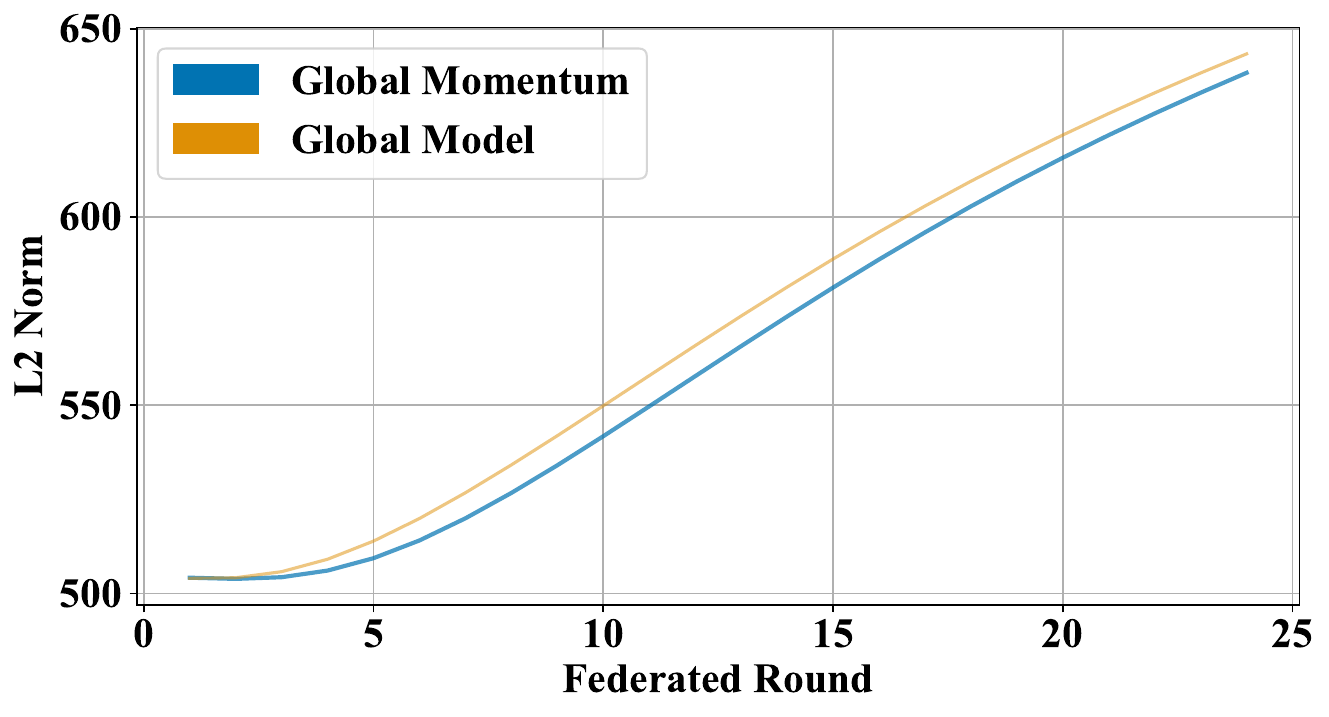}}\\
    \subfloat[]{\includegraphics[width=0.45\textwidth]{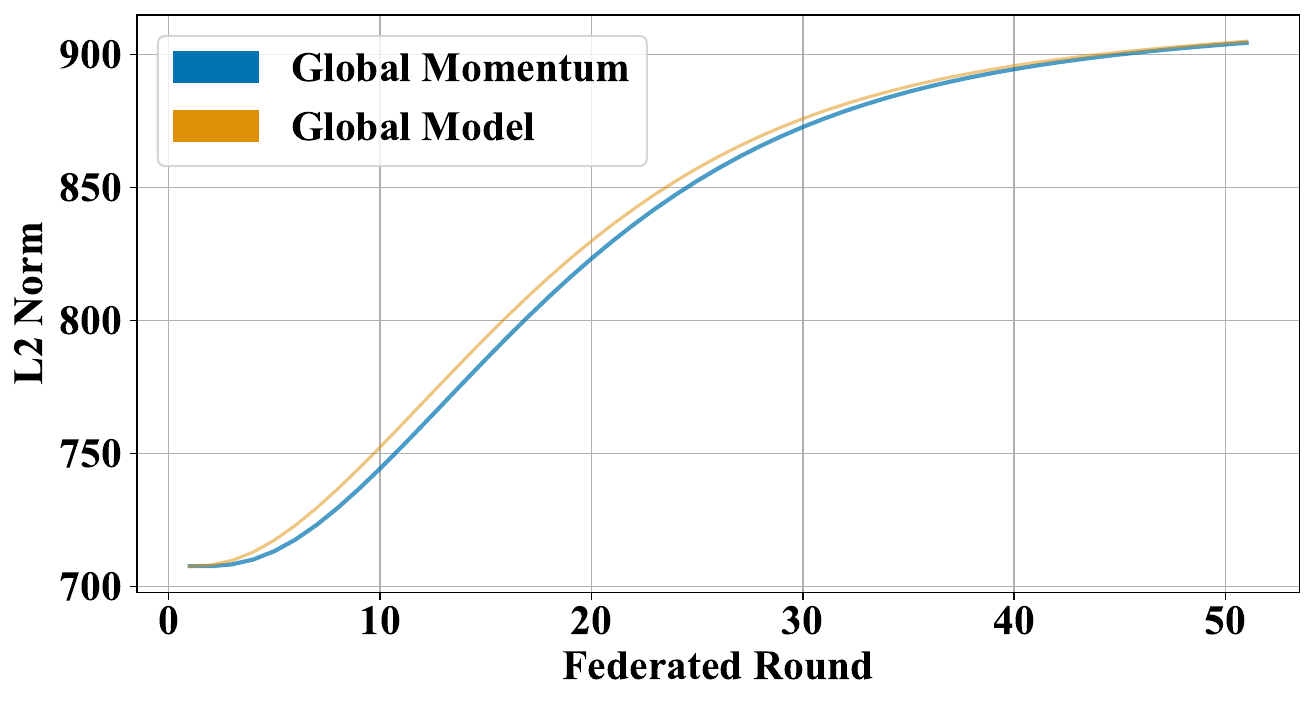}}
    \subfloat[]{\includegraphics[width=0.45\textwidth]{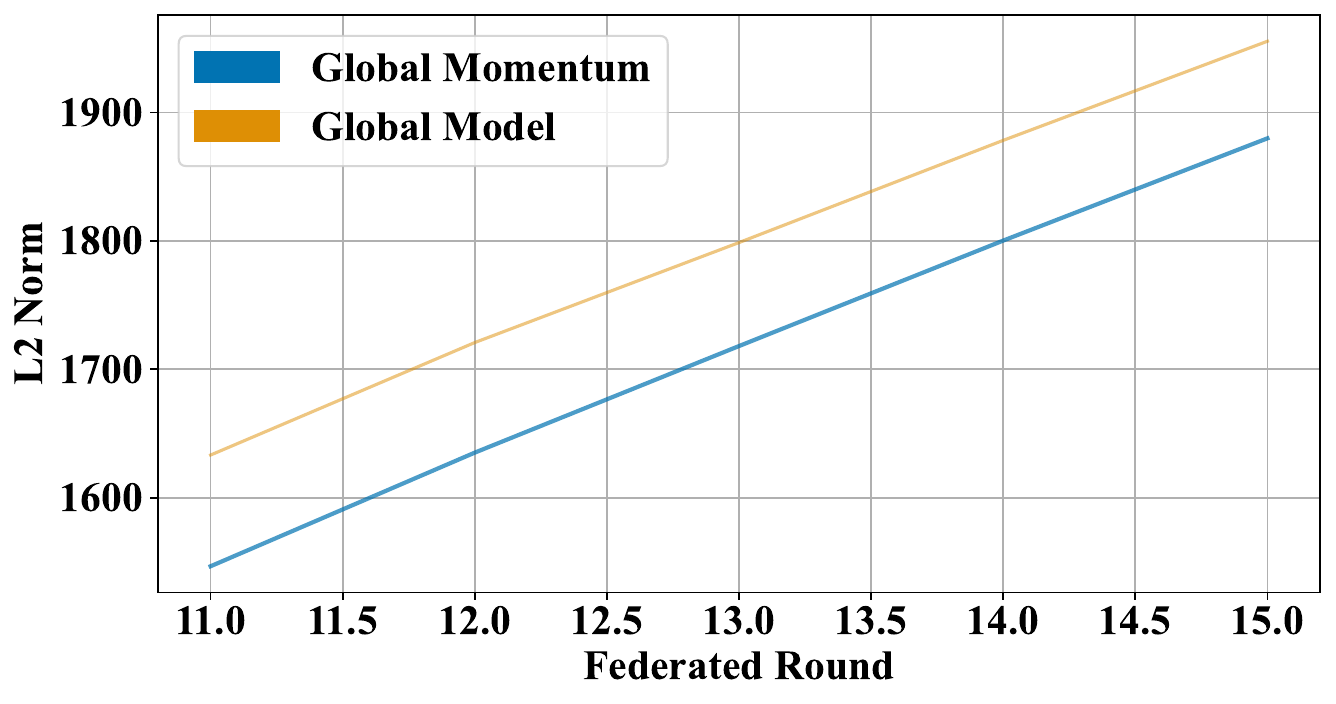}} 
    \caption{The $l_2$ norm of the global model versus the server-side Nesterov momentum, which seeks to track an exponential moving average of the model using $\beta=0.7$, for our 75M~(a), 125M~(b), 350M~(c), and 1.3B~(d) experiments. }
    \label{fig:fed:momentum_vs_model-(generic-scale)}
\end{figure} 


\begin{figure}[ht]
    \centering
    \subfloat[]{\includegraphics[width=0.45\textwidth]{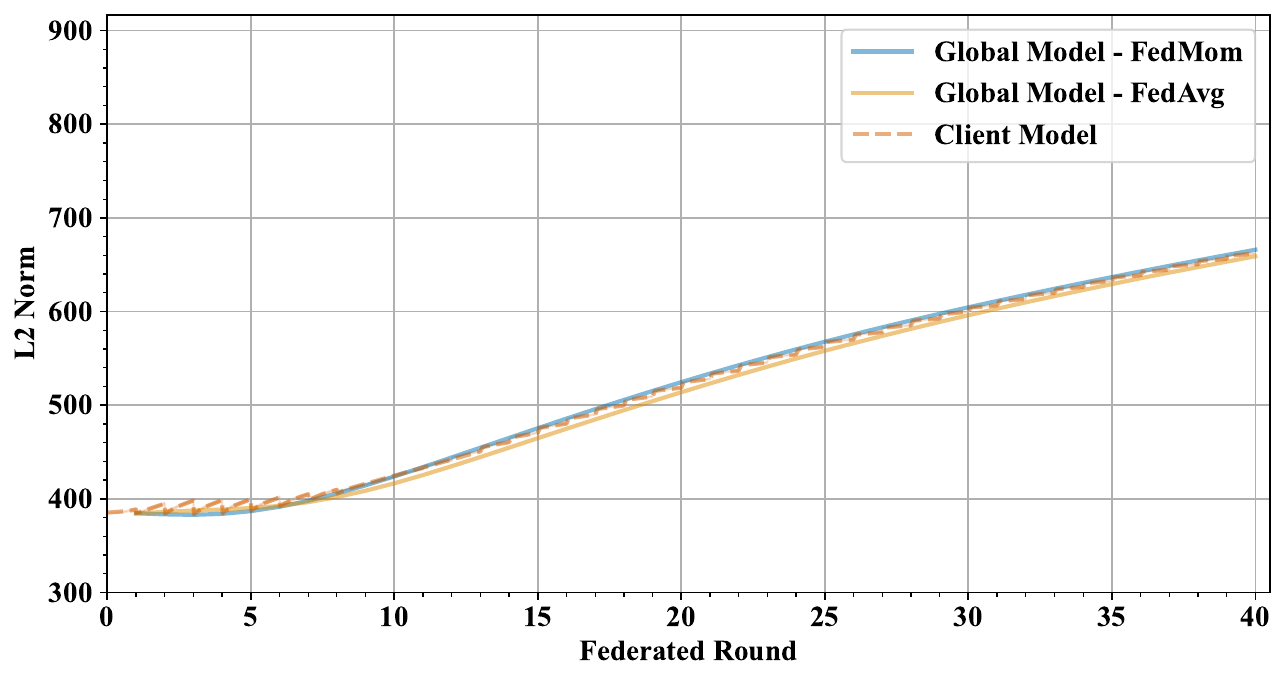}}
    \subfloat[]{\includegraphics[width=0.45\textwidth]{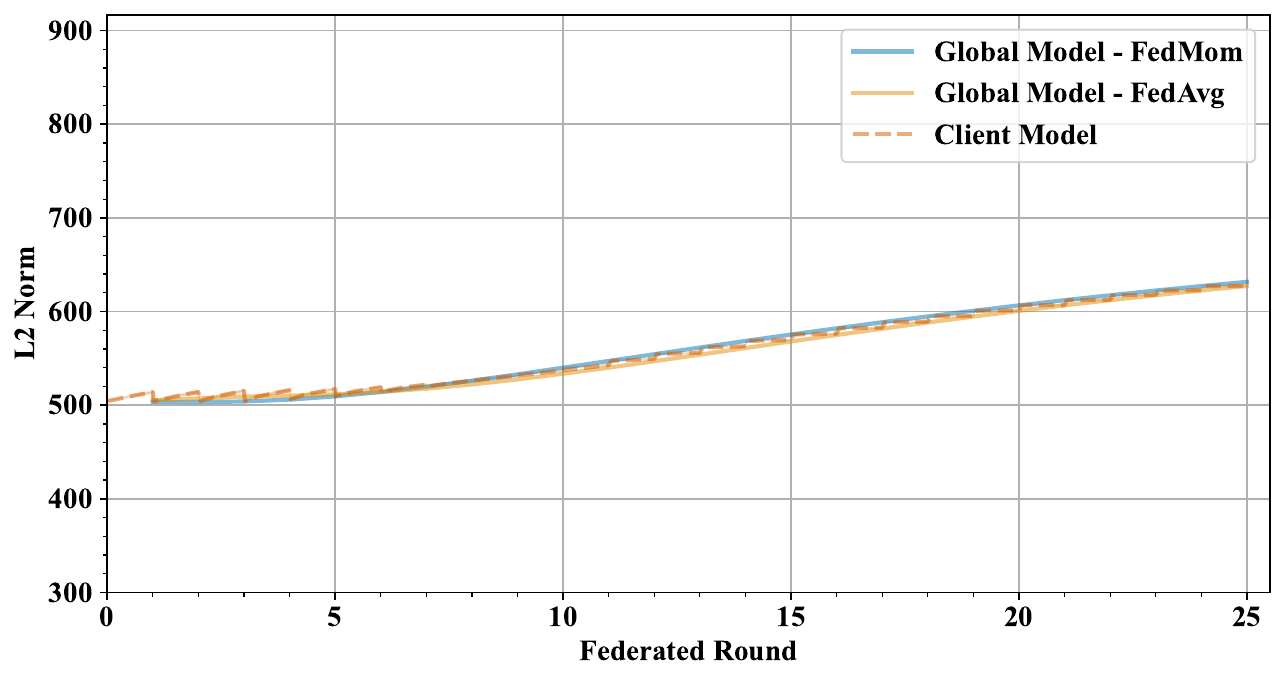}}
    \caption{The $l_2$ norms of the global model, client models, and the average of client models for our $75$M~(a) and $125$M~(b) experiments on our naturally heterogeneous partition of \emph{The Pile}. Despite the heterogeneous data, the momentum mechanism of the server and the decaying learning rate allow it to enforce a consensus effectively. }
    \label{fig:fed:norm-pile}
\end{figure} 

\begin{figure}[ht]
    \centering
    \subfloat[]{\includegraphics[width=0.45\textwidth]{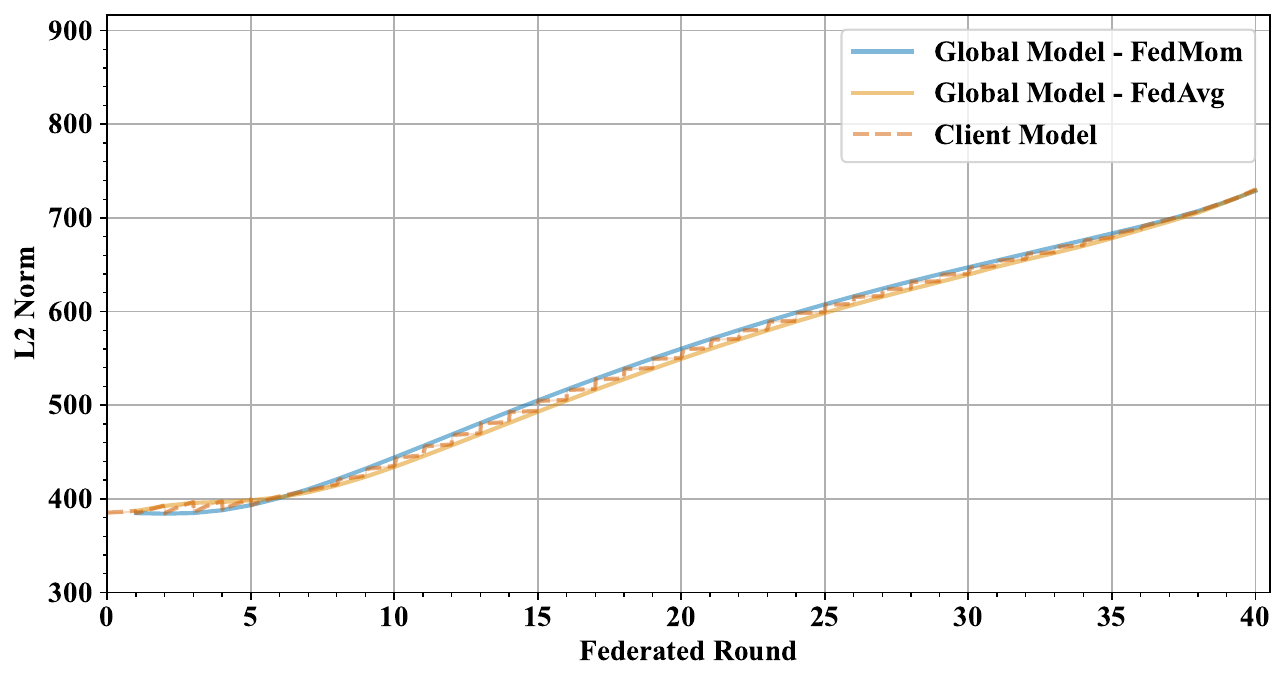}}
    \subfloat[]{\includegraphics[width=0.45\textwidth]{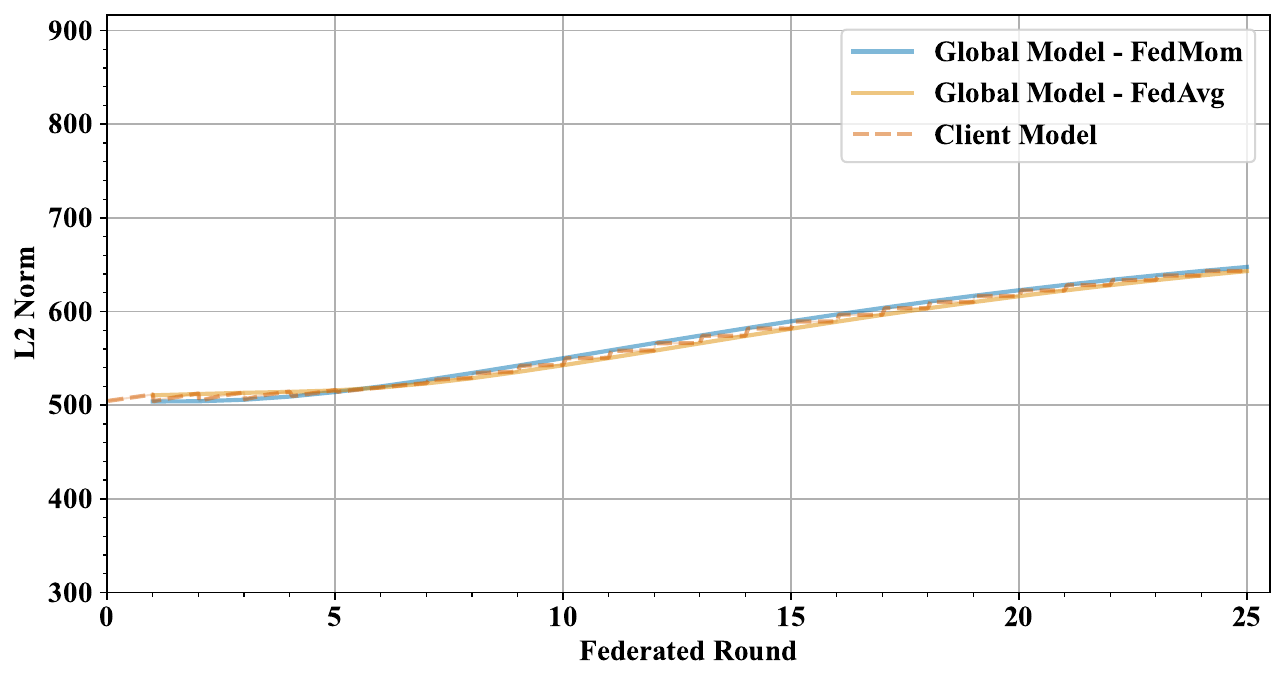}}
    \caption{The $l_2$ norms of the global model, client models, and the average of client models for our $75$M~(a) and $125$M~(b) experiments on our partial participation experiments using $4$ clients per round out of $64$ total \emph{C4} clients, i.e., the $6\%$ of the federated population. Despite the FL using only a small sample of clients every round, the global model follows the exact same trend as in the full participation scenario.}
    \label{fig:fed:norm-partial}
\end{figure}

\begin{figure}[ht]
    \centering
    \subfloat[]{\includegraphics[width=0.45\textwidth]{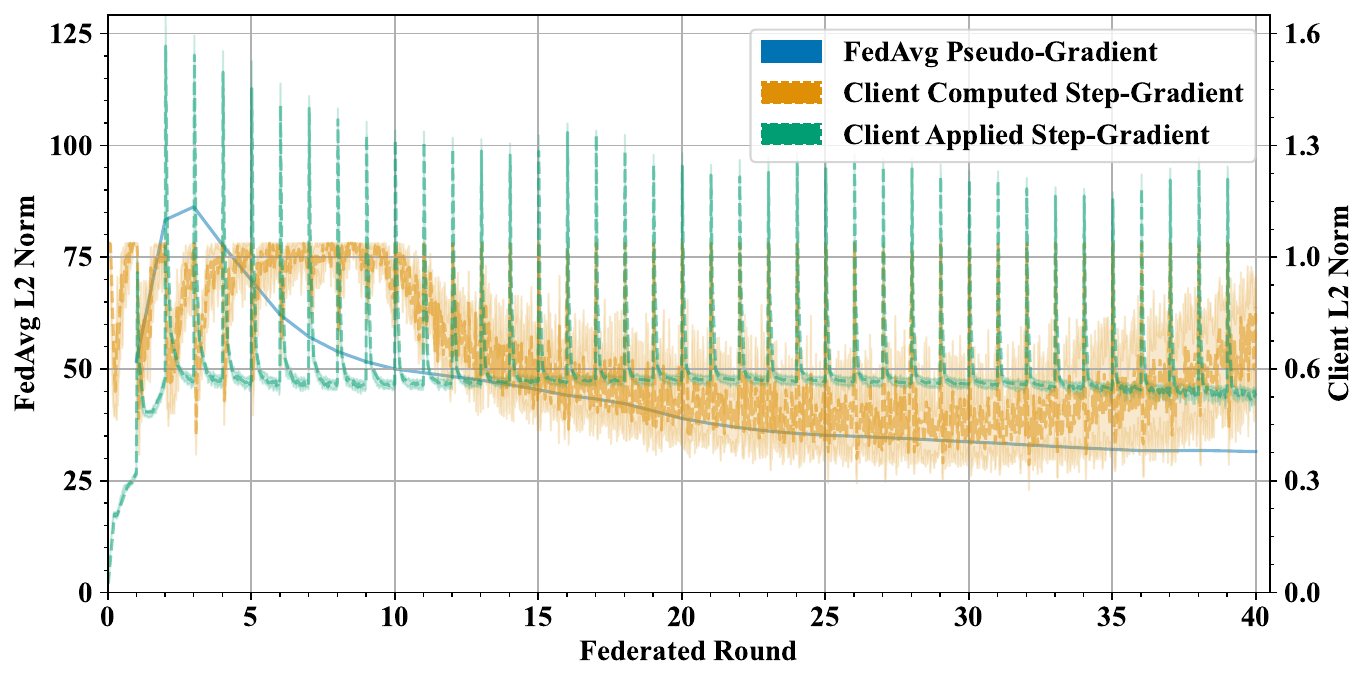}}
    \subfloat[]{\includegraphics[width=0.45\textwidth]{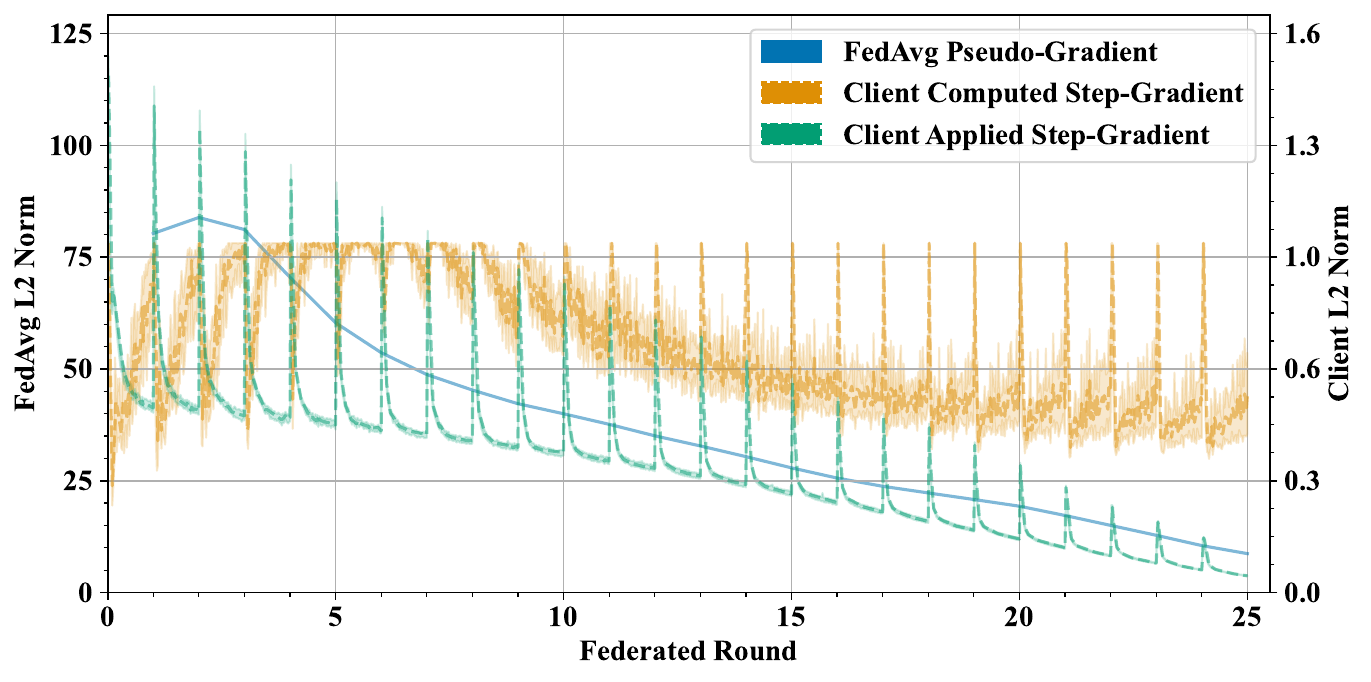}}
    \caption{The $l_2$ norms, for our $75$M~(a) and $125$M~(b) experiments on \emph{The Pile}, of the FedAvg Pseudo-Gradient~(average of client deltas relative to the server model of the previous round), the client model gradients computed on a per-step basis, and the client gradients applied to the model when considering learning rate, weight decay, and clipping. Unlike the IID case, the FedAvg Pseudo-Gradient decays much faster than local client gradients, indicating that its decrease results from the model adapting to data heterogeneity rather than being a consequence of learning rate decay.}
    \label{fig:fed:pseudograd-vs-localgrad-pile}
\end{figure}

\begin{figure}[ht]
    \centering
    \subfloat[]{\includegraphics[width=0.45\textwidth]{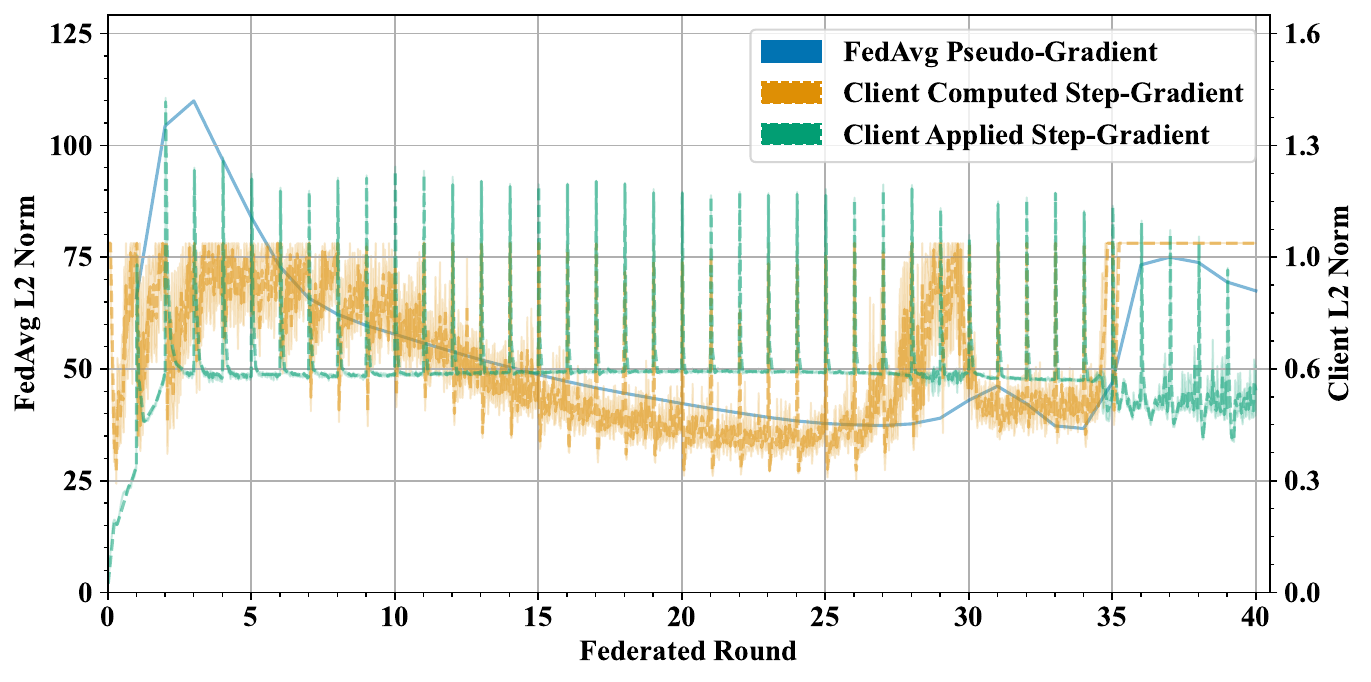}}
    \subfloat[]{\includegraphics[width=0.45\textwidth]{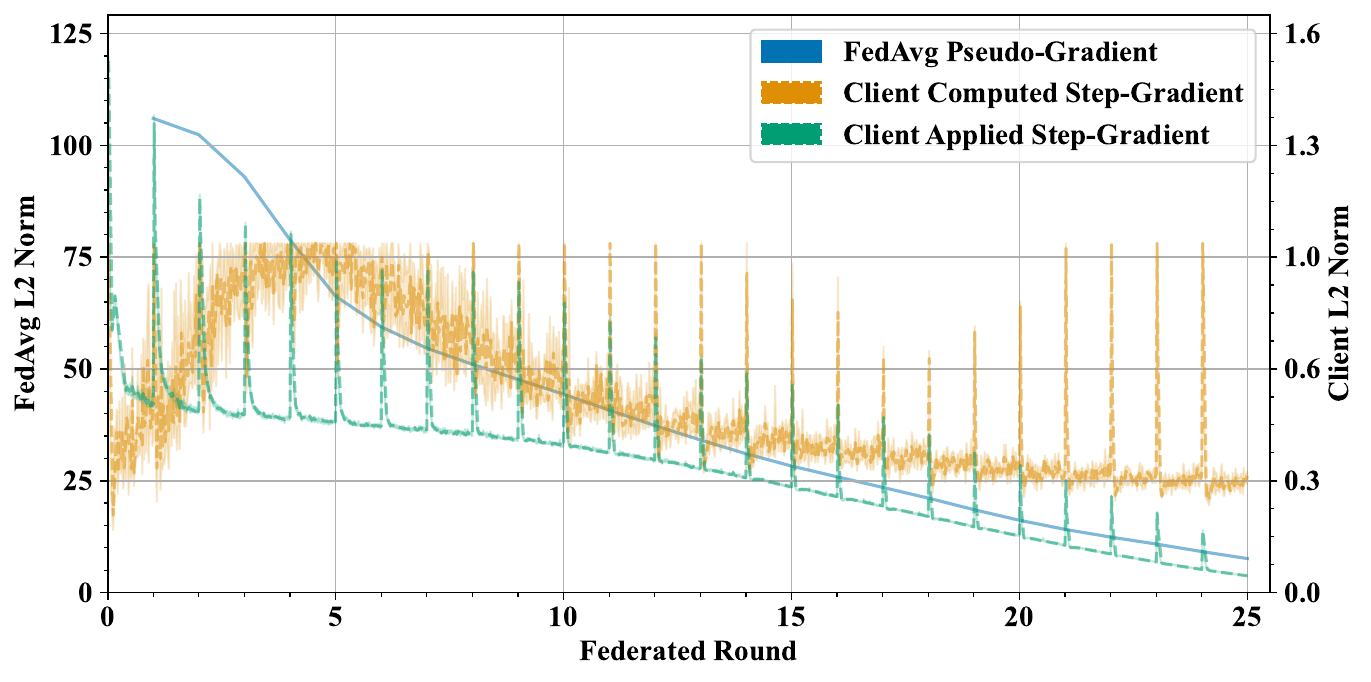}}
    \caption{The $l_2$ norms, for our $75$M~(a) and $125$M~(b) partial participation experiments on \emph{C4}, of the FedAvg Pseudo-Gradient~(average of client deltas relative to the server model of the previous round), the client model gradients computed on a per-step basis, and the client gradients applied to the model when considering learning rate, weight decay, and clipping. The relationship between the FedAvg Pseudo-Gradient and the local steps is the same as in the full-participation case despite a very small sample of clients being considered every round.}
    \label{fig:fed:pseudograd-vs-localgrad-partial}
\end{figure} 

\FloatBarrier

\end{document}